\documentclass[10pt,twocolumn,letterpaper]{article} 
 
\usepackage{iccv} 
\usepackage{times} 
\usepackage{epsfig} 
\usepackage{graphicx} 
\usepackage{amsmath} 
\usepackage{amssymb} 
 
\usepackage[pagebackref=true,breaklinks=true,letterpaper=true,colorlinks,bookmarks=false]{hyperref} 
 
\iccvfinalcopy 
 
\def\iccvPaperID{4} 

\ificcvfinal\fi 
 
\begin{document} 
 
\title{Few-shot $\mathbf{1/a}$ Anomalies Feedback: Damage Vision Mining Opportunity and Embedding Feature Imbalance} 
 
\author{Takato Yasuno\\ 
Research Institute for Infrastructure Paradigm Shift 
{\tt\small yasunotkt@gmail.com} 
} 
 
\maketitle 
\ificcvfinal\thispagestyle{empty}\fi 
 
\begin{abstract} 
Over the past decade, previous balanced datasets have been used to advance algorithms for classification, object detection, semantic segmentation, and anomaly detection in industrial applications. In urban infrastructures and living environments, damage data mining cannot avoid imbalanced data issues because of rare unseen events and the high-quality status of improved operations. 
For visual inspection, the deteriorated class acquired from the surface of concrete and steel components are occasionally imbalanced. From numerous related surveys, we conclude that imbalanced data problems can be categorised into four types: 1) missing range of target and label valuables, 2) majority-minority class imbalance, 3) foreground background of spatial imbalance, and 4) long-tailed class of pixel-wise imbalance. Since 2015, many imbalanced studies have been conducted using deep-learning approaches, including regression, image classification, object detection, and semantic segmentation. However, anomaly detection for imbalanced data is not well known.   
In this study, we highlight a one-class anomaly detection application, whether anomalous class or not, and demonstrate clear examples of imbalanced vision datasets: medical disease, hazardous behaviour, material deterioration, plant disease, river sludge, and disaster damage.  
As illustrated in Figure \ref{fig:DamageVisionMining}, we provide key results on the advantage of damage-vision mining, hypothesising that the more effective the range of the positive ratio, the higher the accuracy gain of the anomalies feedback.  
In our imbalanced studies, compared with the balanced case with a positive ratio of $1/1$, we find that there is an applicable positive ratio $1/a$ where the accuracy is consistently high. However, the extremely imbalanced range is from one shot to $1/2a$, the accuracy of which is inferior to that of the applicable ratio. In contrast, with a positive ratio ranging over $2/a$, it shifts in the over-mining phase without an effective gain in accuracy. 
\end{abstract} 
 
\begin{figure} 
\centering 
\includegraphics[width=0.47\textwidth] 
{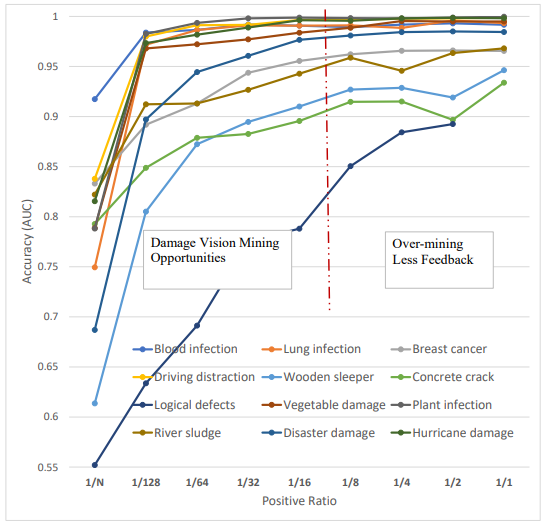} 
\caption{Illustration of damage vision mining processing on the positive ratio of target anomalous class in imbalanced datasets.} 
\label{fig:DamageVisionMining} 
\end{figure} 
 
\section{Introduction} 
\subsection{Related Works for Imbalanced Vision Data} 
\subsubsection{Imbalanced Regression} 
\begin{figure} 
\centering 
\includegraphics[width=0.43\textwidth] 
{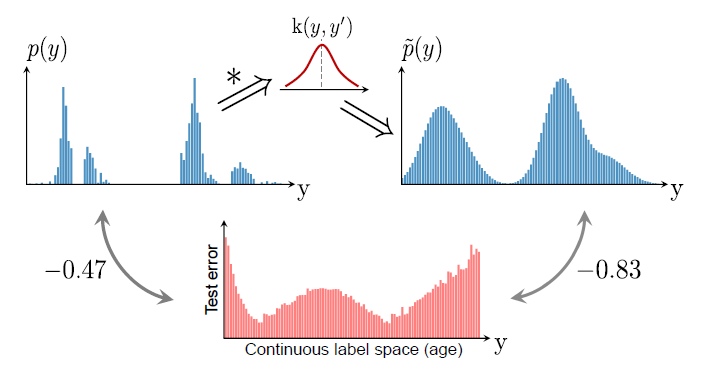} 
\caption{Imbalanced regression: e.g., Label distribution smoothing using a symmetric kernel (Yang et al. 2021).} 
\label{fig:SmoothReg} 
\end{figure}

In this study, we reviewed previous machine learning approaches for imbalanced data. As shown in Table~\ref{tab:imbset}, we identified the imbalanced settings and unbiased techniques.  
Regarding imbalanced regression, Yang et al. defined Deep Imbalanced Regression (DIR) dealing with potential missing data for continuous target values, and {\it distribution smoothing} techniques for both labels and features were presented \cite{Yang2021}, as illustrated in Figure \ref{fig:SmoothReg}. Smoothing regression techniques were applied to five datasets with missing value ranges such as age, categorical label, and health score.    
Stocksieker et al. presented a data augmentation (DA) algorithm that combines weighted resampling (WR) and a DA procedure \cite{Stocksieker2023}. Using Generalized Additive Models under synthetic data generators, six DA approaches were compared: Gaussian noise, smoothed bootstrap, k-nearest Neighbors, Gaussian Mixture Models, and Gaussian Copula. However, the effect of the mean square error using the DA-WR approach was dependent on the choice of the imbalanced data generator, and no DA technique was clearly found to improve imbalanced regression.    
 
\begin{figure} 
\centering 
\includegraphics[width=0.43\textwidth] 
{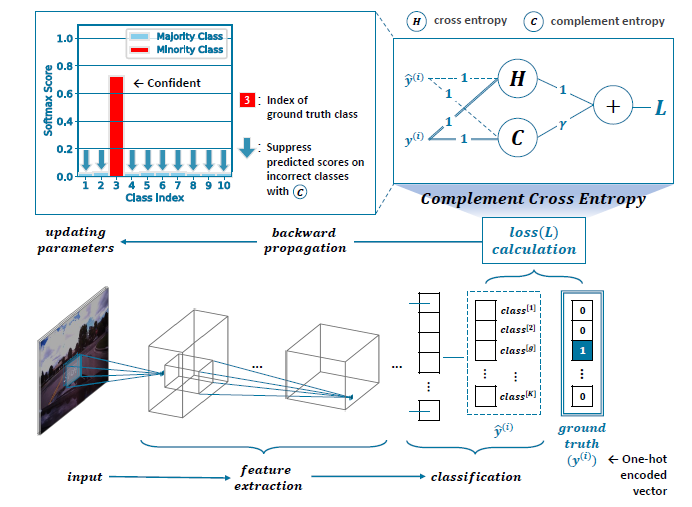} 
\caption{Imbalanced image classification: e.g., Complement cross entropy (Kim et al. 2020).} 
\label{fig:CompleCE} 
\end{figure} 
 
\subsubsection{Imbalanced Image Classification} 
To address the imbalanced image classification between majority and minority classes, a previous common preprocess is a resampling-based approach to the initial setting dataset. These are categorised as synthetic oversampling from minority classes \cite{Kim2019,Piras2012}, undersampling from majority classes \cite{Koziarski2020}, or a combination of both \cite{Batista2004,Zhu2020}.  
Another loss-function-based approach employs cost-sensitive learning, such as reweighting sample-wise loss by inverse class frequency and by assigning a relatively higher loss \cite{Lin2017,Wang2020}.  
In addition, Chen et al. represented the {\it complement cross-entropy} (CCE) \cite{Chen2019,Kim2020}, as illustrated in Figure \ref{fig:CompleCE}.  
The CCE evenly suppresses softmax probabilities on incorrect classes during training and demonstrates efficacy. However, it has a limitation in that it induces a complement training time that is approximately two times longer because of twice the backpropagation per iteration.   
 
For medical image classification, Kieu et al. conducted a survey on deep learning for lung disease 
detection\cite{Kieu2020}. They discussed data imbalance as an issue and the future direction of lung disease detection using deep learning. When training the image classification, if the number of samples of one class is significantly higher than that of the other class, the resulting model will be biased. Thus, it is preferable to have a balanced number of images in each class. However, lung disease is a relatively rare event and not a balanced case. In the initial phase of insufficient infection vision mining, when learning a multiclass classification of COVID-19, pneumonia, and normal lungs, the number of pneumonia images far exceeded the number of unseen infection images. After many pandemic experiences and international collaborations with medical doctors, the COVID-19 radiography database \cite{Kaggle2021} has been updated so that 3616 positive images of COVID-19 in chest X-rays are publicly available \cite{Khan2022}. 
In this study, we present ablation studies on imbalanced positive ratios in a lung infection dataset that includes several small ratios of COVID-19 anomalies and normal images.    
 
\begin{table*}[h] 
\centering 
\caption{Damage learning methods and clear examples under imbalanced settings.} 
\label{tab:imbset} 
\begin{tabular}{|c|c|c|c|} 
\hline 
\textbf{Imbalance} & \textbf{Typical data} & \textbf{Learning method} & \textbf{Clear examples} \\ 
\hline 
Missing & Continuous Target, Label & Regression & Distribution Smoothing\cite{Stocksieker2023,Yang2021} \\ 
Class & Majority-Minority & Image Classification & Resample\cite{Kim2019,Koziarski2020}, Reweight \cite{Lin2017,Wang2020}, Complement\cite{Chen2019} \\ 
Object & Bounding Box & Object Detection & Handling scale \& objective imbalance\cite{Agarwal2018,Oksuzy2020,Zou2018} \\ 
Spatial & Foreground-Background & Object Detection & Handling spatial imbalance\cite{Agarwal2018,Oksuzy2020} \\ 
Pixel-wise & Long-tailed classes & Semantic Segmentation & Center Collapse, Equiangular Tight Frame \cite{Li2022,Zhong2023} \\ 
\hline 
\end{tabular} 
\end{table*} 
 
\subsubsection{Imbalanced Object Detection} 
Since 2015, comprehensive object-detection surveys \cite{Agarwal2018,Zou2018} have been conducted on methods for handling scale and spatial imbalances. Surveys have also been conducted on industrial-domain-specific object detection, such as vehicle detection\cite{Sun2006}, pedestrian detection\cite{Dollar2012}, and face detection\cite{Zafeiriou2015,Bai2017}. Sun et al. \cite{Sun2006} and Dollar et al. \cite{Dollar2012} covered these methods from an imbalance perspective because they presented a comprehensive analysis of feature extraction methods that handle scale imbalance. In addition, Litjens et al. \cite{Litjens2017} discussed deep learning applications to medical image analysis that presented challenges with their possible solutions, including a limited exploration of the class imbalance problem. 
 
\begin{figure} 
\centering 
\includegraphics[width=0.43\textwidth] 
{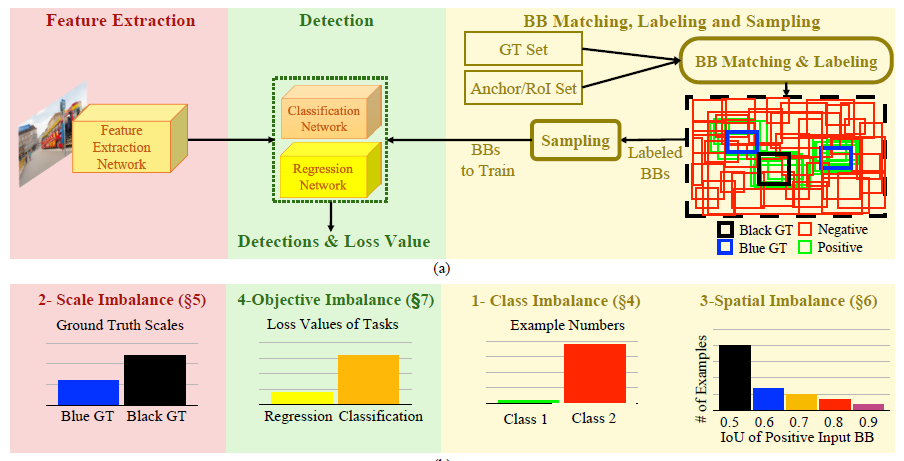} 
\caption{Imbalanced object detection: illustration of 4 imbalance categories (Oksuzy et al. 2020).} 
\label{fig:Imb4Catego} 
\end{figure}

Oksuzy et al. systematically reviewed the object detection literature and identified eight imbalance problems \cite{Oksuzy2020}, as illustrated in Figure \ref{fig:Imb4Catego}. 
These are grouped into four main types: {\it class imbalance, scale imbalance, spatial imbalance}, and {\it objective imbalance}. First, a class imbalance occurs when there is significant inequality among the number of samples pertaining to each class. Class imbalance is considered the foreground-to-background relationship and imbalance among the foreground classes. Second, a scale imbalance occurs when objects have various scales and numbers of samples pertaining to different scales. Third, spatial imbalance refers to a set of factors related to the spatial properties of the bounding boxes, such as regression penalty, location, and IoU. Fourth, an objective imbalance occurs when there are multiple loss functions to minimize, e.g., classification and regression losses \cite{Oksuzy2020}. 
 
\subsubsection{Imbalanced Semantic Segmentation} 
Previous studies on 2D \& 3D semantic segmentation have mainly focused on network architecture and module design, ignoring the impact of data distribution. Because semantic segmentation datasets naturally follow a heavily imbalanced distribution among classes, neural networks perform poorly when trained on them \cite{Kang2020,Menon2020,Ren2020,Zhong2021}. 
Some studies have attempted to induce a neural collapse in imbalanced learning to improve the accuracy of minor classes \cite{Thrampoulidis2022,Xie2022,Yang2022}. However, these studies on neural collapse are limited in recognition. As discovered by Papyan et al. \cite{Papyan2020}, the {\it neural collapse} phenomenon in that the within-class means of features and classifier weight vectors converge to the vertices of a {\it simplex equiangular tight frame} (simplex ETF) at the end of the classification training.   
 
\begin{figure} 
\centering 
\includegraphics[width=0.43\textwidth] 
{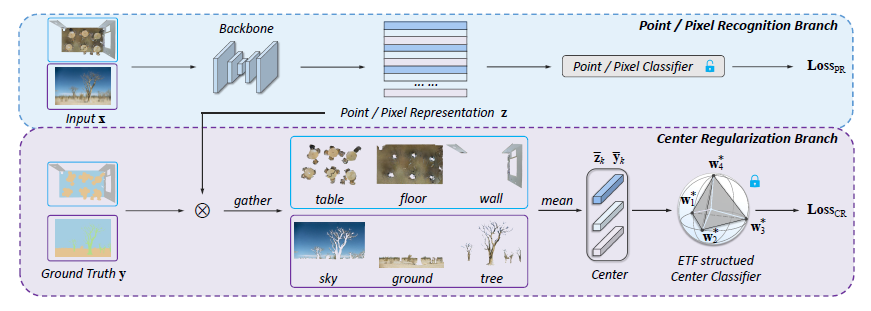} 
\caption{Imbalanced semantic segmentation: e.g., Center collapse regularizer using an equiangular tight frame (Zhong et al. 2023).} 
\label{fig:SimplexETF} 
\end{figure} 
 
The simplex ETF structure in neural collapse renders feature centres equiangular separation and maximal discriminating ability, which can effectively improve the performance of minor classes in long-tailed recognition \cite{Li2022,Zhu2022}. To understand the imbalanced semantic segmentation, the structures of feature centres and classifiers were explored by Zhong et al. \cite{Zhong2023}, as illustrated in Figure \ref{fig:SimplexETF}.  
Surprisingly, symmetric equiangular separation, as instructed by the neural collapse phenomenon in image recognition, did not hold in semantic segmentation for either feature centres or classifiers. To exploit the equiangular and maximum separation properties for better performance on minor classes, the Center Collapse (CeCo) regularizer was presented for imbalanced semantic segmentation problems. 
 
\subsection{Anomaly Detection for Damage Vision} 
Naturally, an anomaly detection algorithm always encounters more or less imbalanced data properties. Because the damaged phenomena rarely occur, such as damage, deterioration, deficit, disease, infection, accident, and devastation. This paper highlights damage vision and feature-imbalanced properties.  
There are many anomaly detection techniques for damage vision data.  
Over the past decade, anomaly detection techniques have attracted significant attention in the widespread domain of applications, assisted by machine learning and deep learning methodologies. Previous survey papers have provided fruitful systematic overviews \cite{Chalapathy2019,Chandola2009,Ruff2020,Yuan2022} focusing on the model property, application domain, and trustworthiness to be more interpretable, fair, robust, and privacy settings.  
Specifically, vision-based deep learning applications have emerged from two driving forces: computing accessibility and a digitalised society that accelerates the creation of many datasets annotated with several class labels. Over 20 datasets of surface damage to industrial products have focused on various materials, such as steel, metals, aluminium, tiles, fabrics, printed boards, solar panels, concrete, roads, pavements, bridges, and rails \cite{Saberironaghi2023}.  
 
Modern anomaly detection approaches can be divided into three primary categories: one-class classification, patch-wise embedding similarity, and reconstruction-based models \cite{Yasuno2023}. Inspired by \cite{Ruff2020}, these anomaly detection approaches were reviewed in a unified manner, progressing from less to more complexity scale through several categories of localisation models. Anomaly detection approaches based on a less complexity scale include the one-class support vector machine (OC-SVM) \cite{Chalapathy2018}, support vector data description (SVDD) \cite{Tax2014}, principal component analysis (PCA) \cite{Hawkins1974}, and kernel-PCA \cite{Hoffmann2007}. Anomaly detection approaches based on more complexity scales include deep SVDD \cite{Ruff2018}, fully convolutional data description (FCDD) \cite{Liznerski2021}, variational autoencoder (VAE) \cite{An201  
 5,Kingma2019}, and adversarial autoencoders (AAE) \cite{Zhou2017}. However, reconstruction-based models cannot always reconstruct synthetic outputs accurately based on their susceptibility to background noise. By contrast, one-class classification models depend neither on synthetic reconstruction nor on probabilistic assumptions. This study highlighted one-class classification models for damage vision anomaly detection.  
 
In addition, the patch-wise embedding approach minimises the background noise per patch image. To localise anomalous features in a patch image, patch-wise embedding similarity models show that the normal reference can be a sphere feature containing embeddings from normal images. In this case, the anomaly score is the distance between the embedding vectors of a test image and the reference vectors representing the normality from the dataset. Embedding-similarity-based models include SPADE \cite{Cohen2020}, PaDiM \cite{Thomas2020}, and PatchCore \cite{Roth2021}. However, these models are based on supervised learning, which additionally requires optimisation algorithms such as greedy coreset selection and a nearest neighbour search on a set of normal embedding vectors; thus, the inference complexity scales linearly to the size of the training dataset. In contrast, the one-class classification approach can learn efficiently using a rare class of imbalanced dataset with fewer scales. The author presented a one-class classification-based anomaly detection method for imbalanced damage vision and compared the accuracy between our method and patch-wise embedding similarity models, typically PaDiM and PatchCore. 
 
\subsection{Few-shot Damage Vision Feedback for Class Imbalance} 
For condition-based maintenance, automating visual inspection is crucial to ensure high quality. Deterioration prognostic attempts to optimise the fine decision process for predictive maintenance and proactive repair. In civil infrastructures and living environments, damage data mining cannot avoid imbalanced data issues because of rare unseen events and the high-quality status of improved operations. 
In Table~\ref{tab:imbset}, we highlight the imbalanced settings and clear examples. From the aforementioned related studies, we summarise that imbalanced data problems can be categorised into four types: 1) missing range of target and label variables, 2) majority-minority class imbalance, 3) foreground background of spatial imbalance, and 4) long-tailed class of pixel-wise imbalance. Since 2015, many studies on imbalanced data have been conducted based on deep-learning algorithms using regression, image classification, object detection, and semantic segmentation. However, an anomaly detection approach for imbalanced data is not yet sufficiently known.   
 
Previous imbalanced studies were based on {\it non-damage vision} datasets. No damaged region of interest has been included in the existing research scope using learning methods such as regression, image classification, object detection, and semantic segmentation.  
In the present imbalanced studies, illustrated in Figure \ref{fig:DamageVisionMining}, we provide key results on the significance of damage vision mining, hypothesising that the more effective the ratio of the anomalous class, the higher the accuracy gain of anomaly detection applications. We hypothesised that the damage-vision mining process is changed into significant two phases. The damage-vision mining phases contain 1) {\it damage-vision mining opportunity}, the former phase with a higher accuracy gain, and 2) {\it over-mining}, the latter phase without a further gain in accuracy. The former phase of damage vision mining opportunity is significantly beneficial because of its higher performance than unsupervised learning without anomalous data mining, as well as the promising advantage of accuracy by damage vision mining.  
 
In this study, we focused on the class-imbalanced anomaly-detection problem using typical 12 damage vision datasets. The anomaly detection problem is essential for two imbalanced classes between small anomalies and large normal images. We consider a {\it feature imbalance}, in which minor clusters were distributed in the feature embedding space. The authors analysed the imbalanced distribution of the damage vision embedding feature space and implemented our contrastive damage representation learning using our MN-pair contrastive learning and density-based clustering \cite{Yasuno2023MNPair}. Surprisingly, the number of damage feature clusters was greater than 10, rather than the number of initial classes that contained normal and anomalies. Specifically, the damage feature class was clustered into a narrow region on the embedding space.           
Furthermore, we intend to find the effective positive ratio of anomalies versus a relatively large normal class without over-mining to achieve stable accuracy and avoid wasting time and resources on damage vision mining.  
 
\section{Anomaly Detection and Embedding Feature} 
\subsection{One-class Classification Using Deeper FCDDs} 
The authors have already presented deeper FCDDs\cite{Yasuno2023} and applied them to inspection datasets of bridges, dams, and buildings. However, an anomaly detection approach for imbalanced data is not yet sufficiently known. To apply imbalanced vision datasets, we summarise a classification-based anomaly-detection method using deeper Fully-Convolutional Data Descriptions (deeper FCDDs\cite{Yasuno2023}).  
 
Let $I_k$ be the $k$-th image in an imbalanced vision dataset of size $h\times w$. 
We consider the number of training images and weight $W$ of the fully convolutional network (FCN).  
Let $\Phi^b_W(I_k)$ denote the mapping of the deeper CNN to backbone $b$ based on the input image. The one-class classification model was formulated using the cross-entropy loss function as follows: 
\begin{equation} 
\begin{split} 
\mathcal{L}_{DeepSVDD} =& - \frac{1}{n} \sum_{k=1}^{n} (1-a_k) \log \ell (\Phi^b_W(I_k)) \\ 
                    &+ a_k \log [ 1 -  \ell (\Phi^b_W(I_k)) ], 
\end{split} 
\end{equation} 
where $a_k =1$ denotes the anomalous label of the $k$-th damage vision and $a_k =0$ denotes the normal label of the $k$-th non-damage vision. A pseudo-Huber loss function was introduced to obtain a more robust loss formulation \cite{Ruff2021icml} in Equation (2). Let $\ell(u)$ be the loss function and define the pseudo-Huber loss as follows:    
\begin{equation} 
\ell(u) = \exp(-H(u)),~ H(u) = \sqrt{\|u\|^2 + 1} -1. 
\end{equation} 
Then, a deeper FCDD loss function can be formulated as 
\begin{equation} 
\begin{split} 
&\mathcal{L}_{deeperFCDD} = \frac{1}{n} \sum_{k=1}^{n} \frac{(1-a_k)}{uv} \sum_{x,y} H_{x,y} (\Phi^b_W(I_k)) \\  
                           &- a_k \log \left[ 1 -  \exp\left\{ \frac{-1}{uv} \sum_{x,y} H_{x,y} (\Phi^b_W(I_k)) \right\} \right], \label{eqn:deepFCDD} 
\end{split} 
\end{equation} 
where $H_{x,y}(u)$ are the elements $(x,y)$ of the receptive field of size $u\times v$ under a deeper FCDD.  
In equation (\ref{eqn:deepFCDD}), if we set an unsupervised learning, the positive second term cancels out. If we use imbalanced vision data that include fewer anomalous images and relatively large normal images, a deeper FCDD loss function (\ref{eqn:deepFCDD}) is less influenced by the positive second term.   
The anomaly score $S_k$ of the $k$-th image is expressed as the sum of all elements of the receptive field as follows: 
\begin{equation} 
S_k(b) = \sum_{x,y} H_{x,y} (\Phi^b_W(I_k)),~k=1,\cdots,n. 
\end{equation}   
Herein, we present the construction of a baseline FCDD \cite{Yasuno2023} with an initial backbone $b=0$ and perform CNN27 mapping $\Phi^0_W(I_k)$ from the input image $A_k$ in the imbalanced vision dataset. We also present deeper FCDDs focusing on elaborate backbones $b\in \{$VGG16, ResNet101, Inceptionv3$\}$ with the mapping operation $\Phi^{b}_W(I_k)$ to achieve better performance. In this paper, we presented ablation studies on the anomalous ratio of imbalanced datasets for detecting material deterioration and disaster damage.  
 
\subsection{Damage-mark Heatmap Upsampling} 
Convolutional neural network (CNN) architectures comprising millions of common parameters have exhibited remarkable performance; however, the underlying reasons for this superiority remain unclear. Heatmap visualisation techniques for detecting and localising anomalous features are typically categorised as masked sampling and activation-map approaches.  
The former includes methods, such as occlusion sensitivity \cite{Zeiler2013} and local interpretable model-agnostic explanations \cite{Ribeiro2016}.  
The latter category includes activation maps such as class activation maps (CAMs) \cite{Zhou2015} and gradient-based extensions (Grad-CAM) \cite{Selvaraju2017}.  
Nonetheless, the aforementioned disadvantages include the requirement for parallel computation resources and iterative computation time for local partitioning, masked sampling, and generating a gradient-based heatmap. 
 
In this study, we adopt a receptive field upsampling approach \cite{Liznerski2021} to visualise anomalous features using an upsampling-based activation map with Gaussian upsampling from the receptive field of the FCN. The primary advantages of the upsampling approach are the reduced computational resource requirements and shorter computation times. The proposed upsampling algorithm generates a full-resolution anomaly heatmap from the input of a low-resolution receptive field $u\times v$. 
  
Let $H\in {R}^{u\times v}$ be a low-resolution receptive field (input) and let $H'\in {R}^{h\times w}$ be a full-resolution damage-mark heatmap (output). 
We define the 2D Gaussian distribution $G_2(m_1,m_2,\sigma)$ as follows:  
\begin{equation} 
\begin{split} 
&[G_2(m_1,m_2,\sigma)]_{x,y} \equiv \\ 
 &\frac{1}{2\pi\sigma^2}\exp\left(-\frac{(x-m_1)^2+(y-m_2)^2}{2\sigma^2}\right).   
\end{split} 
\end{equation} 
The Gaussian upsampling algorithm from the receptive field was implemented as follows: 
\begin{enumerate} 
\item $H' \leftarrow 0 \in {R}^{h\times w}$ 
\item for all output pixels $d$ in $H \leftarrow 0 \in {R}^{u\times v}$ 
\item \qquad $u(d) \leftarrow$ is upsampled from a receptive field of $d$ 
\item \qquad $(c_1(u),c_2(u)) \leftarrow$ is the center of the field $u(d)$ 
\item \qquad $H' \leftarrow H' + d\cdot G_2(c_1,c_2,\sigma)$ 
\item end for 
\item return $H'$  
\end{enumerate} 
After conducting experiments with various datasets, we determined that a receptive field size of $28 \times 28$ is a practical value. When generating a deterioration heatmap reveals a damage mark, we must unify the display range corresponding to the anomaly scores, which range from the minimum to maximum value. To strengthen the defective regions and highlight hazard marks, we define a display range of [min, max/4], where the quartile parameter is 0.25. This resulted in a histogram of the anomaly scores with a long-tailed shape. If we were to include the complete anomaly score range, the colours would weaken to blue or yellow on the maximum side. 
 
\subsection{MN-pair Contrastive Damage Representation} 
The authors \cite{Yasuno2023MNPair} formulated an MN-pair contrastive damage representation and clustering method and found its applicability to a few datasets of steel and concrete. However, another damage representation of imbalanced vision is not yet sufficiently understood. To analyse the feature imbalance, we summarised the MN-pair damage representation method for contrastive learning embedding feature similarity and for density-based clustering of the normal and anomalies on the imbalanced vision datasets.  
 
Let $x$ denote the set of input images on the damage vision dataset. Let $e_i=f(x_i;\vartheta) \in R^L$ be the $i$–th damage embedding of input $i\in \{1,\cdots,n \}$  which preserves the semantic aspects of the damage. Here, $n$ is the number of input images. Furthermore, $\vartheta$ is a shared parameter under a CNN for damage metric learning, and $L$ is the dimension of the damage embedding space.  
Let $F_i=e_i/||e_i||_2$ be the $\ell_2$–normalized version. The damage similarity can be measured from the distance between two images $i_1$ and $i_2$ using the normalised cosine similarity: 
\begin{equation} 
s_{\vartheta}(x_{i_1},x_{i_2}) = (F_{i_1})^T F_{i_2} 
\end{equation} 
where larger values indicate greater similarities. Here, the suffix $T$ denotes the transposed operation.  
The N-pair loss approach \cite{Sohn2016} creates a multi-class classification, in which we create a set of $N-1$ negative $\{x_k^{-} \}_{k=1}^{N-1}$ and one positive $x_j^{+}$ for every anchor image $x_i$. We defined the following N-pair loss function for each set: 
\begin{equation} 
\begin{split} 
&\mathcal{L}_{N-pair}\left( \vartheta;x_i,x_j^{+},\{x_k^{-} \}_{k=1}^{N-1} \right) = \\  
                           & \log \left[ 1 -  \exp\left( s_{\vartheta}(x_i,x_j^{+}) \right) + \sum_{k=1}^{N-1} \exp\left( s_{\vartheta}(x_i,x_k^{-}) \right) \right], \label{eqn:Npair} 
\end{split} 
\end{equation} 
For a simple expression, we denote the cosine similarities as 
\begin{equation} 
s_{i,j}^{0,+} = s_{\vartheta}(x_{i},x_{j}^{+})/\tau,~s_{i,k}^{0,-} = s_{\vartheta}(x_{i},x_{k}^{-})/\tau 
\end{equation} 
Here, the suffix $0$ denotes the anchor, which is divided by the normalised temperature scale $\tau$. 
This scale enhances small values to ensure that the N-pair loss can be trained efficiently; for example, we can set the scale $\tau=0.3$ in the present study. Thus, the N-pair loss in Equation (\ref{eqn:Npair}) can be expressed as follows:  
\begin{equation} 
\begin{split} 
(7) = -\log \frac{ \exp\left( s_{i,j}^{0,+} \right) } { \exp\left( s_{i,j}^{0,+} \right) + \sum_{k=1}^{N-1} \exp\left( s_{i,k}^{0,-} \right) } \label{eqn:simNpair} 
\end{split} 
\end{equation} 
The N-pair loss was identical to the InfoNCE loss \cite{Oord2018, Khosla2020}. However, N-pair loss is a slow starter because of the presence of only one positive image toward N-1 negative images. A positive signal is important to bond the inner embedding space around the same class of each anchor.  
Thus, we propose an MN-pair weighting loss \cite{Yasuno2023MNPair} instead of (\ref{eqn:Npair})-(\ref{eqn:simNpair}), in which we create a set of $N-1$ negative $\{x_k^{-} \}_{k=1}^{N-1}$ and $M-1$ positive $\{x_j^{+} \}_{j=1}^{M-1}$ for every anchor $x_i$: 
\begin{equation} 
\begin{split} 
&\mathcal{L}_{MN-pair}\left( \vartheta;x_i, \{x_j^{+}\}_{j=1}^{M-1}, \{x_k^{-} \}_{k=1}^{N-1} \right) = \\  
& -\log \frac{ \pi\sum_{j=1}^{M-1} \exp\left( s_{i,j}^{0,+} \right) }  
{ \pi\sum_{j=1}^{M-1} \exp\left( s_{i,j}^{0,+} \right) + (1-\pi)\sum_{k=1}^{N-1} \exp\left( s_{i,k}^{0,-} \right) }, \label{eqn:MNpair} 
\end{split} 
\end{equation} 
where $\pi$ is a positive weight and $1-\pi$ is a negative weight constrained by the sum of both weights being one; for example, $\pi=0.15$ was applicable in the present study. To train the parameters $\vartheta$ under a CNN for damage-metric learning, we can minimise the MN-pair loss function $\mathcal{L}_{MN-pair}$ using a standard optimiser, such as Adam. 
 
\subsection{Density-Based Damage Feature Clustering} 
Using the damage-embedded feature with the dimension, we can reduce its dimension into two axes of scores using the t-SNE algorithm \cite{Lindermany2019}. Several different concepts exist for clusters of damage representation, including, 1) well-separated clusters, 2) center-based clusters within a specified radius, and 3) density-based clusters. In a two-dimensional damage-embedding space, we can use either a centre- or density-based approach. The former is based on the distance from neighbouring points to the centre, such as K-means \cite{Hartigan1979} and k-nearest neighbour \cite{Comak2008}. To provide an effective approach for representing a heterogeneous subdivided region of damage beyond predefined classes, we used a density-based clustering algorithm (DBSCAN) \cite{Evangelos1996}. The points of the embedded damage features were classified into 1) core points in the interior of a dense region, 2) border points on the edge of a dense region, and 3) noise points in a sparsely occupied region (a noise or background). The DBSCAN algorithm is formally expressed as     
 
\begin{enumerate} 
\item Label all points as core, border, or noise points. 
\item Eliminate noise points. 
\item Put an edge between all core points that are within a user-specified distance parameter of each other. 
\item Make each group of connected core points into a separate cluster. 
\item Assign each border point to one of the clusters of its associated core points. 
\end{enumerate} 
 
As we used a density-based definition of a cluster, it was relatively resistant to noise and could handle clusters with arbitrarily damaged shapes. Therefore, DBSCAN can determine numerous damage clusters that cannot be identified using a centre-based algorithm such as k-means. For damage representation clustering, we can set a distance parameter; for example, $\varepsilon=3$ was applicable in the present study, and a minimum number of neighbours for core points; for example, 10 was feasible in damage vision studies.  
 
\section{Applied Results} 
\subsection{Class Imbalanced Vision Datasets} 
In this study, we highlighted three imbalanced datasets as shown in Table~\ref{tab:datarail}; the training data contains anomalous and normal images. In this study, we implemented deeper backbone studies using VGG16, ResNet101, Inceptionv3.  
During the training of the anomaly detector, we fixed the input size to $256^2$ training for the wooden sleeper dataset and to $224^2$ training for concrete deterioration and disaster damage. To train the model, we set the mini-batch size to 32 and ran it for 60 epochs.  
In this study, we used the Adam optimiser with a learning rate of 0.0001, gradient decay factor of 0.9, and squared gradient decay factor of 0.99. The training images were partitioned at a ratio of 65:15:20 for the training, calibration, and testing images in each dataset. Here, $N_d$ $(d=1,2,3)$ denotes the number of training data in dataset $d$, and $M_d$ represents the total number of datasets that contain the calibration and testing images. 
\begin{enumerate} 
\item {\it Blood infection $N_1=1300, M_1=2000.$ } 
\begin{itemize} 
\item malaria parasitized in blood smear images by cell-level (\cite{Rajaraman2018}).  
\end{itemize} 
\item {\it Lung infection $N_2=1300, M_2=2000.$ } 
\begin{itemize} 
\item COVID-19 in chest X-ray images (\cite{Kaggle2021,Khan2022}).  
\end{itemize} 
\item {\it Breast cancer $N_3=3120, M_3=4800.$ } 
\begin{itemize} 
\item Breast cancer specimens scanned images that were extracted patches annotated to IDC(Invasive Ductal Carcinoma) negative and positive (\cite{Janowczyk2016,KaggleBreast}).  
\end{itemize} 
\item {\it Driving distraction $N_4=1300, M_4=2000.$ } 
\begin{itemize} 
\item 4-classes: distracted driving images, i.e., texting-left and -right, talking on the phone-left and -right (\cite{Darapaneni2022,Driver2016}).  
\end{itemize} 
\item {\it Wood deterioration $N_5=1300, M_5=2000.$ } 
\begin{itemize} 
\item decayed wooden sleeper in rural railway (\cite{Yasuno2023Imbalanced}).  
\end{itemize} 
\item {\it Concrete damage $N_6=1300=650\times2, M_6=2000.$ } 
\begin{itemize} 
\item 2-classes: crack on pavement and deck (SDNET \cite{Dorafshan2018}).  
\end{itemize} 
\item {\it Logical defects $N_7=1300, M_7=2000.$ } 
\begin{itemize} 
\item 5-classes: logical and structural detects, i.e., breakfast box, juice bottle, pushpins, screw bag, and splicing connectors (MVTec LOCO \cite{Paul2022,MVTecLoco}).  
\end{itemize} 
\item {\it Vegetable damage $N_8=1300=325\times4, M_8=2000.$ }  
\begin{itemize} 
\item old and damaged vegetables in 4-classes: tomato, bell pepper, chili pepper, and new Mexico chili (VegNet \cite{Yogesh2022}).  
\end{itemize} 
\item {\it Plant infection $N_9=1300, M_9=2000.$ }  
\begin{itemize} 
\item infected leaves in 12-classes: apple, blueberry, cherry, corn, grape, peach, potato, raspberry, soybean, strawberry, and tomato (\cite{Sadman2023,Siddiqua2022}).  
\end{itemize} 
\item {\it River sludge $N_{10}=1300=650\times2, M_{10}=2000.$ } 
\begin{itemize} 
\item floating sludge on the river surface images (\cite{Yasuno2022River}).  
\end{itemize} 
\item {\it Disaster damage $N_{11}=1300=325\times4, M_{11}=2000.$ }  
\begin{itemize} 
\item 4-classes: building collapse, flooding area, traffic incidents, fire/smoke (AIDER \cite{AIDER2019}).  
\end{itemize} 
\item {\it Hurricane damage $N_{12}=1300, M_{12}=2000.$ } 
\begin{itemize} 
\item Hurricane satellite imagery (\cite{CaoBuilding2018,CaoData2018}).  
\end{itemize} 
\end{enumerate} 
 
\begin{table}[h] 
\centering 
\caption{Imbalanced training dataset $d$ of target damage.(Each class has $N_d$ images in the normal and anomalous class. At least, the calibration 300 and test 400 images were fixed respectively.)} 
\label{tab:datarail} 
\begin{tabular}{|c|c|c|} 
\hline 
\textbf{Positive ratio} & \textbf{Anomalous} & \textbf{Normal} \\ 
\hline 
1/1(supervised) & $N_d$ & $N_d$ \\ 
1/2 & $N_d/2$ & $N_d$ \\ 
1/4 & $N_d/4$ & $N_d$ \\ 
1/8 & $N_d/8$ & $N_d$ \\ 
1/16 & $N_d/16$ & $N_d$ \\ 
1/32 & $N_d/32$ & $N_d$ \\ 
1/64 & $N_d/64$ & $N_d$ \\ 
1/128 & $N_d/128$ & $N_d$ \\ 
1/$N_d$(one-shot) & 1 & $N_d$ \\ 
\hline 
\end{tabular} 
\end{table} 
 
\subsection{Blood Infection} 
\subsubsection{Backbone Studies of Supervised Detection} 
As shown in Table~\ref{tab:accBackboneBlood}, our deeper FCDD-ResNet101 outperformed in terms of the $F_1$, precision, and recall rather than the baseline CNN27 and other backbone-based deeper FCDDs in this blood smear images dataset for detecting malaria-parasitized cells. 
\begin{table}[h] 
\caption{Backbone ablation studies on malaria-parasitized cell detection using our proposed deeper FCDDs.} 
\label{tab:accBackboneBlood} 
\centering 
\begin{tabular}{|c|c|c|c|c|} 
\hline 
\textbf{Backbone} & \textbf{AUC} & \boldmath{$F_1$} & \textbf{Precision} & \textbf{Recall} \\ 
\hline 
CNN27 & 0.9853 & 0.9468 & 0.9589 & 0.9350 \\ \hline 
VGG16 & 0.9932 & 0.9689 & 0.9629 & 0.9750 \\ 
\textbf{ResNet101} &\textbf{0.9917} & \textbf{0.9774} & \textbf{0.9799} & \textbf{0.9750} \\ 
Inceptionv3 &0.9913 & 0.9759 & 0.9872 & 0.9650 \\ \hline 
\end{tabular} 
\end{table} 
 
\subsubsection{Imbalanced-to-unsupervised Training Results} 
As shown in Table~\ref{tab:accImbalanceBlood}, we conducted ablation studies on the imbalanced data containing smaller anomalous and relatively large normal images. In this study, we applied our deeper FCDD-ResNet101 and achieved high performance in the aforementioned supervised results.   
Compared with the balanced case of a positive ratio of 1/1, we found that there was an applicable range from an imbalanced ratio of 1/2 to 1/16, where the accuracy of recall was consistently less than 4\%.  
However, in the extremely imbalanced range of 1/32 to 1/1300, the accuracy was inferior to the applicable range, that is, the recall was greater than 5\%. The rare positive ratio of 1/32 represents imbalanced data containing few 41 anomalous images and relatively large 1300 normal images. In this case, additional anomalous images should be acquired and added to the initial dataset.    
The marginal gain in accuracy was relatively high by adding the blood infection images of malaria-parasitized cells.  
\begin{table}[h] 
\caption{Imbalanced data studies using our deeper FCDD-ResNet101 for Blood infection detection $N_1=1300$.} 
\label{tab:accImbalanceBlood} 
\centering 
\begin{tabular}{|c|c|c|c|c|} 
\hline 
\textbf{Positive ratio} & \textbf{AUC} & \boldmath{$F_1$} & \textbf{Precision} & \textbf{Recall} \\ 
\hline 
\textbf{1/1(ano.$N_1$)} & \textbf{0.9917} & \textbf{0.9774} & \textbf{0.9799} & \textbf{0.9750} \\ \hline 
1/2(ano.650) & 0.9931 & 0.9735 & 0.9797 & 0.9675 \\  
1/4(ano.325) & 0.9918 & 0.9683 & 0.9820 & 0.9550 \\  
1/8(ano.163) & 0.9893 & 0.9660 & 0.9721 & 0.9600 \\  
1/16(ano.81)& 0.9907 & 0.9636 & 0.9672 & 0.9600 \\ \hline 
\textbf{1/32(ano.41)}&\textbf{0.9919} & \textbf{0.9619} & \textbf{0.9768} & \textbf{0.9475} \\  
\textbf{1/64(ano.20)}&\textbf{0.9866} & \textbf{0.9532} & \textbf{0.9641} & \textbf{0.9425} \\  
\textbf{1/128(ano.10)}&\textbf{0.9839} & \textbf{0.9590} & \textbf{0.9816} & \textbf{0.9375} \\  
\textbf{1/$N_1$(ano.1)} &\textbf{0.9174} & \textbf{0.8239} & \textbf{0.8610} & \textbf{0.7900} \\ \hline 
\end{tabular} 
\end{table} 
 
\begin{figure}[h] 
\centering 
\includegraphics[width=0.4\textwidth]{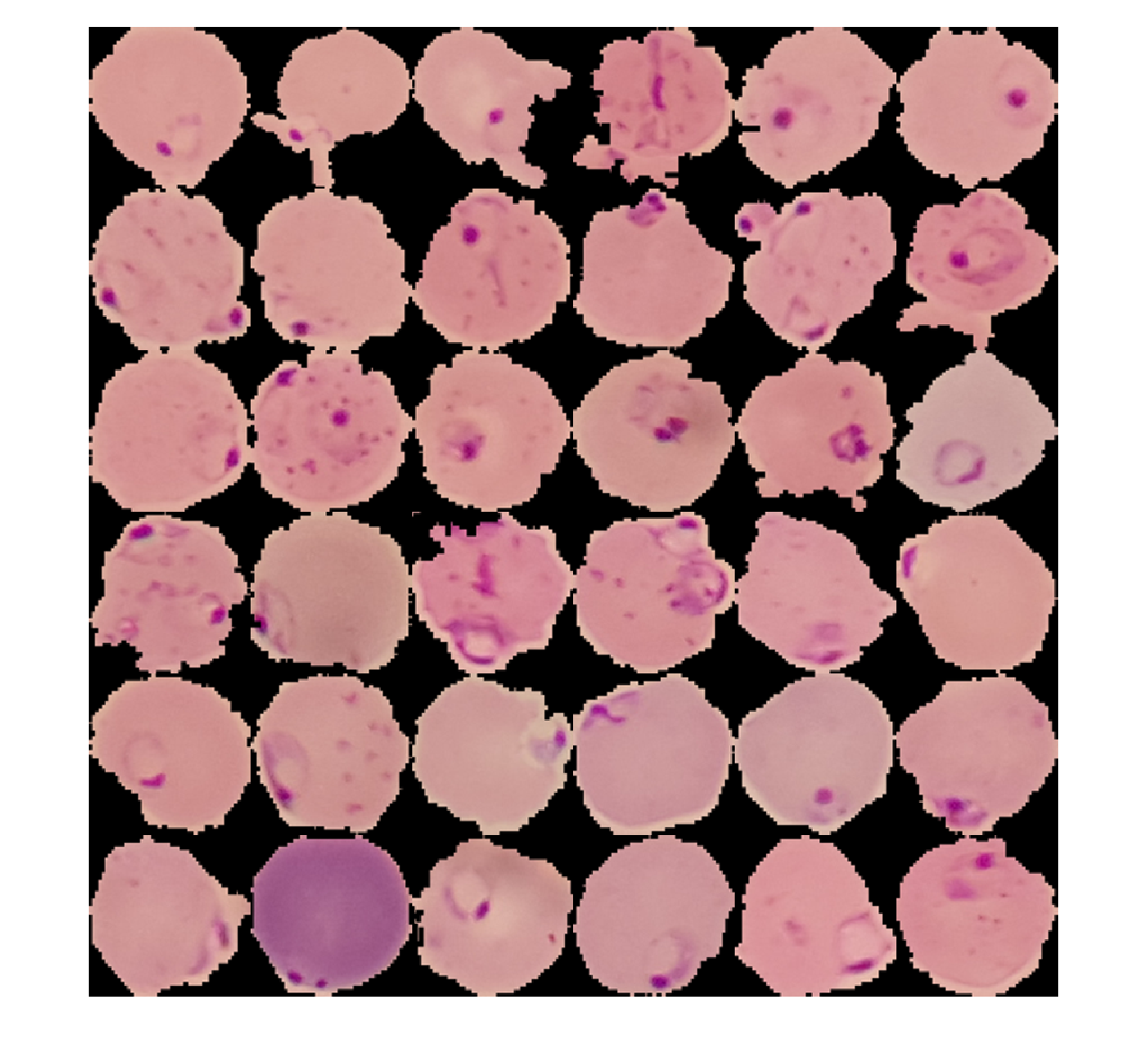} \\ 
\includegraphics[width=0.4\textwidth]{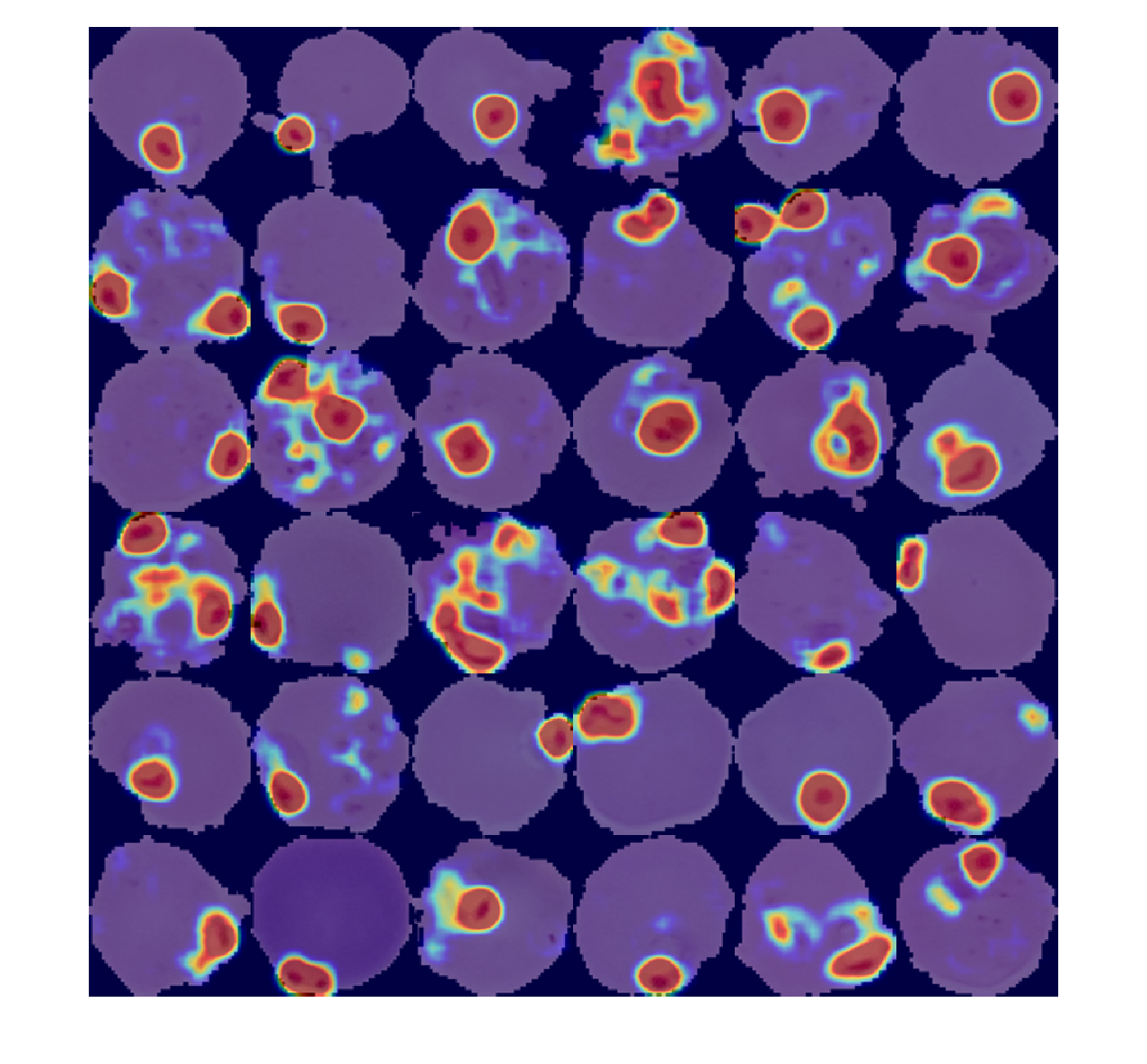} 
\caption{Imbalanced damage images (top) with positive ratio $1/16$, and damage-mark heatmaps (bottom) of blood infection using our deeper FCDD-ResNet101.} 
\label{fig:rawBlood} 
\end{figure} 
\begin{figure}[h] 
\centering 
\includegraphics[width=0.37\textwidth]{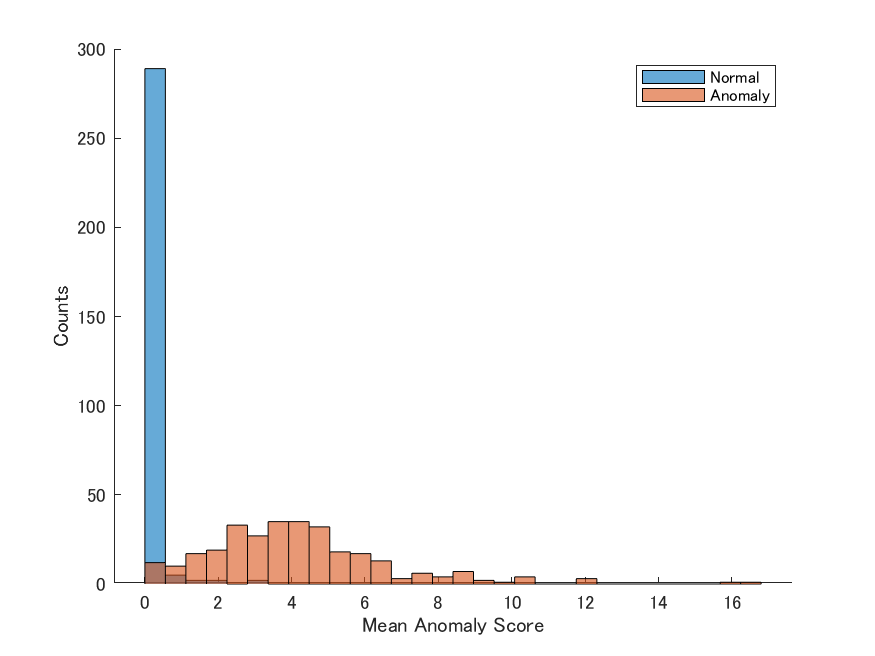} 
\caption{Histogram of blood infection scores using our deeper FCDD-ResNet101 corresponding to the imbalanced damage images with positive ratio $1/16$.} 
\label{fig:histBlood} 
\end{figure} 
 
\begin{figure}[h] 
\centering 
\includegraphics[width=0.37\textwidth]{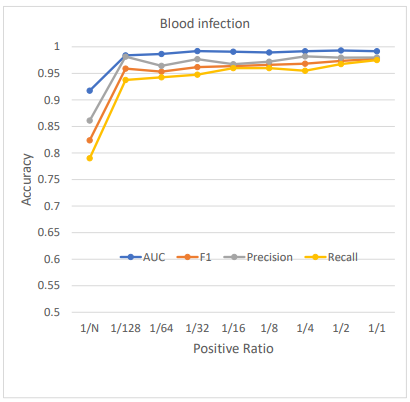} 
\caption{Anomalous vision mining studies on blood infection, indicates that the more anomalies of damage vision, the higher performance of anomaly detection.} 
\label{fig:EffectBlood}  
\end{figure} 
 
\subsubsection{Damage-mark Heatmaps on Blood Infection} 
We visualised the damage features using Gaussian upsampling in our deeper FCDD-ResNet101 network. In addition, we generated a histogram of the anomaly scores of the test images in the imbalanced case with a positive ratio of $1/16$. In Fig. \ref{fig:rawBlood}, a damage-mark explanation is represented. The red region in the heatmap represents the malaria-parasitized cells of features in the blood smear images.  
Fig. \ref{fig:histBlood} illustrates that a few overlapping bins exist in the boundary of the uninfected class and malaria-parasitized class along the horizontal anomaly scores. Thus, to detect malaria-parasitized cells, the score range was well separated in the blood smear image dataset. 
 
\subsubsection{Feedback Effect on Damage Class Mining} 
As shown in Figure \ref{fig:EffectBlood}, from the perspective of all accuracies, the imbalanced studies on blood infection implied that all accuracies consistently converged into significant phases.  
Ranging with a positive ratio of less than $1/8$, we could understand that there were damage vision mining opportunities with an accuracy gain in terms of the AUC. In contrast, ranging with a positive ratio over $1/4$, it shifted in the over-mining phase without any gain in the AUC. The former phase of damage vision mining opportunities has become beneficial owing to the promising advantage of higher accuracy in all of them. 
 
\subsubsection{Embedding Damage Representation} 
\begin{figure}[h] 
\centering 
\includegraphics[width=0.35\textwidth]{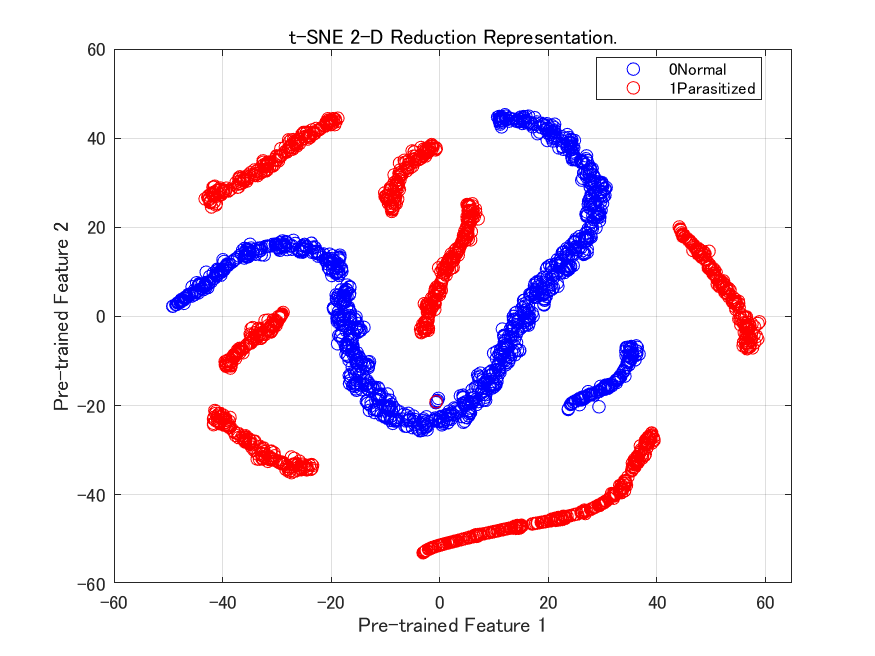} \\ 
\includegraphics[width=0.35\textwidth]{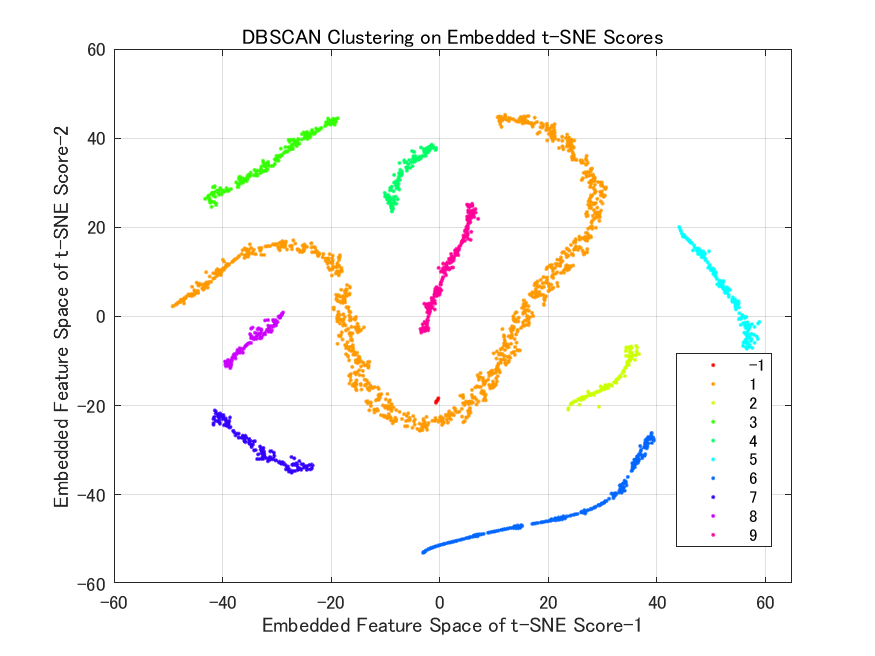} 
\caption{MN-pair contrastive damage representation (top) and density-based clustering (bottom) of blood infection.} 
\label{fig:mndbBlood} 
\end{figure}

As shown in Figure \ref{fig:mndbBlood}, we analysed the feature imbalance in the blood infection embedding space and implemented our MN-pair contrastive damage representation learning and density-based clustering.  
Surprisingly, the number of blood infection feature clusters was nine rather than the initial two classes. The parasitized features of the seven clusters are distributed into narrow regions in the embedding feature space. 
 
\subsection{Lung Infection} 
\subsubsection{Backbone Studies of Supervised Detection} 
As shown in Table~\ref{tab:accBackboneLung}, our deeper FCDD-ResNet101 outperformed in terms of the $F_1$, precision, and recall rather than the baseline CNN27 and other backbone-based deeper FCDDs in this chest X-ray images dataset for detecting lung infection of COVID-19. 
\begin{table}[h] 
\caption{Backbone ablation studies on lung infection detection using our proposed deeper FCDDs.} 
\label{tab:accBackboneLung} 
\centering 
\begin{tabular}{|c|c|c|c|c|} 
\hline 
\textbf{Backbone} & \textbf{AUC} & \boldmath{$F_1$} & \textbf{Precision} & \textbf{Recall} \\ 
\hline 
CNN27 & 0.9359 & 0.8677 & 0.8095 & 0.9350 \\ \hline 
VGG16 & 0.9925 & 0.9662 & 0.9674 & 0.9650 \\ 
\textbf{ResNet101} &\textbf{0.9933} & \textbf{0.9725} & \textbf{0.9701} & \textbf{0.9750} \\ 
Inceptionv3 &0.9918 & 0.9576 & 0.9552 & 0.9600 \\ \hline 
\end{tabular} 
\end{table} 
 
\subsubsection{Imbalanced-to-unsupervised Training Results} 
As shown in Table~\ref{tab:accImbalanceLung}, we implemented ablation studies on the imbalanced data containing smaller anomalies and relatively large normal images. In this study, we applied our deeper FCDD-ResNet101 and achieved high performance in the aforementioned supervised results.   
Compared with the balanced case of a positive ratio of 1/1, we found that there was an applicable range from an imbalanced ratio of 1/2 to a ratio of 1/16, where the accuracy of $F_1$ was consistently greater than 95\%.  
However, in the extremely imbalanced range of 1/32 to 1/1300, the accuracy was inferior to that of the applicable range, for example, $F_1$ exceeded 95\%. The rare positive ratio of 1/32 represents imbalanced data containing quite a little 41 anomalous images and relatively large 1300 normal images. In this case, additional anomalous images should be acquired and added to the initial dataset.    
The marginal gain in accuracy was relatively high when lung infection images of COVID-19 were added.  
\begin{table}[h] 
\caption{Imbalanced data studies using our deeper FCDD-ResNet101 for Lung infection detection $N_2=1300$.} 
\label{tab:accImbalanceLung} 
\centering 
\begin{tabular}{|c|c|c|c|c|} 
\hline 
\textbf{Positive ratio} & \textbf{AUC} & \boldmath{$F_1$} & \textbf{Precision} & \textbf{Recall} \\ 
\hline 
\textbf{1/1(ano.$N_2$)} & \textbf{0.9933} & \textbf{0.9725} & \textbf{0.9701} & \textbf{0.9750} \\ \hline 
1/2(ano.650) & 0.9959 & 0.9764 & 0.9680 & 0.9850 \\  
1/4(ano.325) & 0.9885 & 0.9650 & 0.9650 & 0.9650 \\  
1/8(ano.163) & 0.9911 & 0.9527 & 0.9738 & 0.9325 \\  
1/16(ano.81)& 0.9908 & 0.9533 & 0.9618 & 0.9450 \\ \hline 
\textbf{1/32(ano.41)}&\textbf{0.9911} & \textbf{0.9404} & \textbf{0.9148} & \textbf{0.9675} \\  
\textbf{1/64(ano.20)}&\textbf{0.9864} & \textbf{0.9468} & \textbf{0.9589} & \textbf{0.9350} \\  
\textbf{1/128(ano.10)}&\textbf{0.9717} & \textbf{0.9154} & \textbf{0.9108} & \textbf{0.9200} \\  
\textbf{1/$N_2$(ano.1)} &\textbf{0.7494} & \textbf{0.7088} & \textbf{0.7027} & \textbf{0.7150} \\ \hline 
\end{tabular} 
\end{table} 
 
\begin{figure}[h] 
\centering 
\includegraphics[width=0.4\textwidth]{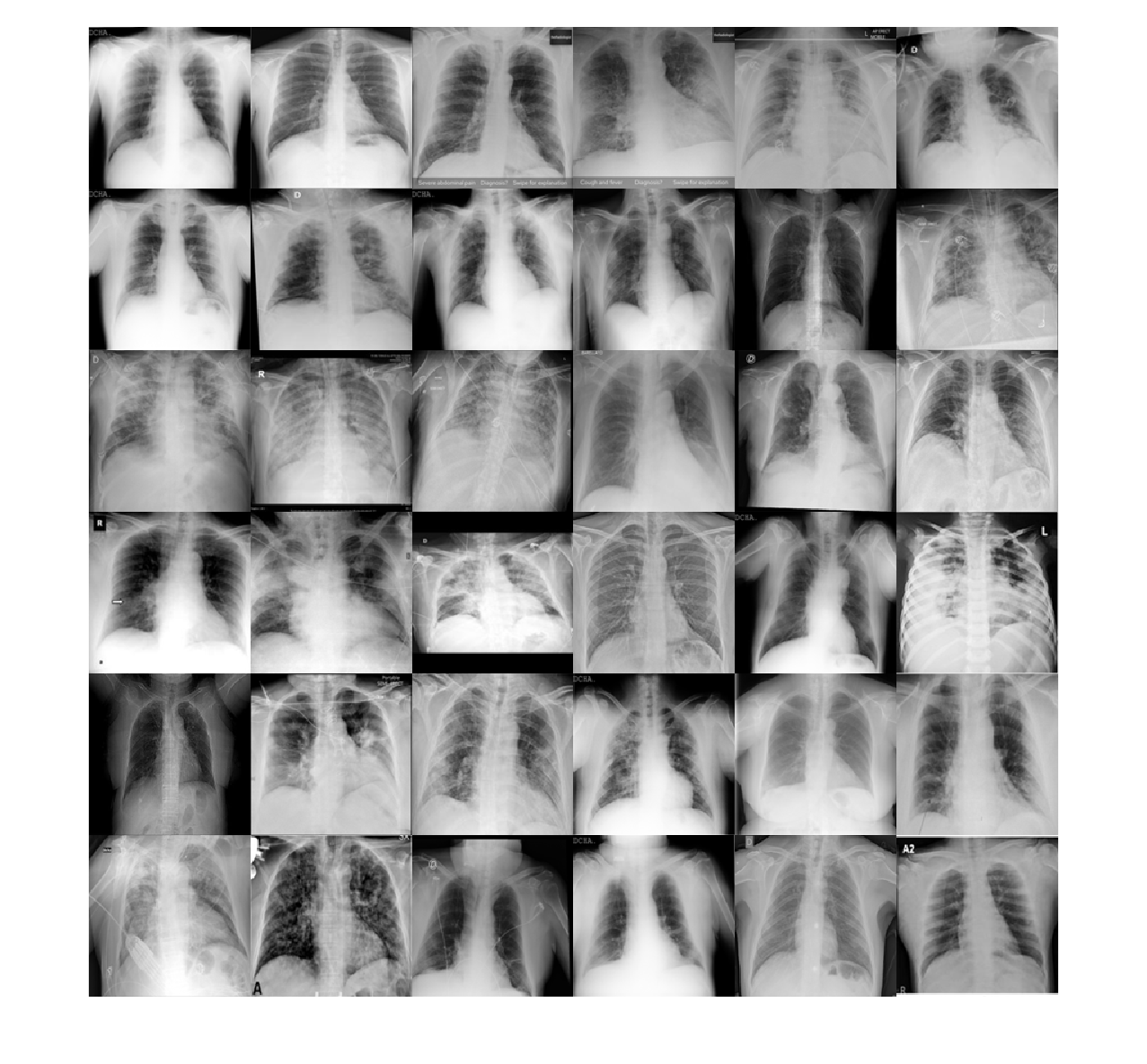} \\ 
\includegraphics[width=0.4\textwidth]{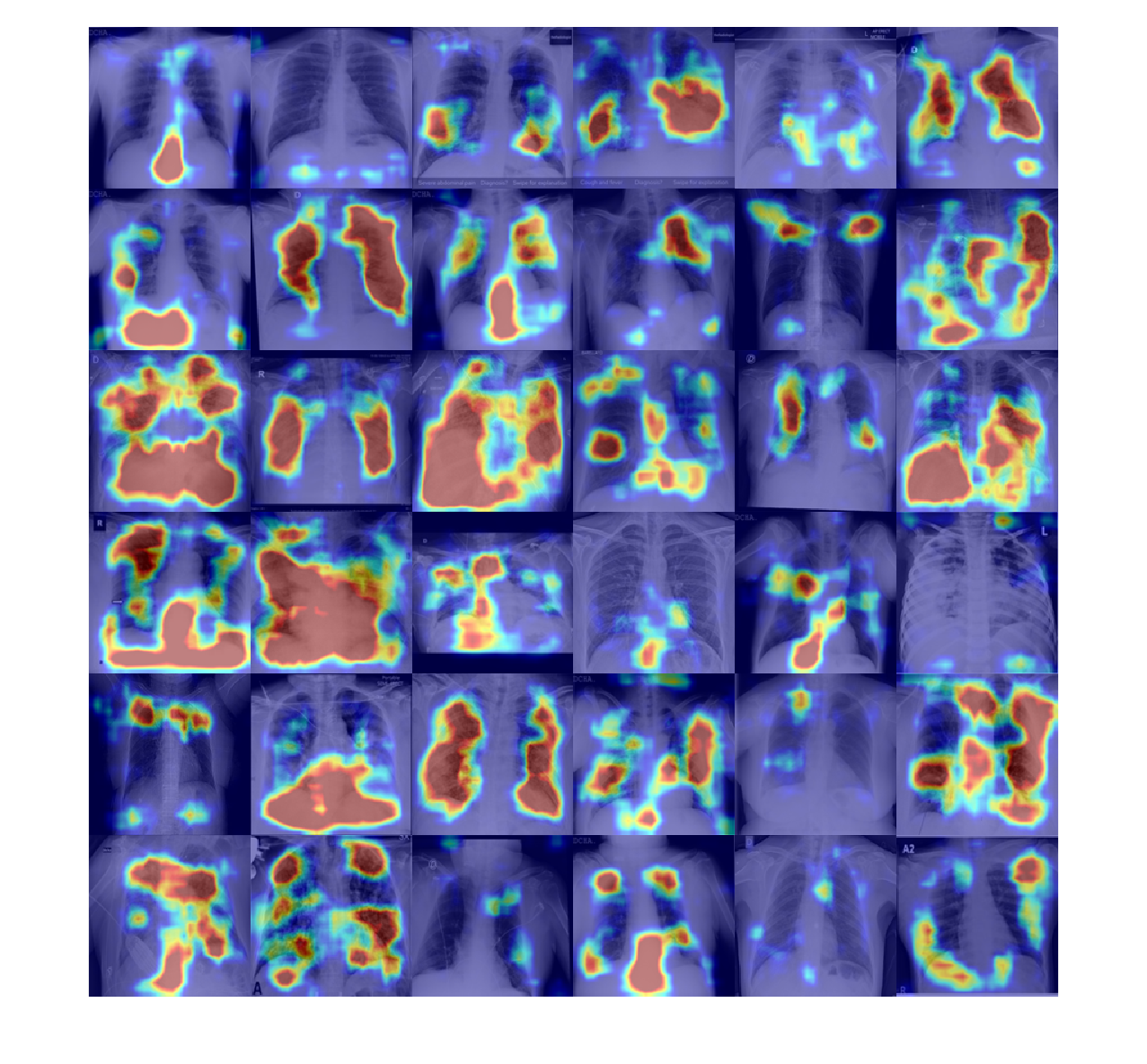} 
\caption{Imbalanced damage images (top) with positive ratio $1/16$, and damage-mark heatmaps (bottom) of lung infection using our deeper FCDD-ResNet101.} 
\label{fig:rawLung} 
\end{figure} 
\begin{figure}[h] 
\centering 
\includegraphics[width=0.37\textwidth]{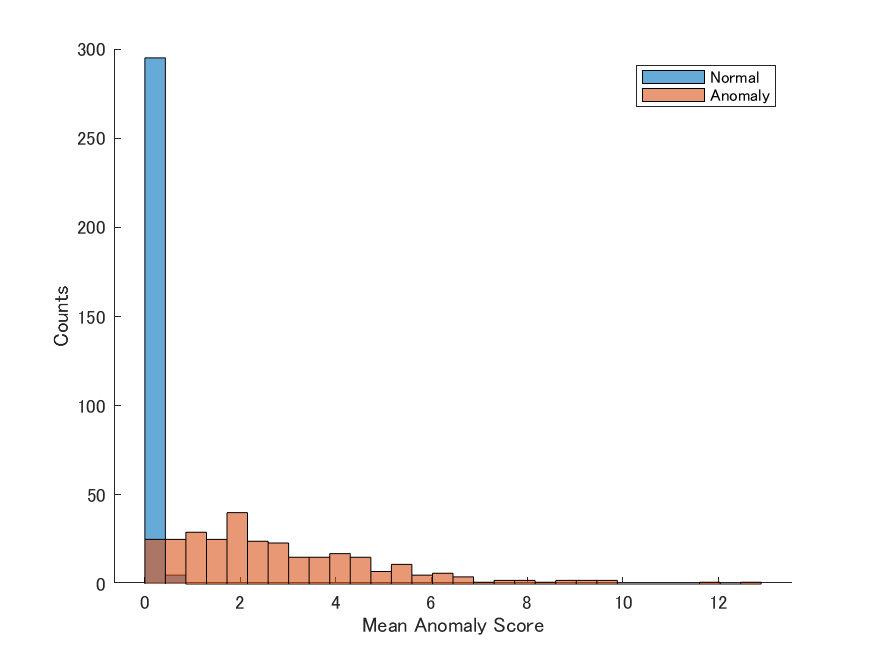} 
\caption{Histogram of lung infection scores using our deeper FCDD-ResNet101 corresponding to the imbalanced damage images with positive ratio $1/16$.} 
\label{fig:histLung} 
\end{figure} 
 
\begin{figure}[h] 
\centering 
\includegraphics[width=0.37\textwidth]{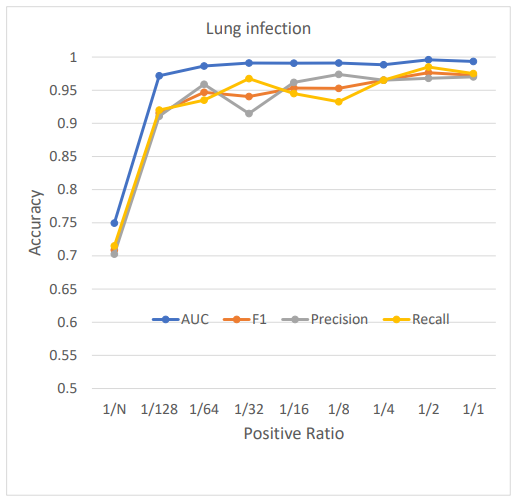} 
\caption{Anomalous vision mining studies on lung infection indicate that the more anomalies of damage vision, the higher the performance of anomaly detection.} 
\label{fig:EffectLung}  
\end{figure} 
 
\subsubsection{Damage-mark Heatmaps on Lung Infection} 
We visualised the damage features using Gaussian upsampling in our deeper FCDD-ResNet101 network. In addition, we generated a histogram of the anomaly scores of the test images in the imbalanced case with a positive ratio of $1/16$. In Fig. \ref{fig:rawLung}, a damage-mark explanation is represented. The red region in the heatmap represents the COVID-19 lung infection of features in the chest X-ray images.  
Fig. \ref{fig:histLung} illustrates that a few overlapping bins exist in the boundary of the normal class and COVID-19 infected class along the horizontal anomaly scores. Thus, to detect lung infection due to COVID-19, the score range was well-separated in the chest X-ray images dataset. 
 
\subsubsection{Feedback Effect on Damage Class Mining} 
As shown in Figure \ref{fig:EffectLung}, from the perspective of all accuracies, the imbalanced studies on lung infection imply that all accuracies consistently converged into significant phases.  
Ranging with a positive ratio of less than $1/8$, we could understand that there were damage vision mining opportunities with an accuracy gain in terms of the AUC. In contrast, ranged with a positive ratio over $1/4$, it shifted in the over-mining phase without any gain in the AUC. The former phase of damage vision mining opportunities has become beneficial owing to its promising advantage of higher accuracy in all of them. 
 
\subsubsection{Embedding Damage Representation} 
\begin{figure}[h] 
\centering 
\includegraphics[width=0.35\textwidth]{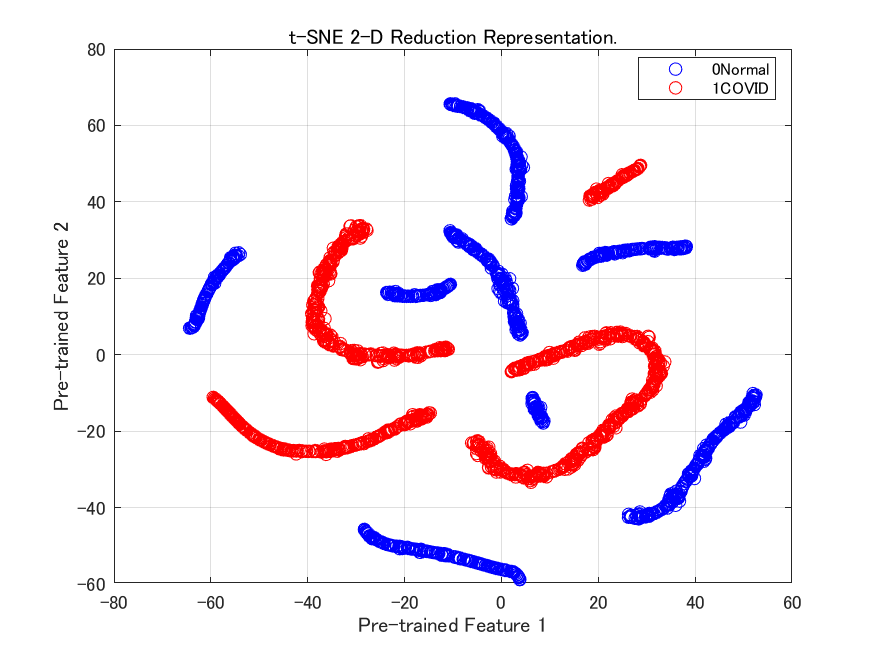} \\ 
\includegraphics[width=0.35\textwidth]{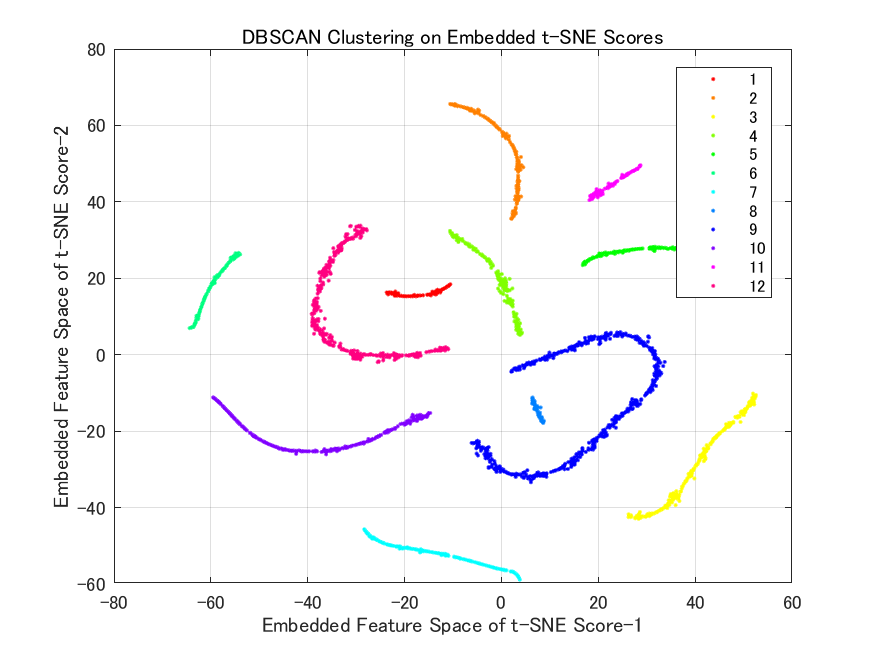} 
\caption{MN-pair contrastive damage representation (top) and density-based clustering (bottom) of lung infection.} 
\label{fig:mndbLung} 
\end{figure}

As shown in Figure \ref{fig:mndbLung}, we analysed the feature imbalance in the lung-infection embedding space and implemented our MN pair contrastive damage representation learning and density-based clustering.  
Surprisingly, the number of lung infection feature clusters was 12, compared to the initial two classes. For the COVID-19 features, the four clusters were densely distributed into longer regions in the embedding feature space. 
 
\subsection{Breast Cancer} 
\subsubsection{Backbone Studies of Supervised Detection} 
As shown in Table~\ref{tab:accBackboneBreast}, our deeper FCDD-ResNet101 outperformed in terms of $F_1$ and precision rather than the baseline CNN27 and other backbone-based deeper FCDDs in this breast cancer specimen scanned image dataset for detecting invasive ductal carcinoma (IDC) positivity. 
\begin{table}[h] 
\caption{Backbone ablation studies on breast cancer IDC detection using our proposed deeper FCDDs.} 
\label{tab:accBackboneBreast} 
\centering 
\begin{tabular}{|c|c|c|c|c|} 
\hline 
\textbf{Backbone} & \textbf{AUC} & \boldmath{$F_1$} & \textbf{Precision} & \textbf{Recall} \\ 
\hline 
CNN27 & 0.9434 & 0.8690 & 0.8504 & 0.8885 \\ \hline 
VGG16 & 0.9638 & 0.8978 & 0.8648 & 0.9333 \\ 
\textbf{ResNet101} &\textbf{0.9656} & \textbf{0.9028} & \textbf{0.8825} & \textbf{0.9239} \\ 
Inceptionv3 &0.9644 & 0.9016 & 0.8739 & 0.9312 \\ \hline 
\end{tabular} 
\end{table} 
 
\subsubsection{Imbalanced-to-unsupervised Training Results} 
As shown in Table~\ref{tab:accImbalanceBreast}, we implemented ablation studies on the imbalanced data containing smaller anomalous and relatively large normal images. In this study, we applied our deeper FCDD-ResNet101 and achieved high performance in the aforementioned results.   
Compared with the balanced case of a positive ratio of 1/1, we found that there was an applicable range from an imbalanced ratio of 1/2 to 1/8, where the accuracy of recall was consistently more than 90\%.  
However, in the extremely imbalanced range of 1/16 to 1/1300, the accuracy was inferior to that of the applicable range, that is, the recall was greater than 90\%. The rare positive ratio of 1/16 represents imbalanced data that contain a small number of anomalous images (81) and relatively large normal images (1300). In this case, additional anomalous images should be acquired and added to the initial dataset.    
The marginal gain in accuracy was relatively high when the breast cancer specimen images were added.  
\begin{table}[h] 
\caption{Imbalanced data studies using our deeper FCDD-ResNet101 for Breast cancer IDC positive detection $N_3=1300$.} 
\label{tab:accImbalanceBreast} 
\centering 
\begin{tabular}{|c|c|c|c|c|} 
\hline 
\textbf{Positive ratio} & \textbf{AUC} & \boldmath{$F_1$} & \textbf{Precision} & \textbf{Recall} \\ 
\hline 
\textbf{1/1(ano.$N_3$)} & \textbf{0.9656} & \textbf{0.9028} & \textbf{0.8825} & \textbf{0.9239} \\ \hline 
1/2(ano.650) & 0.9660 & 0.8946 & 0.9051 & 0.8843 \\  
1/4(ano.325) & 0.9657 & 0.8976 & 0.8783 & 0.9177 \\  
1/8(ano.163) & 0.9622 & 0.8942 & 0.8776 & 0.9114 \\ \hline 
\textbf{1/16(ano.81)}& \textbf{0.9555} & \textbf{0.8832} & \textbf{0.8917} & \textbf{0.8750} \\  
\textbf{1/32(ano.41)}&\textbf{0.9437} & \textbf{0.8729} & \textbf{0.8729} & \textbf{0.8729} \\  
\textbf{1/64(ano.20)}&\textbf{0.9130} & \textbf{0.8368} & \textbf{0.8457} & \textbf{0.8281} \\  
\textbf{1/128(ano.10)}&\textbf{0.8918} & \textbf{0.8182} & \textbf{0.8430} & \textbf{0.7947} \\  
\textbf{1/$N_3$(ano.1)} &\textbf{0.8331} & \textbf{0.7749} & \textbf{0.7532} & \textbf{0.7979} \\ \hline 
\end{tabular} 
\end{table} 
 
\begin{figure}[h] 
\centering 
\includegraphics[width=0.4\textwidth]{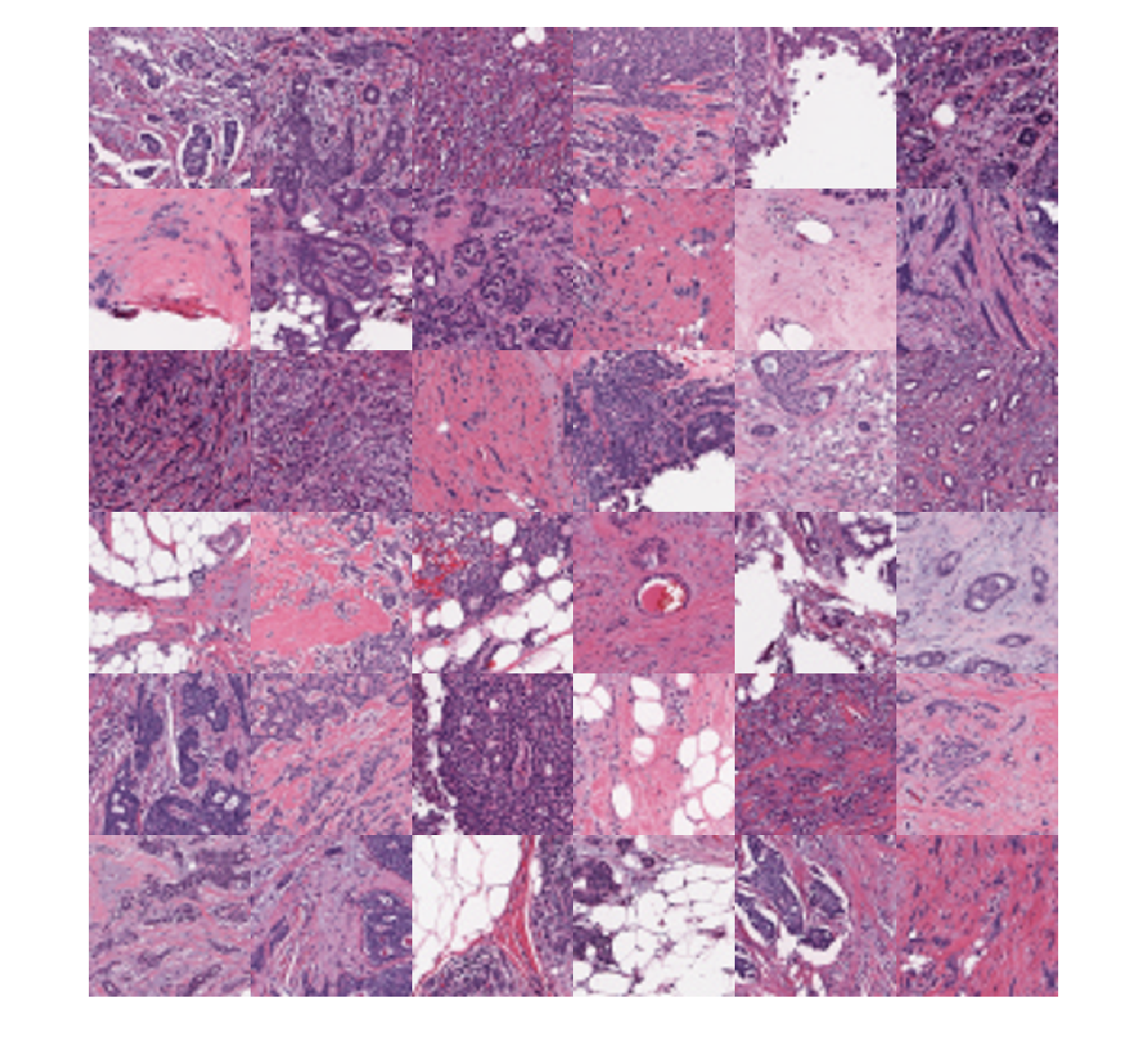} \\ 
\includegraphics[width=0.4\textwidth]{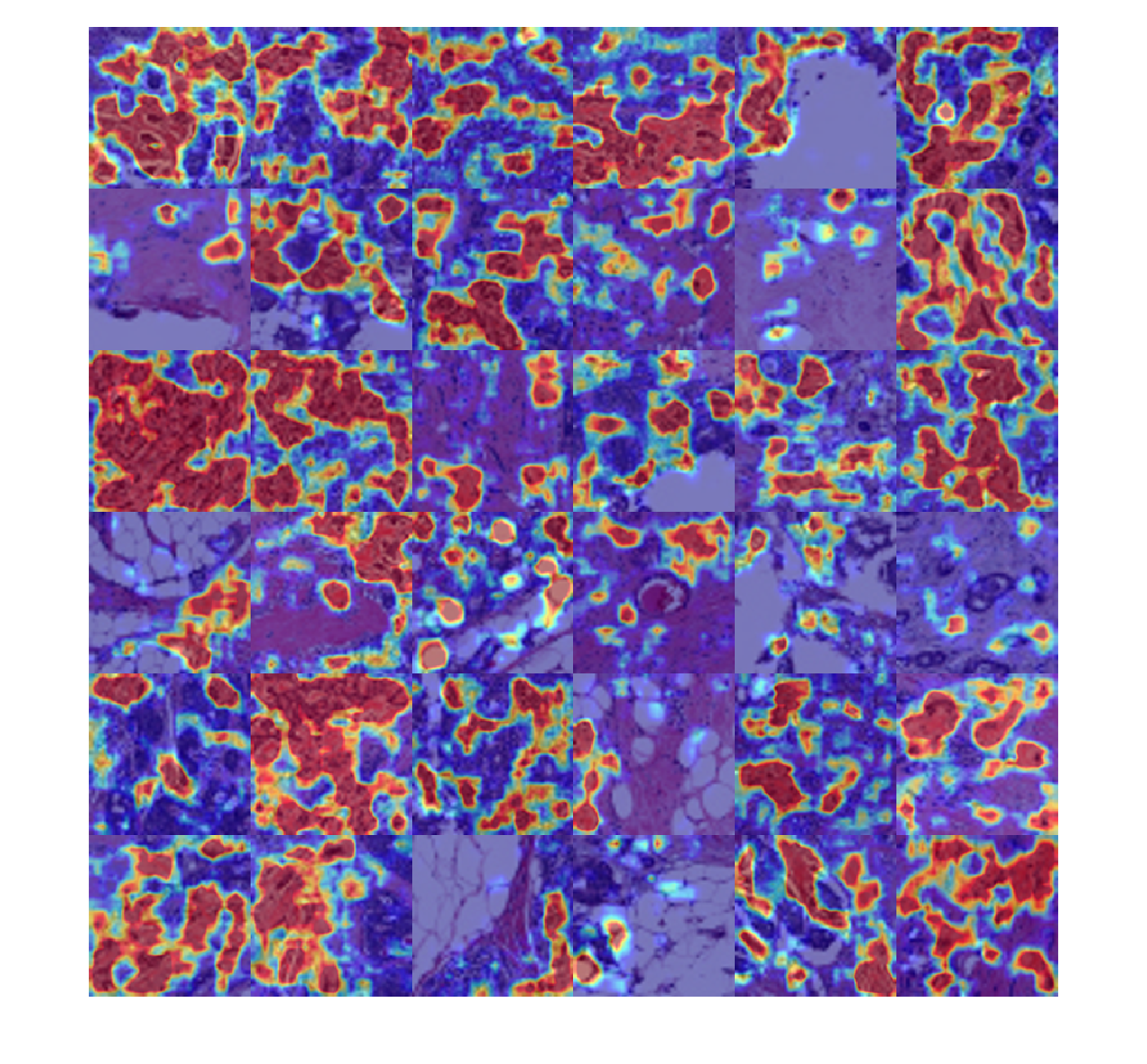} 
\caption{Imbalanced damage images (top) with positive ratio $1/8$, and damage-mark heatmaps (bottom) of breast cancer IDC positive using our deeper FCDD-ResNet101.} 
\label{fig:rawBreast} 
\end{figure} 
\begin{figure}[h] 
\centering 
\includegraphics[width=0.37\textwidth]{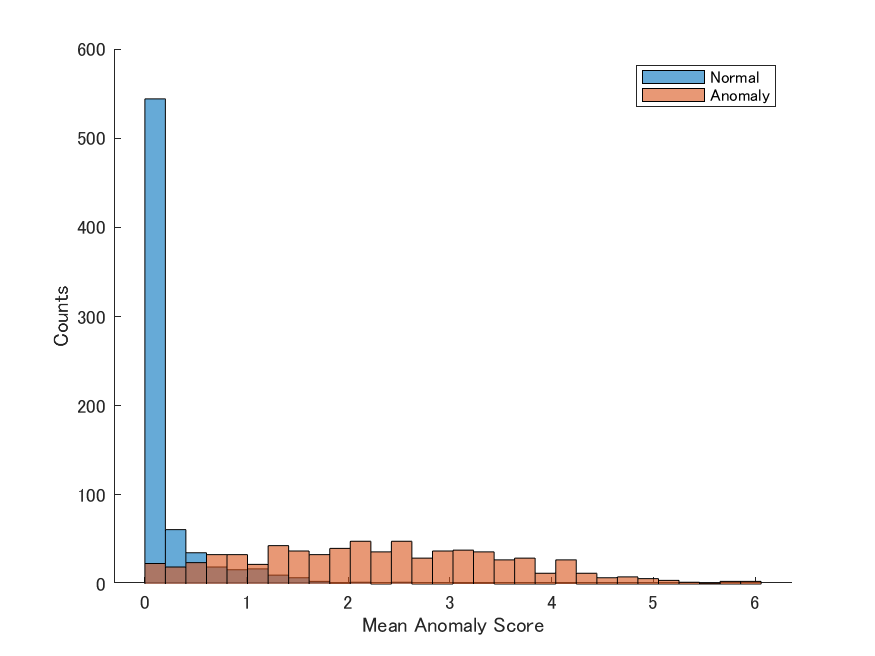} 
\caption{Histogram of breast cancer IDC positive scores using our deeper FCDD-ResNet101 corresponding to the imbalanced damage images with positive ratio $1/8$.} 
\label{fig:histBreast} 
\end{figure} 
 
\begin{figure}[h] 
\centering 
\includegraphics[width=0.37\textwidth]{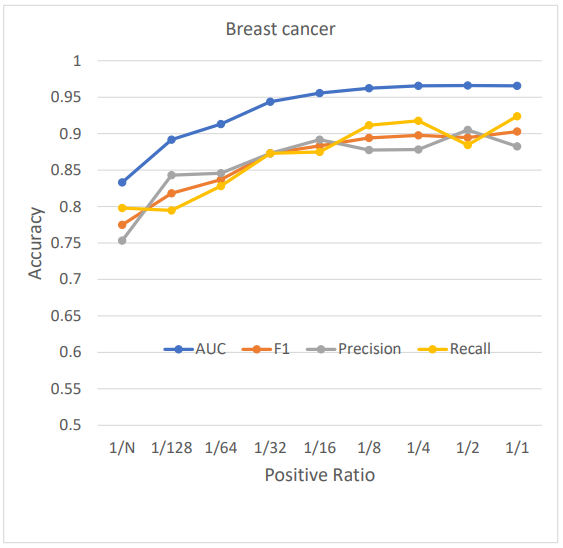} 
\caption{Anomalous vision mining studies on breast cancer indicate that the more anomalies of damage vision, the higher the performance of anomaly detection.} 
\label{fig:EffectBreast}  
\end{figure} 

 \begin{figure}[h] 
\centering 
\includegraphics[width=0.35\textwidth]{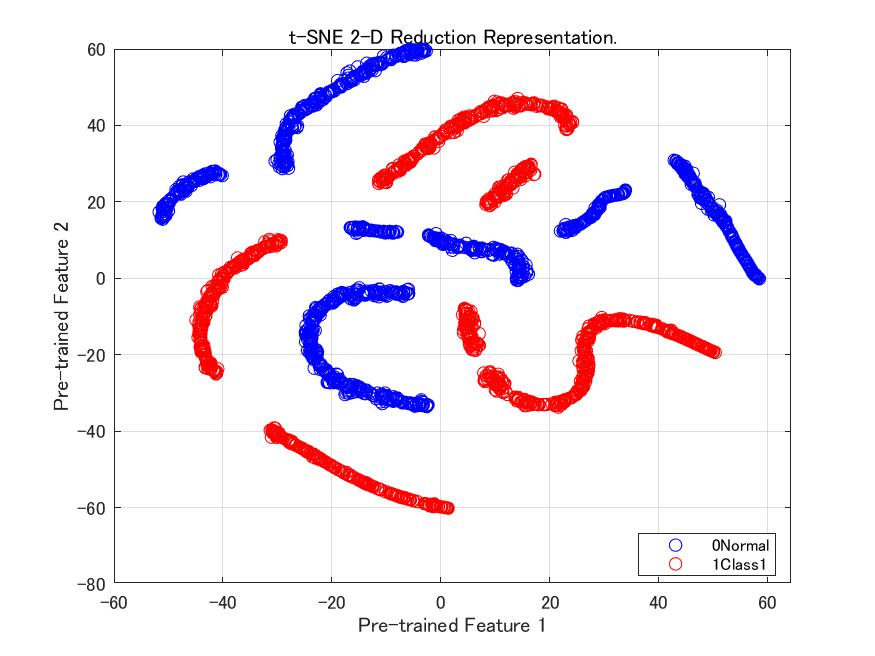} \\ 
\includegraphics[width=0.35\textwidth]{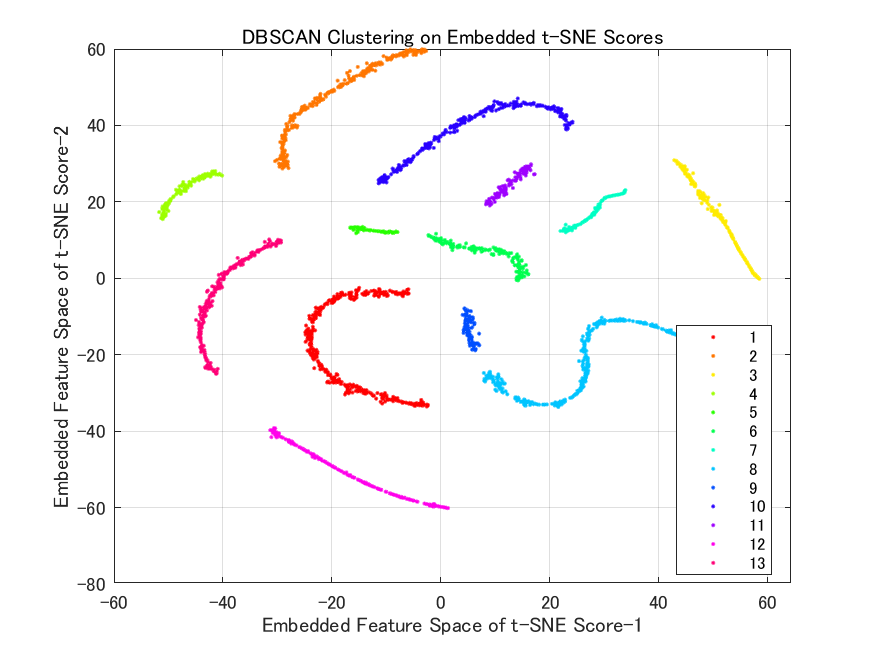} 
\caption{MN-pair contrastive damage representation (top) and density-based clustering (bottom) of breast cancer.} 
\label{fig:mndbBreast} 
\end{figure} 

\subsubsection{Damage-mark Heatmaps on Breast cancer} 
We visualised the damage features using Gaussian upsampling in our deeper FCDD-ResNet101 network. In addition, we generated a histogram of the anomaly scores of the test images in the imbalanced case with a positive ratio of $1/8$. In Fig. \ref{fig:rawBreast}, a damage-mark explanation is represented. The red region in the heatmap represents breast cancer IDC positive features in the specimen patch images.  
Fig. \ref{fig:histBreast} illustrates that a few overlapping bins exist in the boundary of the normal class and breast cancer IDC-positive class along the horizontal anomaly scores. Thus, to detect breast cancer, the score range was well-separated in the specimen-scanned image dataset. 
 
\subsubsection{Feedback Effect on Damage Class Mining} 
As shown in Figure \ref{fig:EffectBreast}, from the viewpoint of all accuracies, the imbalanced studies on breast cancer imply that all accuracies consistently converged into significant phases.  
Ranging with a positive ratio of less than $1/8$, we could understand that there were damage vision mining opportunities with an accuracy gain in terms of the AUC. In contrast, ranging with a positive ratio over $1/4$, it shifted in the over-mining phase without any gain in the AUC. The former phase of damage vision mining opportunities has become beneficial owing to its promising advantage of higher accuracy in all of them. 
 
\subsubsection{Embedding Damage Representation} 
As shown in Figure \ref{fig:mndbBreast}, we analysed feature imbalance in the breast cancer embedding space and implemented MN-pair contrastive damage representation learning and density-based clustering.  
Surprisingly, the number of breast cancer feature clusters was 13, compared to the initial two classes.  
In breast cancer IDC-positive features, a few narrow clusters were distributed in the embedding space. 
 
\subsection{Driving Distraction} 
\subsubsection{Backbone Studies of Supervised Detection} 
As shown in Table~\ref{tab:accBackboneDrive}, our deeper FCDD-Inceptionv3 outperformed in terms of $F_1$, precision, and recall rather than the baseline CNN27 and other backbone-based deeper FCDDs in this driving-distraction image dataset for detecting hazardous driving behaviours. 
\begin{table}[h] 
\caption{Backbone ablation studies on distracted driving detection using our proposed deeper FCDDs.} 
\label{tab:accBackboneDrive} 
\centering 
\begin{tabular}{|c|c|c|c|c|} 
\hline 
\textbf{Backbone} & \textbf{AUC} & \boldmath{$F_1$} & \textbf{Precision} & \textbf{Recall} \\ 
\hline 
CNN27 & 0.9445 & 0.8894 & 0.8839 & 0.8950 \\ \hline 
VGG16 & 0.9981 & 0.9937 & 0.9949 & 0.9925 \\ 
ResNet101 & 0.9955 & 0.9836 & 0.9923 & 0.9750 \\ 
\textbf{Inceptionv3} & \textbf{0.9987} & \textbf{0.9974} & \textbf{1.000} & \textbf{0.9950} \\ \hline 
\end{tabular} 
\end{table} 
 
\subsubsection{Imbalanced-to-unsupervised Training Results} 
As shown in Table~\ref{tab:accImbalanceDrive}, we implemented ablation studies on the imbalanced data containing smaller anomalous and relatively large normal images. In this study, we applied our deeper FCDD-Inceptionv3 and achieved high performance in the aforementioned supervised results.   
Compared with the balanced case of a positive ratio of 1/1, we found that there was an applicable range from an imbalanced ratio of 1/2 to 1/16, where the accuracy loss of recall was consistently less than 3\%.  
However, in the extremely imbalanced range of 1/32 to 1/1300, the accuracy was inferior to that of the applicable range; for example, the loss of $F_1$ was greater than 3\%. The rare positive ratio of 1/32 represents imbalanced data containing quite a little 41 anomalous images and relatively large 1300 normal images. In this case, additional anomalous images were acquired and added to the initial dataset.    
The marginal gain in accuracy was relatively high when driving distraction images of texting and talking on the phone were added.  
\begin{table}[h] 
\caption{Imbalanced data studies using our deeper FCDD-Inceptionv3 for Driving distraction detection $N_3=1300$.} 
\label{tab:accImbalanceDrive} 
\centering 
\begin{tabular}{|c|c|c|c|c|} 
\hline 
\textbf{Positive ratio} & \textbf{AUC} & \boldmath{$F_1$} & \textbf{Precision} & \textbf{Recall} \\ 
\hline 
\textbf{1/1(ano.$N_4$)} & \textbf{0.9987} & \textbf{0.9974} & \textbf{1.000} & \textbf{0.9950} \\ \hline 
1/2(ano.650) & 0.9981 & 0.9886 & 1.000 & 0.9775 \\  
1/4(ano.325) & 0.9976 & 0.9782 & 1.000 & 0.9575 \\  
1/8(ano.163) & 0.9968 & 0.9899 & 0.9949 & 0.9850 \\  
1/16(ano.81)& 0.9965 & 0.9850 & 0.9850 & 0.9850 \\ \hline 
\textbf{1/32(ano.41)}&\textbf{0.9916} & \textbf{0.9664} & \textbf{0.9604} & \textbf{0.9725} \\  
\textbf{1/64(ano.20)}&\textbf{0.9912} & \textbf{0.9542} & \textbf{0.9715} & \textbf{0.9375} \\  
\textbf{1/128(ano.10)}&\textbf{0.9799} & \textbf{0.9107} & \textbf{0.9437} & \textbf{0.8800} \\  
\textbf{1/$N_4$(ano.1)} &\textbf{0.8378} & \textbf{0.7450} & \textbf{0.6971} & \textbf{0.8000} \\ \hline 
\end{tabular} 
\end{table} 
 
\begin{figure}[h] 
\centering 
\includegraphics[width=0.4\textwidth]{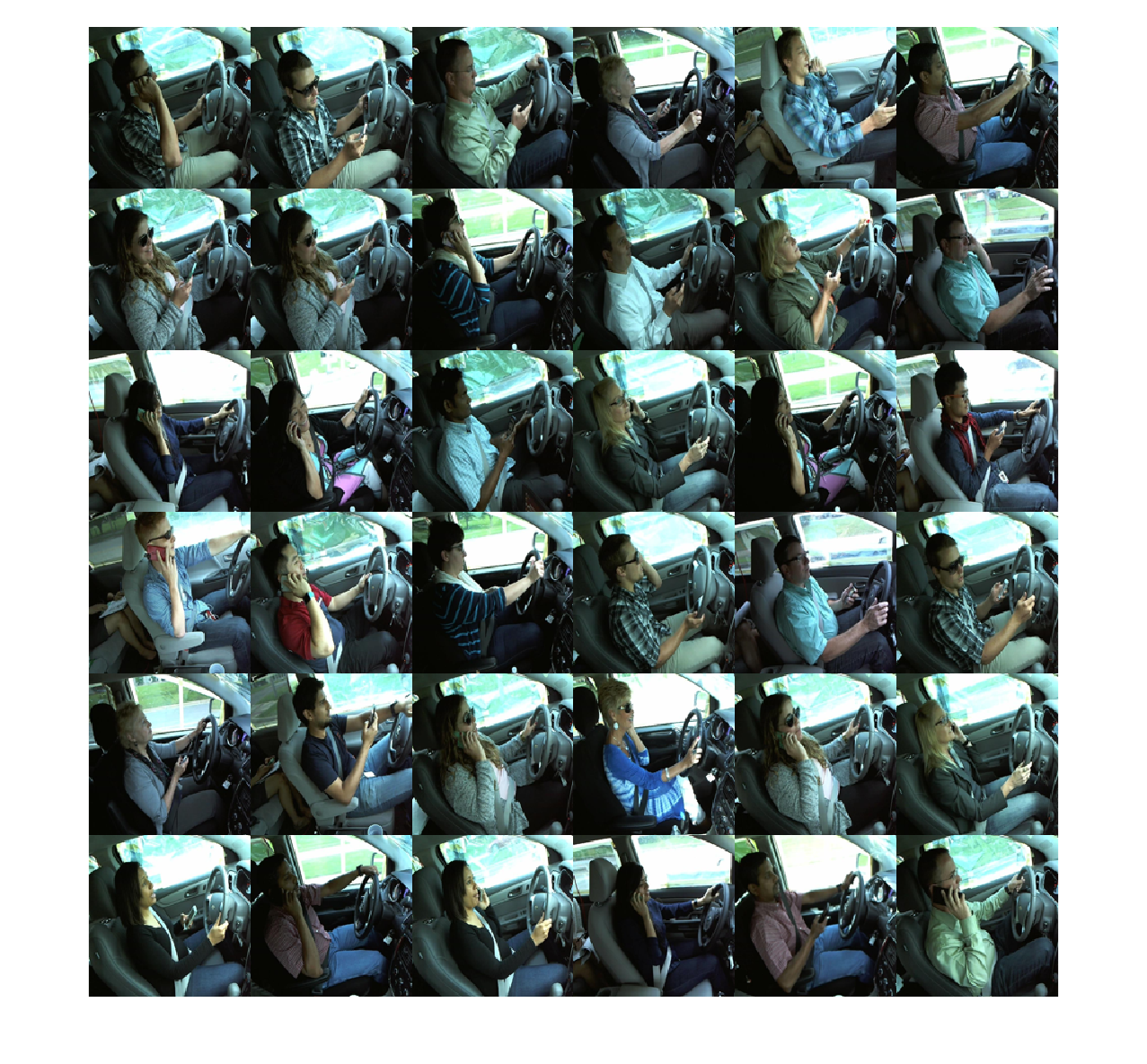} \\ 
\includegraphics[width=0.4\textwidth]{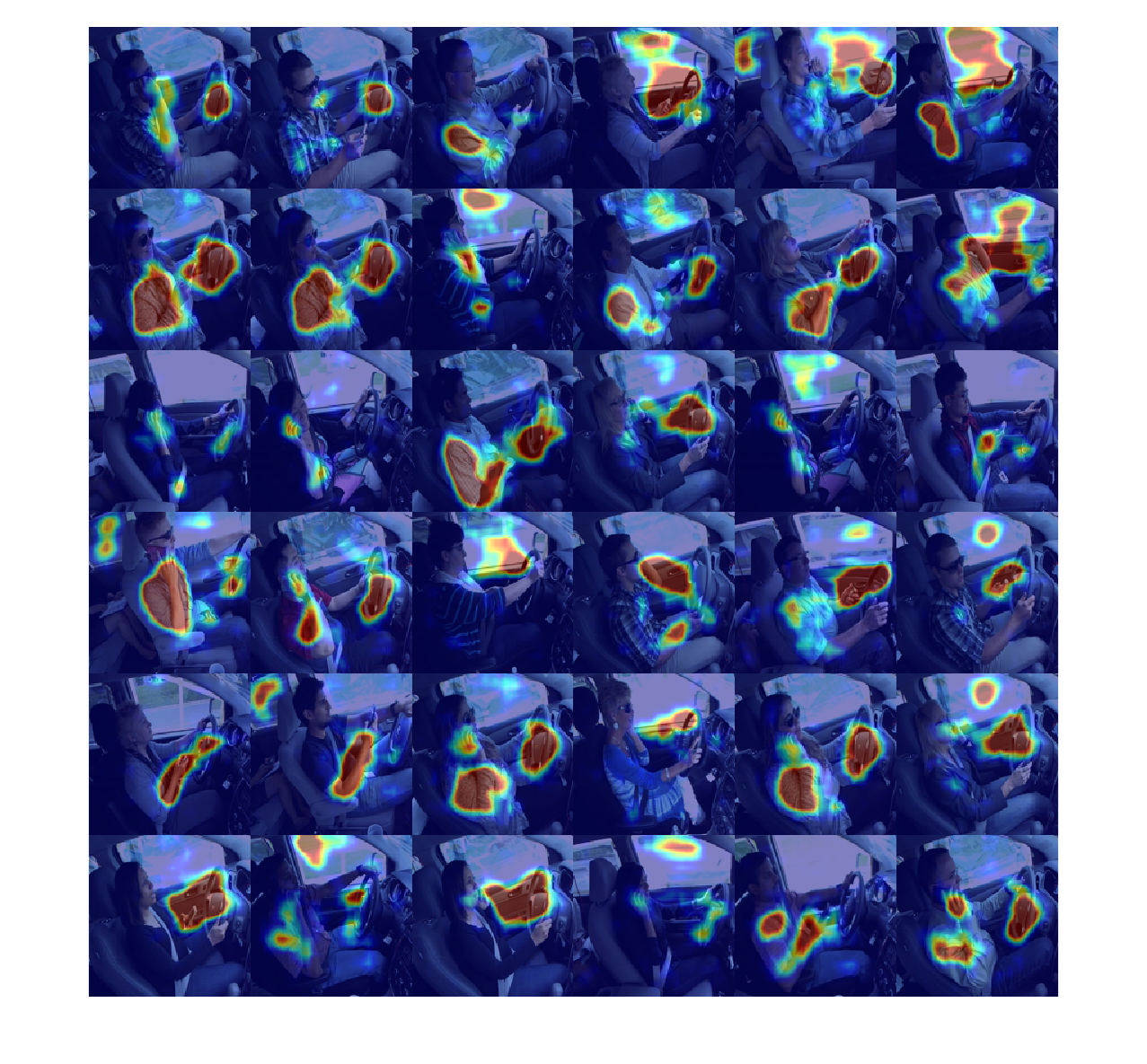} 
\caption{Imbalanced damage images (top) with positive ratio $1/16$, and damage-mark heatmaps (bottom) of distracted driving using our deeper FCDD-Inceptionv3.} 
\label{fig:rawDrive} 
\end{figure} 
\begin{figure}[h] 
\centering 
\includegraphics[width=0.37\textwidth]{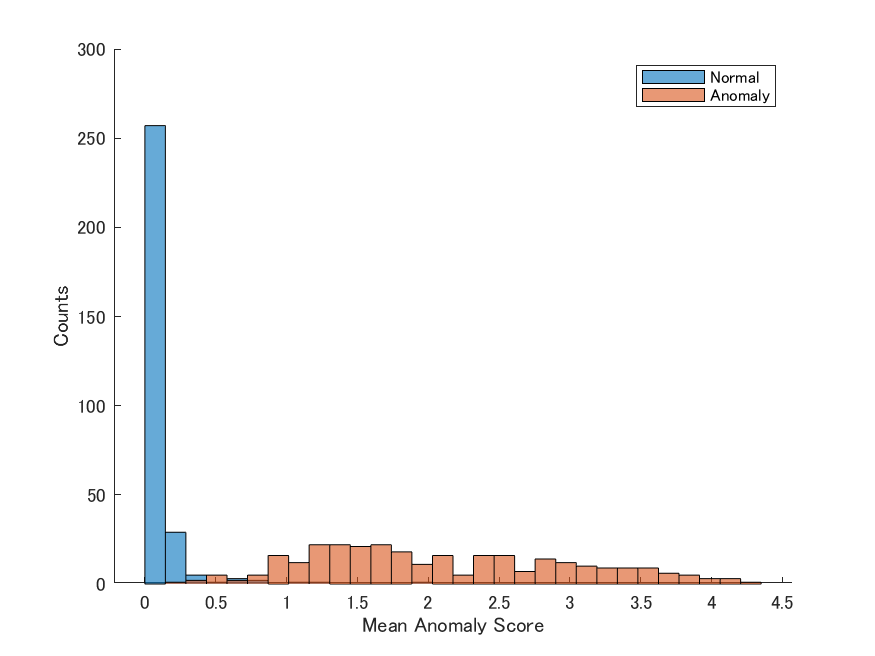} 
\caption{Histogram of distracted driving scores using our deeper FCDD-Inceptionv3 corresponding to the imbalanced damage images with positive ratio $1/16$.} 
\label{fig:histDrive} 
\end{figure} 
 
\begin{figure}[h] 
\centering 
\includegraphics[width=0.37\textwidth]{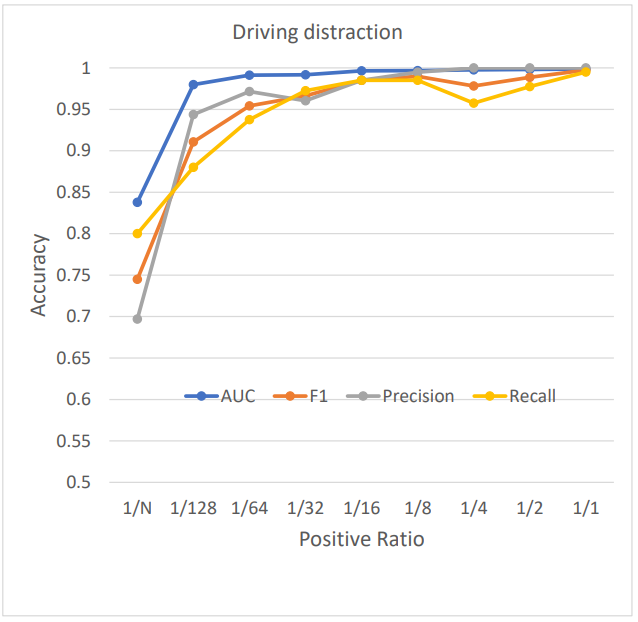} 
\caption{Anomalous vision mining studies on driving distraction indicate that the more anomalies of damage vision, the higher the performance of anomaly detection.} 
\label{fig:EffectDrive}  
\end{figure} 
 
\subsubsection{Damage-mark Heatmaps on Driving Distraction} 
We visualised the damage features using Gaussian upsampling in our deeper FCDD-Inceptionv3 network. In addition, we generated a histogram of the anomaly scores of the test images in the imbalanced case with a positive ratio of $1/16$. In Fig. \ref{fig:rawDrive}, a damage-mark explanation is represented. The red region in the heatmap represents the distracted driving features in which the driver behaves with one hand while texting and talking on the phone without holding the handle in the other hand. These heatmaps focused on anomalous hand movements as a distracted driving behaviour that could potentially cause traffic incidents.  
Fig. \ref{fig:histDrive} illustrates that a few overlapping bins exist in the boundary of the safety driving class and distracted class along the horizontal anomaly scores. Thus, to detect distracted driving behaviours, the score range was well separated in the blood-smear image dataset. 
 
\subsubsection{Feedback Effect on Damage Class Mining} 
As shown in Figure \ref{fig:EffectDrive}, from the viewpoint of all accuracies, the imbalanced studies on driving distraction implied that all accuracies consistently converged into significant phases.  
Ranging with a positive ratio of less than $1/8$, we could understand that there were damage vision mining opportunities with an accuracy gain in terms of the AUC. In contrast, ranging with a positive ratio over $1/4$, it shifted in the over-mining phase without any gain in the AUC. The former phase of damage vision mining opportunities has become beneficial owing to its promising advantage of higher accuracy. 
 
\subsubsection{Embedding Damage Representation} 
\begin{figure}[h] 
\centering 
\includegraphics[width=0.35\textwidth]{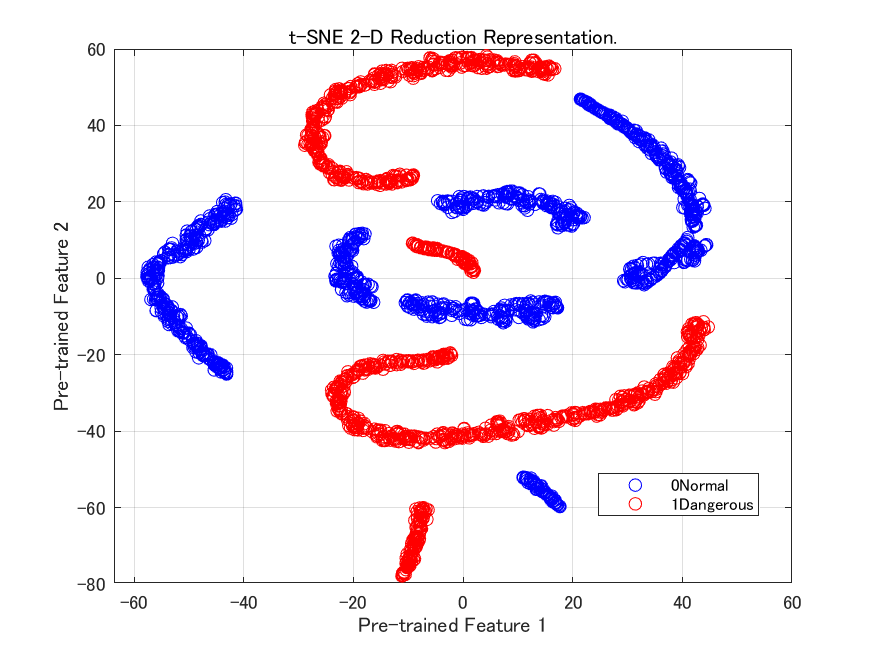} \\ 
\includegraphics[width=0.35\textwidth]{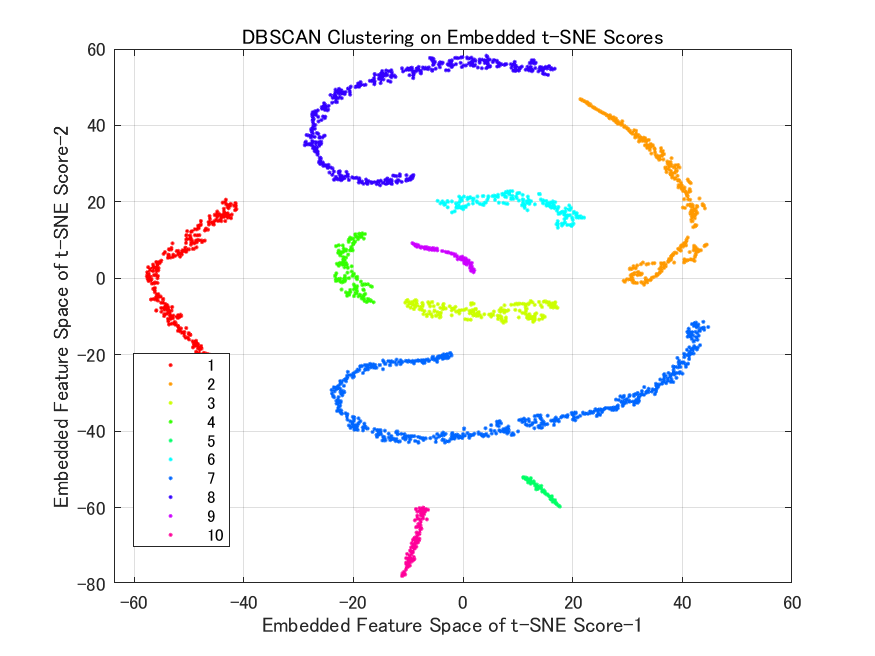} 
\caption{MN-pair contrastive damage representation (top) and density-based clustering (bottom) of driving distraction.} 
\label{fig:mndbDrive} 
\end{figure}

As shown in Figure \ref{fig:mndbDrive}, we analysed feature imbalance in the driving distraction embedding space and implemented MN-pair contrastive damage representation learning and density-based clustering.  
The number of driving distraction feature clusters doubled to 10, twice the initial number of normal and four anomalous classes, that is, texting-left and texting-right, and talking on the phone-left and phone-right.  
In the driving distraction feature, a few narrow clusters were distributed in the embedding space. 
 
\begin{figure}[h] 
\centering 
\includegraphics[width=0.4\textwidth]{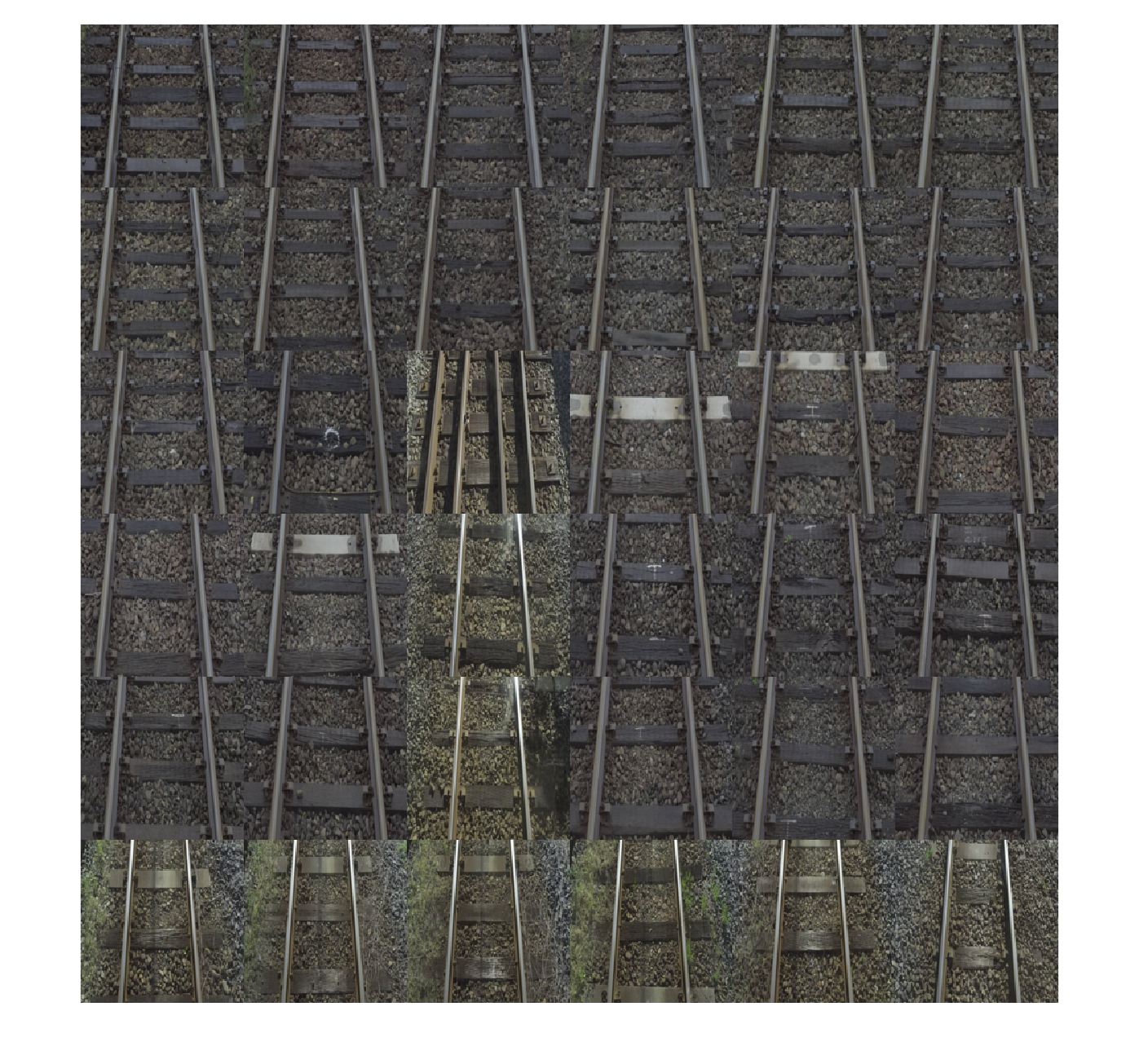} \\ 
\includegraphics[width=0.4\textwidth]{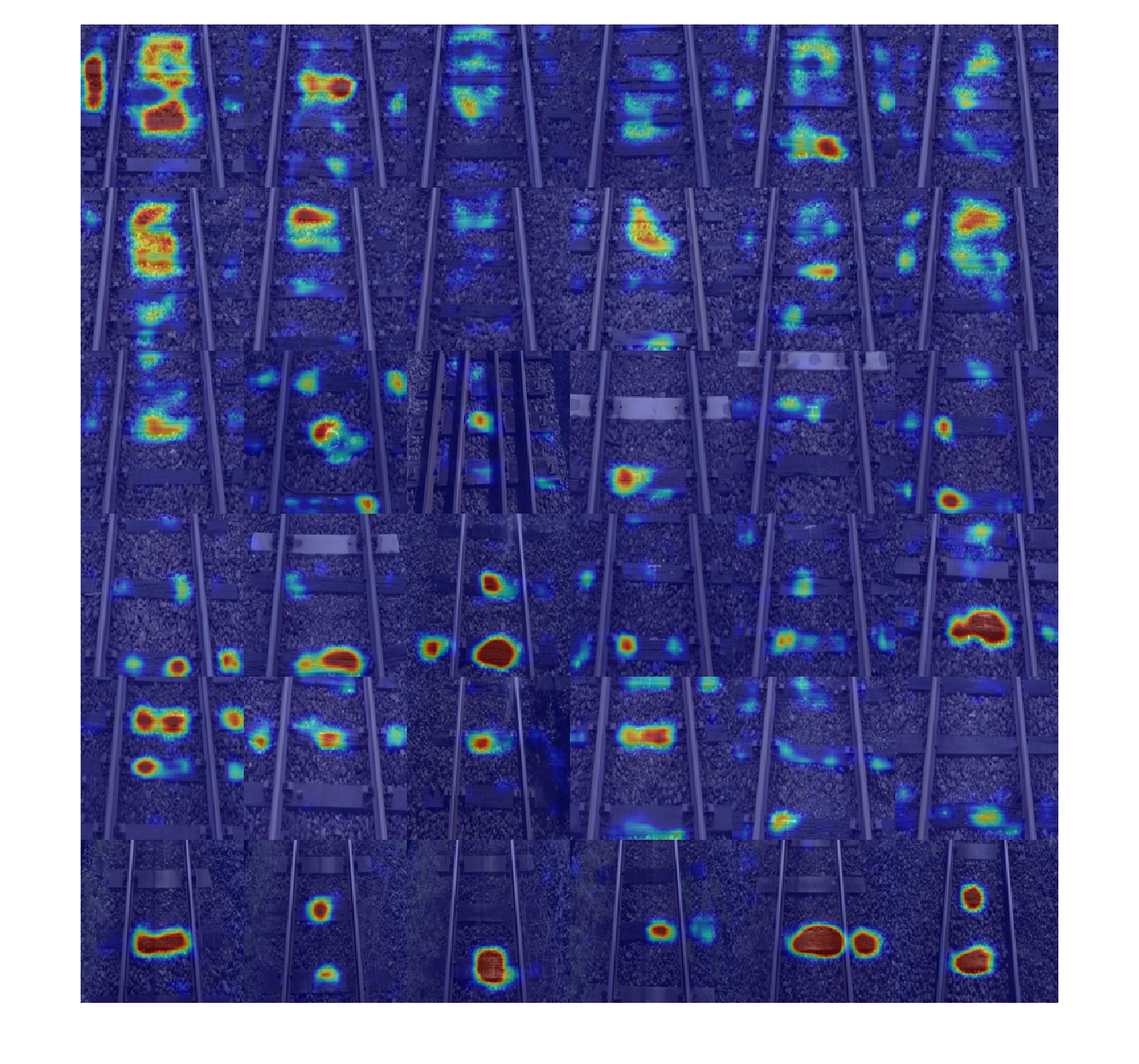} 
\caption{Imbalanced deterioration images (top) with positive ratio $1/16$, and damage-mark heatmaps (bottom) of decayed wooden sleeper using our deeper FCDD-VGG16.} 
\label{fig:rawSleeper} 
\end{figure} 
\begin{figure}[h] 
\centering 
\includegraphics[width=0.37\textwidth]{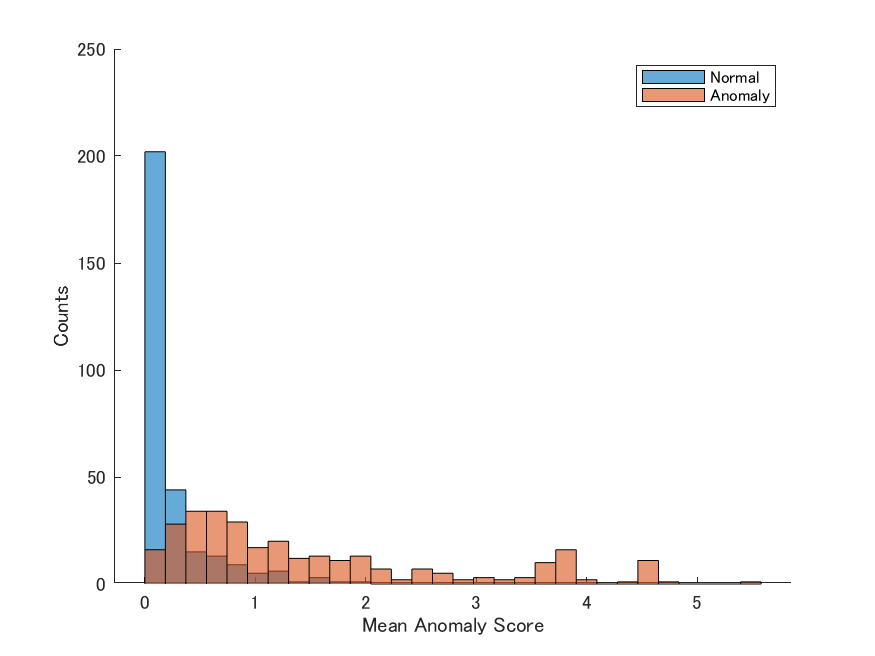} 
\caption{Histogram of decayed wooden sleeper scores using our deeper FCDD-VGG16 corresponding to the imbalanced deterioration images with positive ratio of $1/16$.} 
\label{fig:histSleeper} 
\end{figure} 
 
\subsection{Wooden Deterioration} 
\subsubsection{Backbone Studies of Supervised Detection} 
The number of training images $N_3$ in each class was 1300, and the number for calibration and test are 300 and 400 images respectively, on the dataset of wooden sleeper deterioration.  
As shown in Table~\ref{tab:accBackbone}, our deeper FCDDs based on VGG16 (abbreviated as {\it deeperFCDD-VGG16}) outperformed in terms of the AUC and $F_1$ rather than the baseline CNN27 and other backbone-based deeper FCDDs in this railway dataset for detecting decayed wooden sleeper. 
\begin{table}[h] 
\caption{Backbone ablation studies on defective detection using our proposed deeper FCDDs for Wooden sleepers.} 
\label{tab:accBackbone} 
\centering 
\begin{tabular}{|c|c|c|c|c|} 
\hline 
\textbf{Backbone} & \textbf{AUC} & \boldmath{$F_1$} & \textbf{Precision} & \textbf{Recall} \\ 
\hline 
CNN27 & 0.8624 & 0.7688 & 0.7088 & 0.8400 \\ \hline 
\textbf{VGG16} & \textbf{0.9425} & \textbf{0.8475} & \textbf{0.8770} & \textbf{0.8200} \\ 
ResNet101 &0.9304 & 0.8108 & 0.8823 & 0.7500 \\ 
Inceptionv3 &0.9412 & 0.8041 & 0.8415 & 0.7700 \\ \hline 
\end{tabular} 
\end{table} 
 
\subsubsection{Imbalanced-to-unsupervised Training Results} 
As shown in Table~\ref{tab:accImbalance}, we implemented ablation studies on the imbalanced data containing fewer anomalous and relatively large normal images. In this study, we applied deeper FCDD-VGG16 to achieve high performance in the aforementioned results.   
Compared with the balanced case with a positive ratio of 1/1, we found that there was an applicable range from a balanced ratio of 1/2 to 1/16, where every accuracy was consistently high.  
However, in the extremely imbalanced ranges from 1/32 to 1/128 and 1/1300, the accuracy significantly decreased and was inferior to that of the applicable range. Specifically, the rare positive ratio 1/32 represents imbalanced data that contain a few 41 anomalous images and relatively large 1300 normal images. In this case, additional anomalous images were acquired and added to the initial dataset. The marginal effect of the accuracy gain was significantly high when the anomalous images were added. 
    
\begin{table}[h] 
\caption{Imbalanced data studies using our deeper FCDD-VGG16 for wooden sleeper deterioration $N_3=1300$.} 
\label{tab:accImbalance} 
\centering 
\begin{tabular}{|c|c|c|c|c|} 
\hline 
\textbf{Positive ratio} & \textbf{AUC} & \boldmath{$F_1$} & \textbf{Precision} & \textbf{Recall} \\ 
\hline 
\textbf{1/1(ano.$N_5$)} & \textbf{0.9463} & \textbf{0.8701} & \textbf{0.8379} & \textbf{0.9050} \\ \hline 
1/2(ano.650) & 0.9190 & 0.8751 & 0.8205 & 0.9375 \\  
1/4(ano.325) & 0.9287 & 0.8505 & 0.8274 & 0.8750 \\  
1/8(ano.163) & 0.9269 & 0.8451 & 0.8378 & 0.8525 \\  
1/16(ano.81)& 0.9101 & 0.8547 & 0.8353 & 0.8750 \\ \hline 
\textbf{1/32(ano.41)}&\textbf{0.8947} & \textbf{0.8441} & \textbf{0.8110} & \textbf{0.8800} \\  
\textbf{1/64(ano.20)}&\textbf{0.8724} & \textbf{0.8250} & \textbf{0.7562} & \textbf{0.9075} \\  
\textbf{1/128(ano.10)}&\textbf{0.8051} & \textbf{0.7698} & \textbf{0.7622} & \textbf{0.7775} \\  
\textbf{1/$N_5$(ano.1)} &\textbf{0.6136} & \textbf{0.5823} & \textbf{0.5724} & \textbf{0.5925} \\ \hline 
\end{tabular} 
\end{table} 
 
\subsubsection{Damage-mark Heatmaps on Wooden Decayed} 
We visualised the damage features using Gaussian upsampling in our deeper FCDD-VGG16 network. In addition, we generated a histogram of the anomaly scores of the test images in the imbalanced case with a positive ratio of $1/16$.  
At the bottom of Fig. ref{fig:rawSleeper}, a damage-mark explanation is presented. The red region in the heatmap represents decayed wooden sleepers. There were several regions of background noise over the ballast stones, precast concrete white sleepers in the third and fourth rows, and the grass outside the rail track in the sixth row. 
In addition, Fig. \ref{fig:histSleeper} illustrates that several overlapping bins exist in the boundary of horizontal anomaly scores between the health and deterioration class. Therefore, to detect the decayed wooden sleepers, the anomaly score range was moderately separated from the wooden sleeper deterioration dataset. 
 
\begin{figure}[h] 
\centering 
\includegraphics[width=0.37\textwidth]{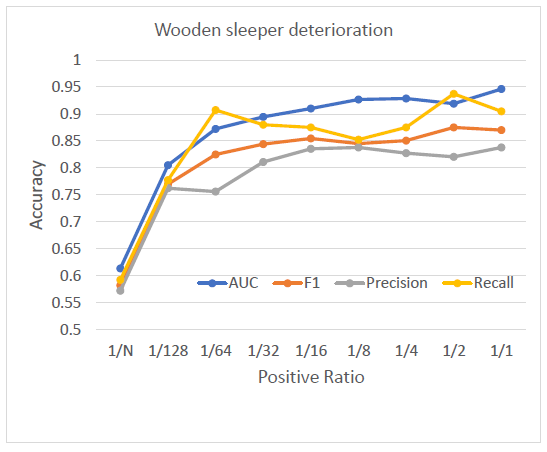} 
\caption{Anomalous vision mining studies on wooden sleeper deterioration indicate that the more anomalies of damage vision, the higher the performance of anomaly detection.} 
\label{fig:EffectSleeper} 
\end{figure} 
 
\subsubsection{Feedback Effect on Deterioration Class Mining} 
As shown in Figure \ref{fig:EffectSleeper}, from the viewpoint of accuracy on the AUC, imbalanced studies on wooden sleeper deterioration imply that the accuracy was moving into stable phases, even though the recall and precision were waving.  
Ranging with a positive ratio of less than $1/8$, we can understand that there were damage vision mining opportunities with an accuracy gain. In contrast, when ranging with a positive ratio over $1/4$, it shifted in the over-mining phase without an effective gain in accuracy. The former phase of damage vision mining opportunities is beneficial because of the advantage of higher accuracy in terms of the AUC. 
 
\subsubsection{Embedding Damage Representation} 
\begin{figure}[h] 
\centering 
\includegraphics[width=0.35\textwidth]{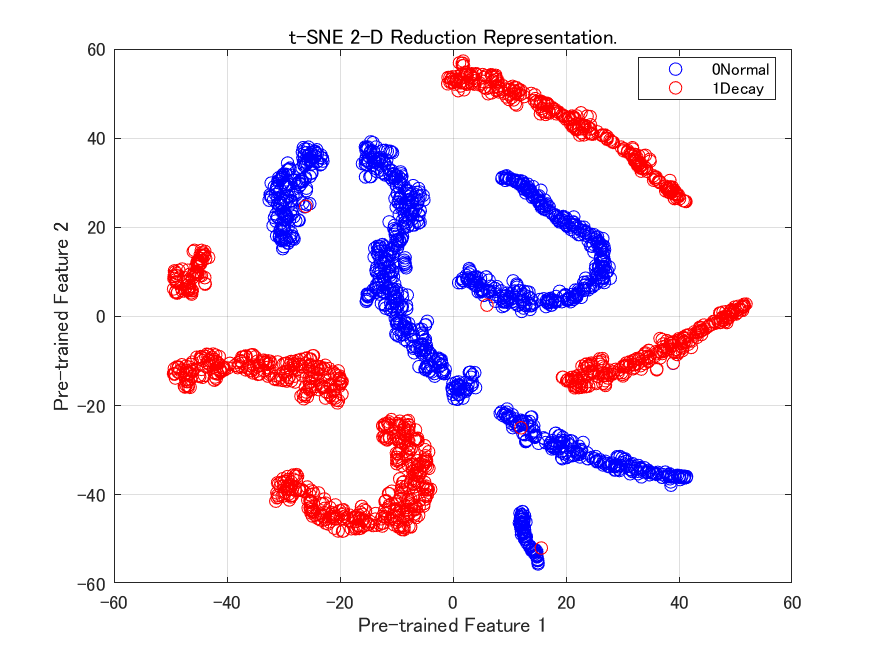} \\ 
\includegraphics[width=0.35\textwidth]{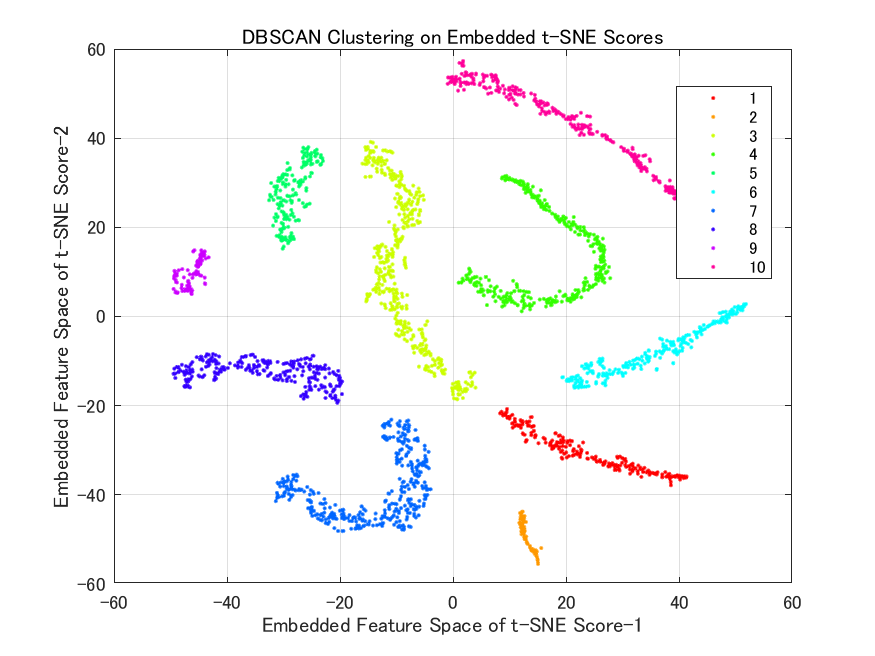} 
\caption{MN-pair contrastive damage representation (top) and density-based clustering (bottom) of wooden deterioration.} 
\label{fig:mndbWood} 
\end{figure}

As shown in Figure \ref{fig:mndbWood}, we analysed the feature imbalance in the wooden deterioration embedding space and implemented our MN-pair contrastive damage representation learning and density-based clustering.  
Surprisingly, the number of wooden deterioration feature clusters increased to 10 rather than the initial two classes. In the wooden deterioration feature, a few narrow clusters were distributed in the embedding space. 
 
\begin{figure}[h] 
\centering 
\includegraphics[width=0.4\textwidth]{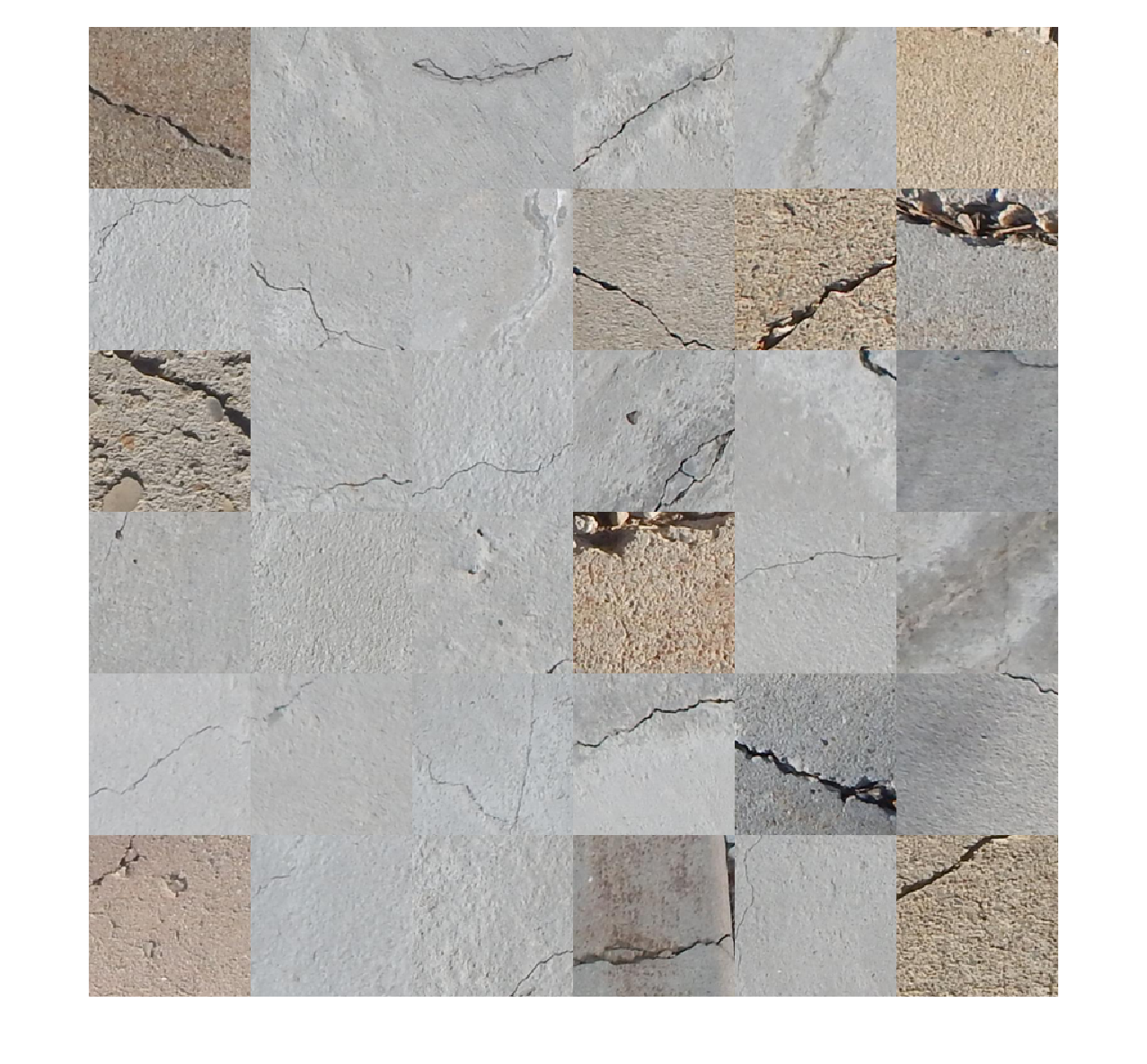} \\ 
\includegraphics[width=0.4\textwidth]{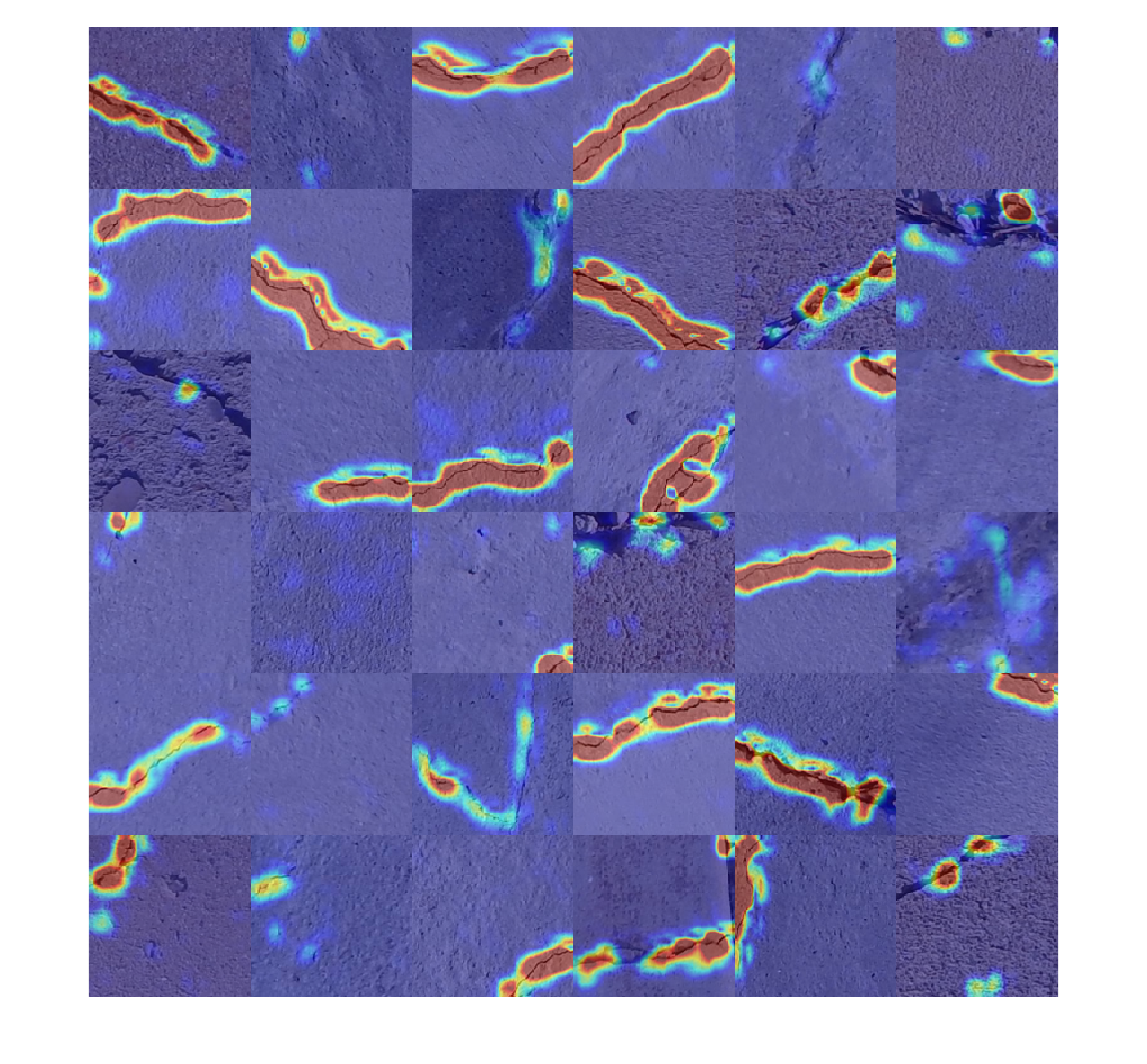} 
\caption{Imbalanced deterioration images (top) with positive ratio $1/8$, and damage-mark heatmaps (bottom) of concrete crack using our deeper FCDD-VGG16.} 
\label{fig:rawCrack} 
\end{figure} 
\begin{figure}[h] 
\centering 
\includegraphics[width=0.37\textwidth]{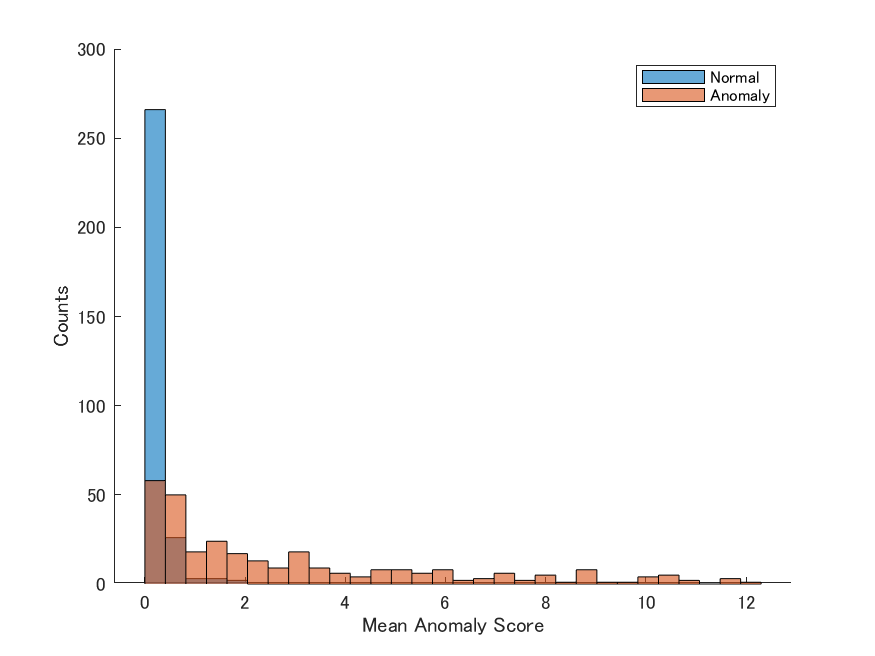} 
\caption{Histogram of concrete crack scores using our deeper FCDD-VGG16 corresponding to the imbalanced deterioration images with positive ratio $1/8$.} 
\label{fig:histCrack} 
\end{figure} 
 
\subsection{Concrete Crack} 
\subsubsection{Backbone Studies of Supervised Detection} 
As shown in Table~\ref{tab:accBackboneCrack}, our deeper FCDD-VGG16 outperformed in terms of accuracy, such as AUC, $F_1$, precision, and recall, rather than those of the baseline CNN27 and other backbone-based deeper FCDDs in this concrete deterioration dataset for detecting cracks. 
\begin{table}[h] 
\caption{Backbone ablation studies on concrete crack detection using our proposed deeper FCDDs.} 
\label{tab:accBackboneCrack} 
\centering 
\begin{tabular}{|c|c|c|c|c|} 
\hline 
\textbf{Backbone} & \textbf{AUC} & \boldmath{$F_1$} & \textbf{Precision} & \textbf{Recall} \\ 
\hline 
CNN27 & 0.7150 & 0.6975 & 0.6245 & 0.7900 \\ \hline 
\textbf{VGG16} & \textbf{0.9287} & \textbf{0.8384} & \textbf{0.8605} & \textbf{0.8175} \\ 
ResNet101 &0.9120 & 0.8206 & 0.8480 & 0.7950 \\ 
Inceptionv3 &0.9119 & 0.8169 & 0.8372 & 0.7975 \\ \hline 
\end{tabular} 
\end{table} 
 
\subsubsection{Imbalanced-to-unsupervised Training Results} 
As shown in Table~\ref{tab:accImbalanceCrack}, we performed ablation studies on the imbalanced damage data containing fewer anomalous and relatively large normal images. In this study, we applied our deeper FCDD-VGG16 to achieve high performance in supervised results.   
Compared with the balanced case of a positive ratio of 1/1, we found that there was an applicable range from a balanced ratio of 1/2 to 1/8, where the accuracy was consistently high in terms of the AUC and $F_1$.  
However, in the extremely imbalanced range of 1/16–1/1300, the accuracy was inferior to that in the aforementioned applicable range. The rare positive ratio of 1/16 represents imbalanced data that contain a small number of anomalous images (81) and relatively large normal images (1300). In this case, additional anomalous images were acquired and added to the initial dataset. The marginal gain in accuracy was relatively high when the deteriorated images were added. 
 
\begin{table}[h] 
\caption{Imbalanced data studies using our deeper FCDD-VGG16 for Concrete crack detection $N_4=1300$.} 
\label{tab:accImbalanceCrack} 
\centering 
\begin{tabular}{|c|c|c|c|c|} 
\hline 
\textbf{Positive ratio} & \textbf{AUC} & \boldmath{$F_1$} & \textbf{Precision} & \textbf{Recall} \\ 
\hline 
\textbf{1/1(ano.$N_6$)} & \textbf{0.9338} & \textbf{0.8265} & \textbf{0.9482} & \textbf{0.7325} \\ \hline 
1/2(ano.650) & 0.8968 & 0.8492 & 0.9016 & 0.8025 \\  
1/4(ano.325) & 0.9151 & 0.8151 & 0.8856 & 0.7550 \\  
1/8(ano.163) & 0.9147 & 0.8178 & 0.8885 & 0.7575 \\ \hline  
\textbf{1/16(ano.81)}&\textbf{0.8956} & \textbf{0.7918} & \textbf{0.8594} & \textbf{0.7750} \\ 
\textbf{1/32(ano.41)}&\textbf{0.8826} & \textbf{0.7918} & \textbf{0.8617} & \textbf{0.7325} \\  
\textbf{1/64(ano.20)}&\textbf{0.8788} & \textbf{0.7614} & \textbf{0.8952} & \textbf{0.6625} \\  
\textbf{1/128(ano.10)}&\textbf{0.8488} & \textbf{0.7104} & \textbf{0.8277} & \textbf{0.6225} \\  
\textbf{1/$N_6$(ano.1)} &\textbf{0.7927} & \textbf{0.6908} & \textbf{0.8235} & \textbf{0.5950} \\ \hline 
\end{tabular} 
\end{table} 
 
\subsubsection{Damage-mark Heatmaps on Concrete Crack} 
We visualised the damage features using Gaussian upsampling in our deeper FCDD-VGG16 network. In addition, we generated a histogram of the anomaly scores of the test images in the imbalanced case with a positive ratio of $1/8$. In the bottom of Fig. \ref{fig:rawCrack}, a damage-mark explanation is presented. The red region in the heatmap represents the cracks on the surface of the concrete deck and pavement. In addition, Fig. \ref{fig:histCrack} illustrates that a few overlapping bins exist in the boundary of horizontal anomaly scores between the health class and crack deterioration class. 
Therefore, to detect cracks, the score range was well separated in the concrete crack dataset of the deck and pavement. 
 
\begin{figure}[h] 
\centering 
\includegraphics[width=0.37\textwidth]{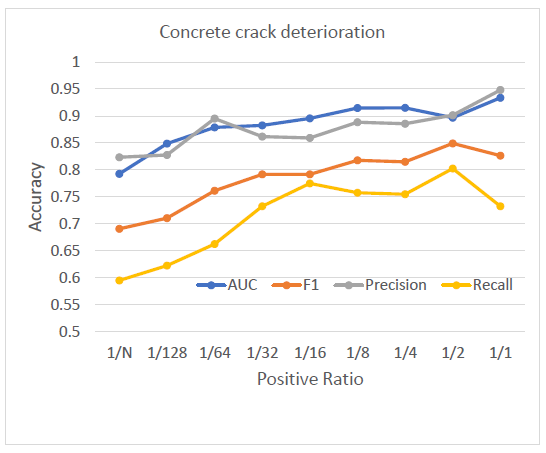} 
\caption{Anomalous vision mining studies on concrete crack, indicates that the more anomalies of damage vision, the higher performance of anomaly detection.} 
\label{fig:EffectCrack} 
\end{figure} 
 
\subsubsection{Feedback Effect on Deterioration Class Mining} 
As shown in Figure \ref{fig:EffectCrack}, from the viewpoint of the primary accuracy AUC, imbalanced studies on concrete cracks imply that the accuracy increased within a range less than the positive ratio $1/8$, although the precision and recall were waving.  
We could understand that the beneficial range was damage vision mining opportunities with an accuracy gain. In contrast, when ranging with a positive ratio over $1/4$, it shifted in the over-mining phase without an effective gain in accuracy. The former phase of damage vision mining opportunities is beneficial because of the advantage of higher accuracy in terms of the AUC. 
 
\subsubsection{Embedding Damage Representation} 
\begin{figure}[h] 
\centering 
\includegraphics[width=0.35\textwidth]{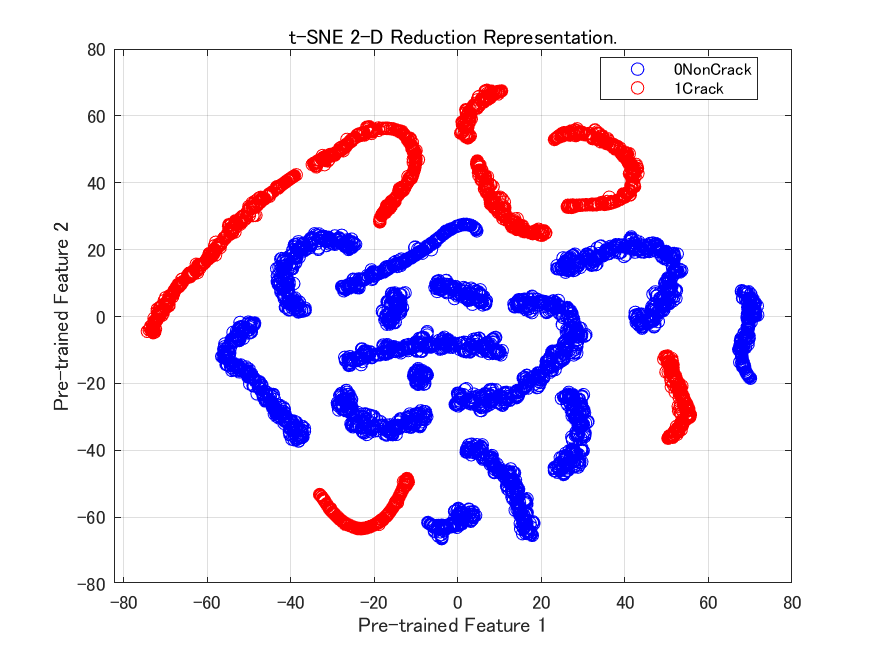} \\ 
\includegraphics[width=0.35\textwidth]{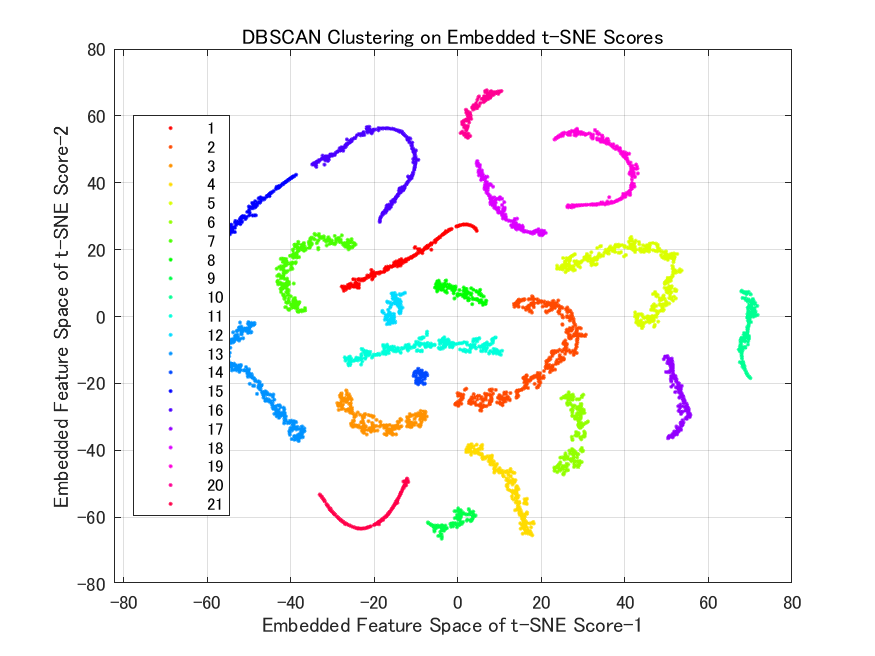} 
\caption{MN-pair contrastive damage representation (top) and density-based clustering (bottom) of concrete deterioration.} 
\label{fig:mndbConcrete} 
\end{figure}

As shown in Figure \ref{fig:mndbConcrete}, we analysed the feature imbalance in the concrete deterioration embedding space and implemented our MN-pair contrastive damage representation learning and density-based clustering.  
Surprisingly, the number of concrete deterioration clusters increased to 21 from the initial two classes. In the concrete deterioration feature, many narrow clusters are distributed in the embedding space. 
 
\subsection{Logical Defects} 
\subsubsection{Backbone Studies of Supervised Detection} 
As shown in Table~\ref{tab:accBackboneLogi}, our deeper FCDD-VGG16 outperformed in terms of accuracy, such as $F_1$ and precision, rather than those of the baseline CNN27 and other backbone-based deeper FCDDs in this logical constraints MVTec LOCO AD dataset for detecting logical defects. 
\begin{table}[h] 
\caption{Backbone ablation studies on logical defects detection using our proposed deeper FCDDs.} 
\label{tab:accBackboneLogi} 
\centering 
\begin{tabular}{|c|c|c|c|c|} 
\hline 
\textbf{Backbone} & \textbf{AUC} & \boldmath{$F_1$} & \textbf{Precision} & \textbf{Recall} \\ 
\hline 
CNN27 & 0.8014 & 0.6266 & 0.5468 & 0.7336 \\ \hline 
\textbf{VGG16} & \textbf{0.8925} & \textbf{0.7146} & \textbf{0.7315} & \textbf{0.6984} \\ 
ResNet101 &0.8669 & 0.6912 & 0.6383 & 0.7537 \\ 
Inceptionv3 &0.8975 & 0.6995 & 0.6859 & 0.7135 \\ \hline 
\end{tabular} 
\end{table} 
 
\subsubsection{Imbalanced-to-unsupervised Training Results} 
As shown in Table~\ref{tab:accImbalanceLogi}, we performed ablation studies on the imbalanced damage data containing fewer anomalous and relatively large normal images. In this study, we applied our deeper FCDD-VGG16 and achieved high performance in the above-supervised results.   
In the MVTec LOCO AD dataset, the number of anomalies was 993, which included 432 structural damages and 561 logical defects. This imbalance study began at a positive ratio of 1/2.  
Compared to the balanced case of a positive ratio of 1/2, we found that there was an applicable range from an imbalanced ratio of 1/4 to 1/8, where the accuracy of the AUC was consistently more than 85\%.  
However, in the extremely imbalanced range of 1/16 to 1/1300, the accuracy was inferior to the applicable range, that is, the AUC was greater than 85\%. The rare positive ratio of 1/16 represents imbalanced data that contain a small number of anomalous images (81) and relatively large normal images (1300). In this case, additional anomalous images were acquired and added to the initial dataset. The marginal accuracy gain was relatively high by adding the logical defective images. 
 
\begin{table}[h] 
\caption{Imbalanced data studies using our deeper FCDD-VGG16 for Logical defects detection $N_7=1300$.} 
\label{tab:accImbalanceLogi} 
\centering 
\begin{tabular}{|c|c|c|c|c|} 
\hline 
\textbf{Positive ratio} & \textbf{AUC} & \boldmath{$F_1$} & \textbf{Precision} & \textbf{Recall} \\ 
\hline 
\textbf{1/2(ano.645)} & \textbf{0.8925} & \textbf{0.7146} & \textbf{0.7315} & \textbf{0.6984} \\ \hline 
1/4(ano.325) & 0.8843 & 0.6986 & 0.7443 & 0.6582 \\  
1/8(ano.163) & 0.8504 & 0.6888 & 0.6531 & 0.7286 \\ \hline  
\textbf{1/16(ano.81)}&\textbf{0.7881} & \textbf{0.6716} & \textbf{0.6700} & \textbf{0.6733} \\ 
\textbf{1/32(ano.41)}&\textbf{0.7756} & \textbf{0.6011} & \textbf{0.7074} & \textbf{0.5226} \\  
\textbf{1/64(ano.20)}&\textbf{0.6913} & \textbf{0.5336} & \textbf{0.4817} & \textbf{0.5979} \\  
\textbf{1/128(ano.10)}&\textbf{0.6337} & \textbf{0.5271} & \textbf{0.5425} & \textbf{0.5125} \\  
\textbf{1/$N_7$(ano.1)} &\textbf{0.5520} & \textbf{0.4019} & \textbf{0.3835} & \textbf{0.4221} \\ \hline 
\end{tabular} 
\end{table} 

 \begin{figure}[h] 
\centering 
\includegraphics[width=0.4\textwidth]{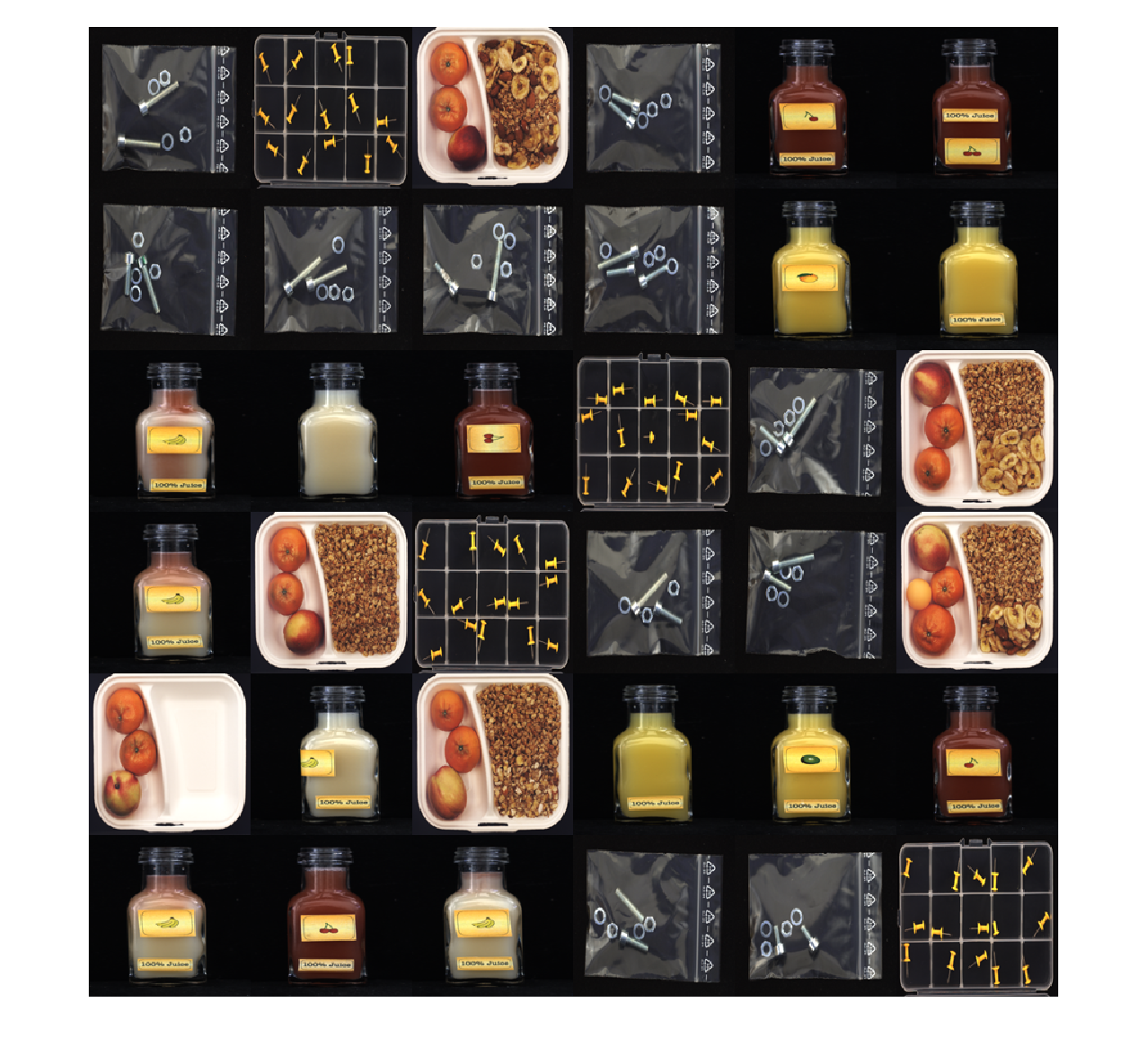} \\ 
\includegraphics[width=0.4\textwidth]{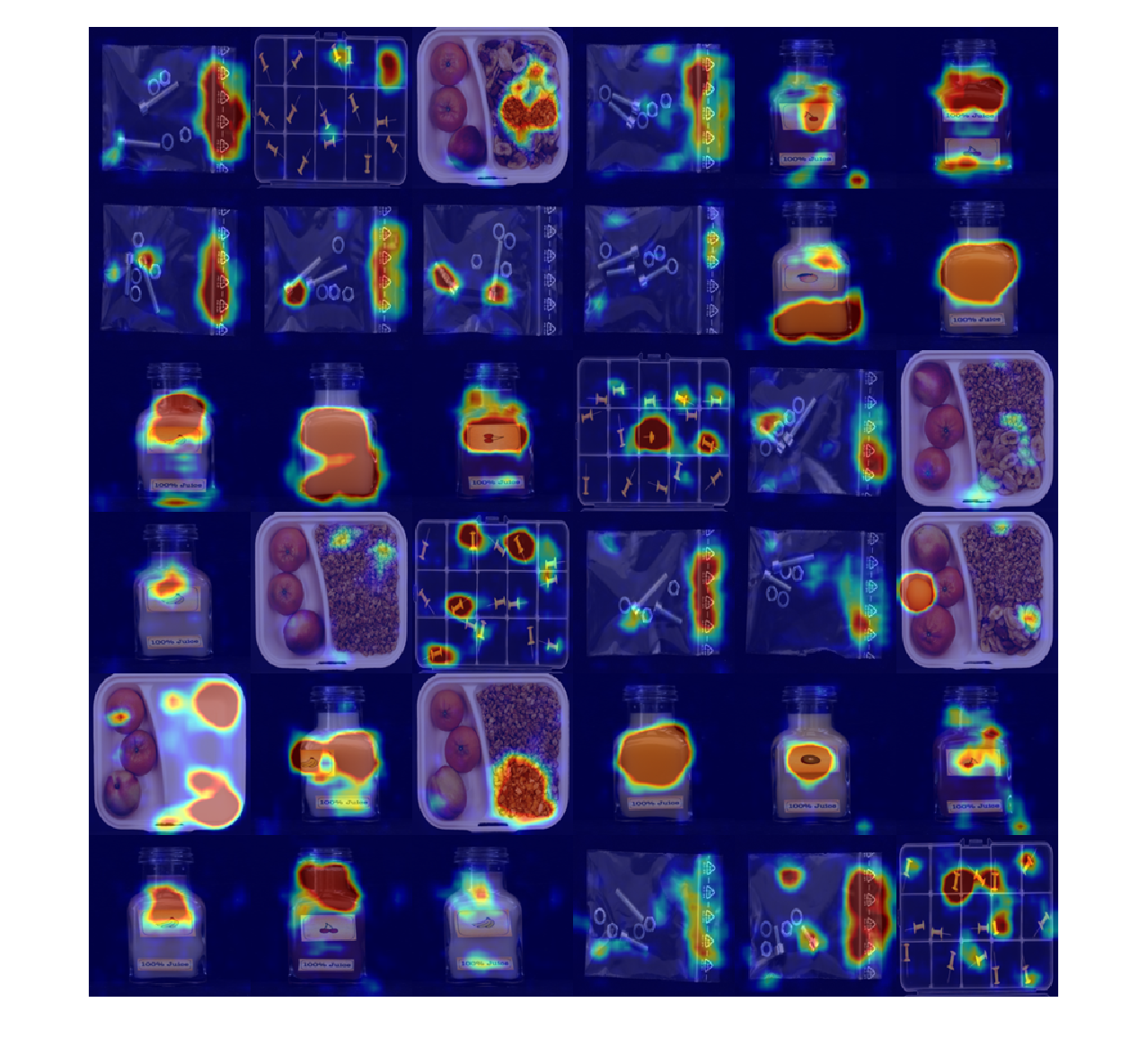} 
\caption{Imbalanced deterioration images (top) with positive ratio $1/8$, and damage-mark heatmaps (bottom) of logical defects using our deeper FCDD-VGG16.} 
\label{fig:rawLogi} 
\end{figure} 
\begin{figure}[h] 
\centering 
\includegraphics[width=0.37\textwidth]{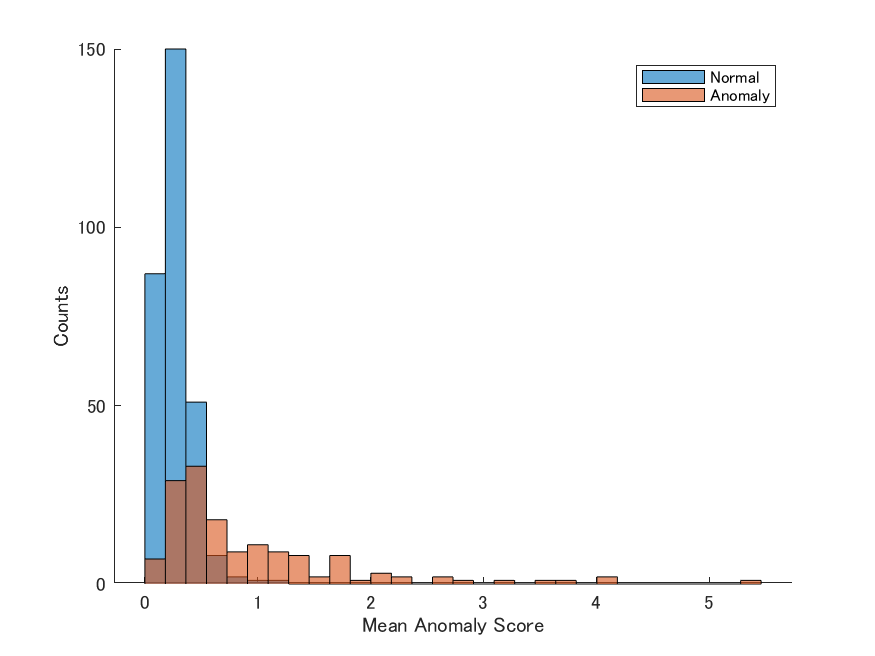} 
\caption{Histogram of logical defects scores using our deeper FCDD-VGG16 corresponding to the imbalanced deterioration images with positive ratio $1/8$.} 
\label{fig:histLogi} 
\end{figure} 

\subsubsection{Damage-mark Heatmaps on Logical Defects} 
We visualised the damage features using Gaussian upsampling in our deeper FCDD-VGG16 network. In addition, we generated a histogram of the anomaly scores of the test images in the imbalanced case with a positive ratio of $1/8$. At the bottom of Fig. \ref{fig:rawLogi}, a damage-mark explanation is presented. The red region in the heatmap represents logical defects in the MVTec logical constraint objects.  
In addition, Fig. \ref{fig:histLogi} illustrates that a few overlapping bins exist in the boundary of horizontal anomaly scores between the health class and logical defective class. 
Therefore, to detect logical defects, the score range was well-separated in the imbalanced logical constraint dataset. 
 
\begin{figure}[h] 
\centering 
\includegraphics[width=0.37\textwidth]{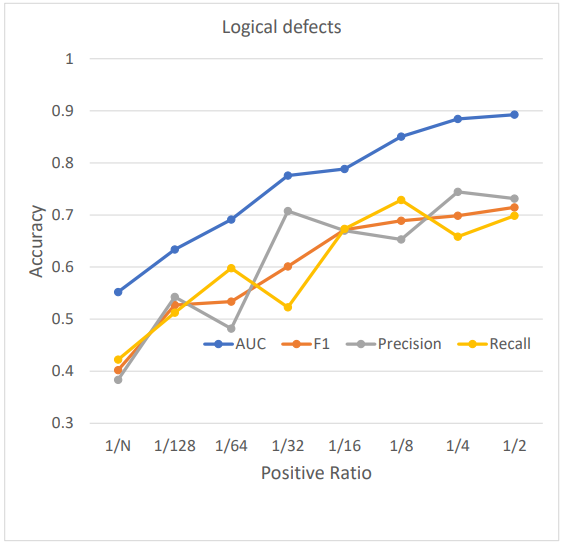} 
\caption{Anomalous vision mining studies on logical  defects indicate that the more anomalies of damage vision, the higher the performance of anomaly detection.} 
\label{fig:EffectLogi} 
\end{figure} 
 
\subsubsection{Feedback Effect on Deterioration Class Mining} 
As shown in Figure \ref{fig:EffectLogi}, from the viewpoint of the primary accuracy AUC, imbalanced studies on logical defects implied that the accuracy increased within a range of less than the positive ratio $1/8$, although the precision and recall were waving.  
We could understand that the beneficial range was damage vision mining opportunities with an accuracy gain. In contrast, when ranging with a positive ratio over $1/4$, it shifted in the over-mining phase without an effective gain in accuracy. The former phase of damage vision mining opportunities is beneficial because of the advantage of higher accuracy in terms of the AUC. 
 
\subsubsection{Embedding Damage Representation} 
\begin{figure}[h] 
\centering 
\includegraphics[width=0.35\textwidth]{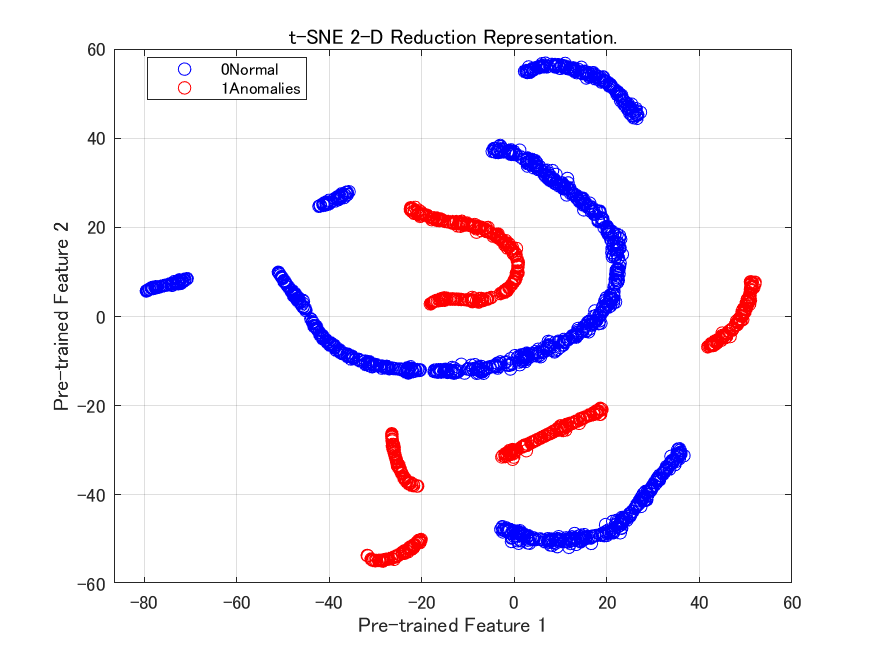} \\ 
\includegraphics[width=0.35\textwidth]{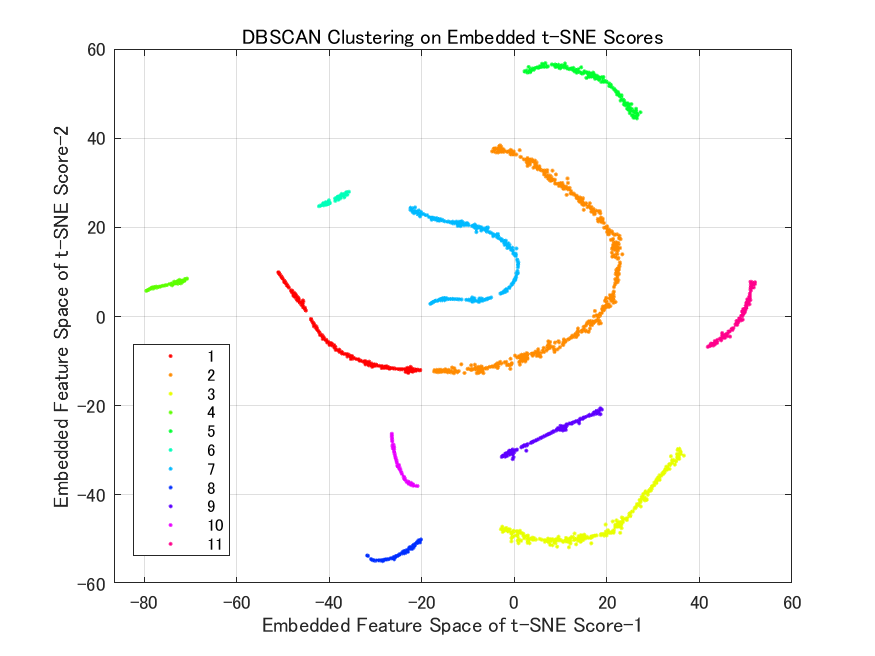} 
\caption{MN-pair contrastive damage representation (top) and density-based clustering (bottom) of logical defects.} 
\label{fig:mndbLoco} 
\end{figure}

As shown in Figure \ref{fig:mndbLoco}, we analysed the feature imbalance in the logical defect embedding space and implemented MN-pair contrastive damage representation learning and density-based clustering.  
Appropriately, the number of logical defect clusters became 11, twice the initial five classes, for example, breakfast box, juice bottle, pushipins, screw bag, and splicing connectors, which contain normal and anomalies.  
In the logical defect feature, several narrow clusters are distributed in the embedding space. 
 
\subsection{Vegetable Damage} 
\subsubsection{Backbone Studies of Supervised Detection} 
As shown in Table~\ref{tab:accBackboneVeg}, our deeper FCDD-ResNet101 outperformed in terms of the $F_1$, and recall rather than the baseline CNN27 and other backbone-based deeper FCDDs in the vegetable image dataset for the detection of old and damaged vegetables. 
\begin{table}[h] 
\caption{Backbone ablation studies on vegetable damage detection using our proposed deeper FCDDs.} 
\label{tab:accBackboneVeg} 
\centering 
\begin{tabular}{|c|c|c|c|c|} 
\hline 
\textbf{Backbone} & \textbf{AUC} & \boldmath{$F_1$} & \textbf{Precision} & \textbf{Recall} \\ 
\hline 
CNN27 & 0.9804 & 0.9150 & 0.8891 & 0.9425 \\ \hline 
VGG16 & 0.9905 & 0.9649 & 0.9673 & 0.9625 \\ 
\textbf{ResNet101} &\textbf{0.9950} & \textbf{0.9664} & \textbf{0.9604} & \textbf{0.9725} \\ 
Inceptionv3 &0.9893 & 0.9513 & 0.9501 & 0.9525 \\ \hline 
\end{tabular} 
\end{table} 
 
\subsubsection{Imbalanced-to-unsupervised Training Results} 
As shown in Table~\ref{tab:accImbalanceVeg}, we implemented ablation studies on the imbalanced data containing smaller anomalous and relatively large normal images. In this study, we applied our deeper FCDD-ResNet101 and achieved high performance in the aforementioned supervised results.   
Compared with the balanced case of a positive ratio of 1/1, we found that there was an applicable range from an imbalanced ratio of 1/2 to a ratio of 1/16, where the accuracy of recall was consistently greater than 95\%.  
However, in the extremely imbalanced range of 1/32–1/1300, the accuracy was inferior to that of the applicable range. A positive ratio of 1/32 represents imbalanced data containing 41 anomalous images and relatively large 1300 normal images. In this case, additional anomalous images were acquired and added to the initial dataset.    
The marginal gain in accuracy was relatively high when old and damaged vegetable images were added.  
\begin{table}[h] 
\caption{Imbalanced data studies using our deeper FCDD-ResNet101 for Vegetable damage detection $N_8=1300$.} 
\label{tab:accImbalanceVeg} 
\centering 
\begin{tabular}{|c|c|c|c|c|} 
\hline 
\textbf{Positive ratio} & \textbf{AUC} & \boldmath{$F_1$} & \textbf{Precision} & \textbf{Recall} \\ 
\hline 
\textbf{1/1(ano.$N_8$)} & \textbf{0.9950} & \textbf{0.9664} & \textbf{0.9604} & \textbf{0.9725} \\ \hline 
1/2(ano.650) & 0.9949 & 0.9568 & 0.9716 & 0.9425 \\  
1/4(ano.325) & 0.9954 & 0.9638 & 0.9603 & 0.9675 \\  
1/8(ano.163) & 0.9887 & 0.9536 & 0.9548 & 0.9525 \\  
1/16(ano.81) & 0.9838 & 0.9407 & 0.9292 & 0.9525 \\ \hline 
\textbf{1/32(ano.41)}&\textbf{0.9771} & \textbf{0.9431} & \textbf{0.9539} & \textbf{0.9325} \\  
\textbf{1/64(ano.20)}&\textbf{0.9722} & \textbf{0.9264} & \textbf{0.9407} & \textbf{0.9125} \\  
\textbf{1/128(ano.10)}&\textbf{0.9680} & \textbf{0.9118} & \textbf{0.9061} & \textbf{0.9175} \\  
\textbf{1/$N_8$(ano.1)} &\textbf{0.7886} & \textbf{0.6952} & \textbf{0.6810} & \textbf{0.7100} \\ \hline 
\end{tabular} 
\end{table} 
 
\begin{figure}[h] 
\centering 
\includegraphics[width=0.4\textwidth]{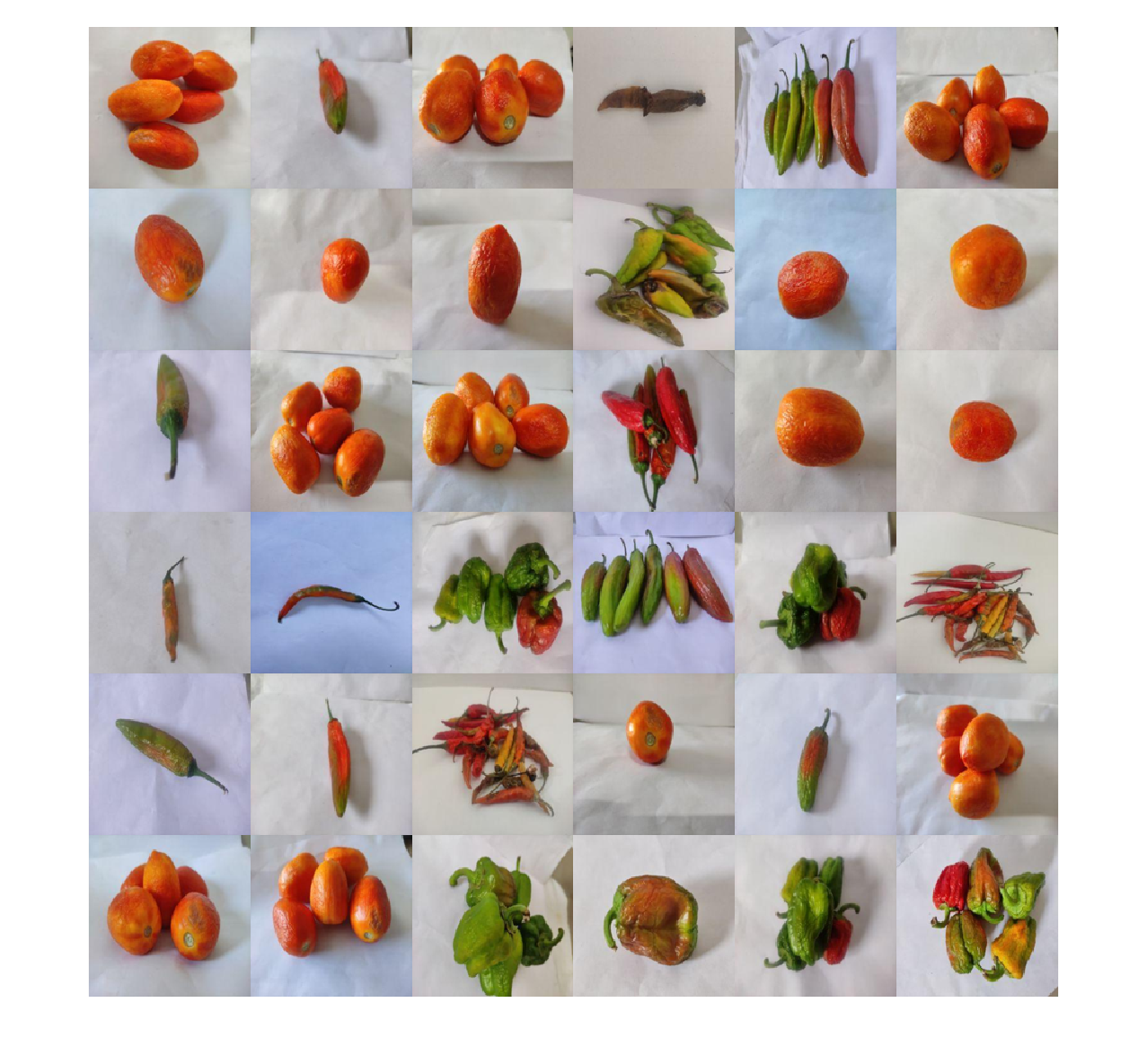} \\ 
\includegraphics[width=0.4\textwidth]{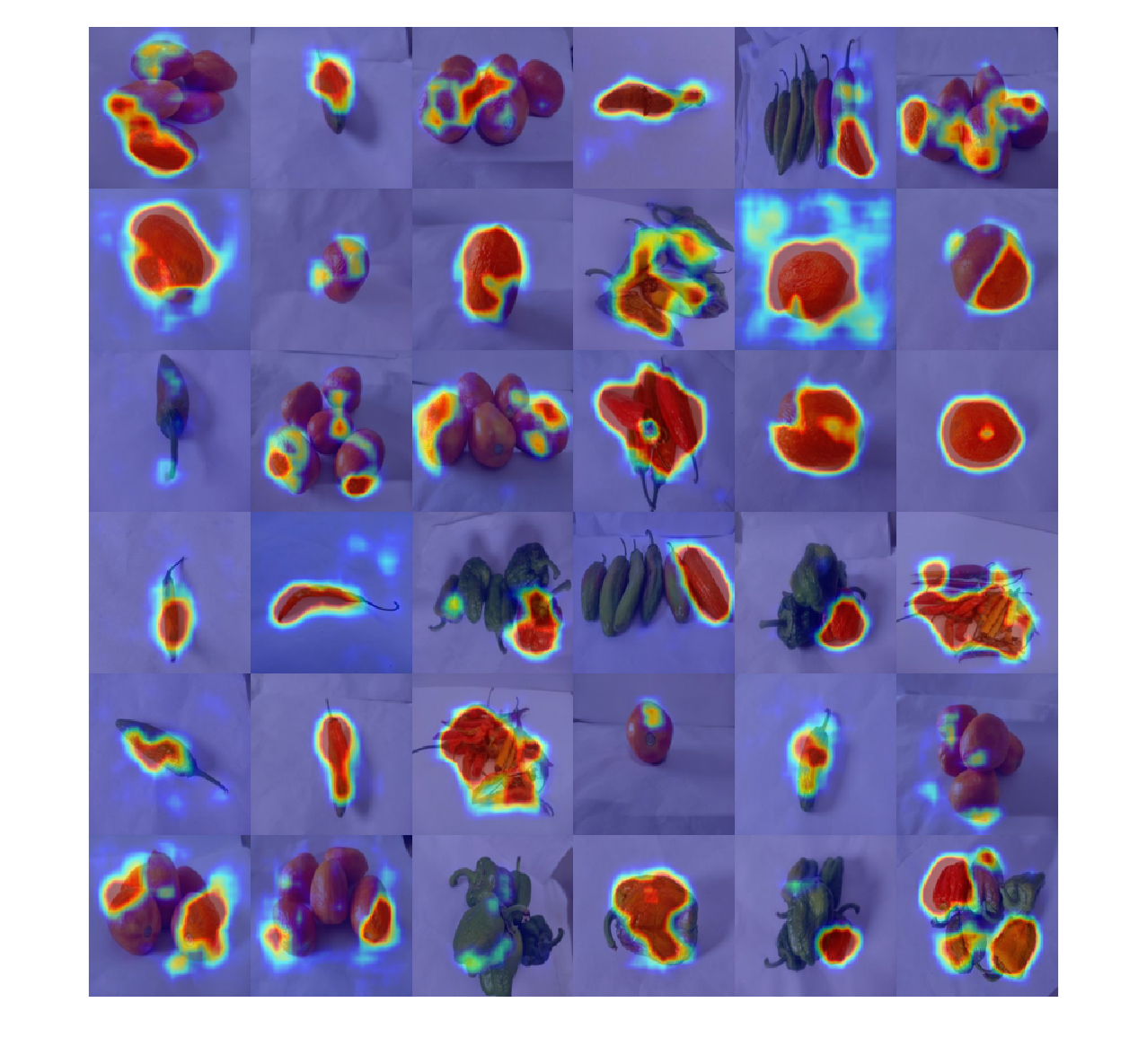} 
\caption{Imbalanced damage images (top) with positive ratio $1/16$, and damage-mark heatmaps (bottom) of vegetable damage using our deeper FCDD-ResNet101.} 
\label{fig:rawVeg} 
\end{figure} 
\begin{figure}[h] 
\centering 
\includegraphics[width=0.37\textwidth]{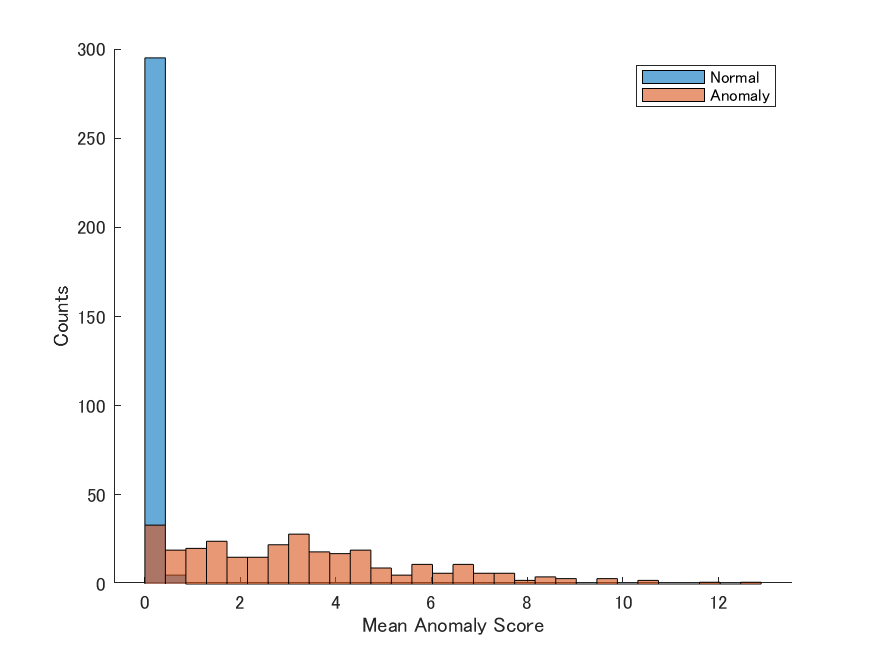} 
\caption{Histogram of vegetable damage scores using our deeper FCDD-ResNet101 corresponding to the imbalanced damage images with positive ratio $1/16$.} 
\label{fig:histVeg} 
\end{figure} 
 
\begin{figure}[h] 
\centering 
\includegraphics[width=0.37\textwidth]{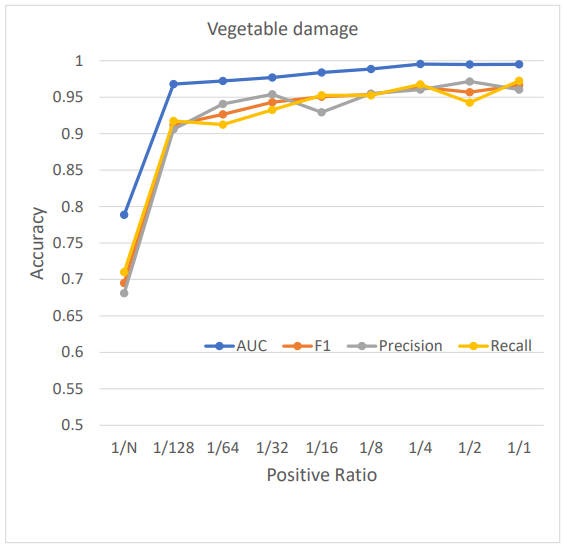} 
\caption{Anomalous vision mining studies on vegetable damage, indicates that the more anomalies of damage vision, the higher performance of anomaly detection.} 
\label{fig:EffectVeg}  
\end{figure} 

 \begin{figure}[h] 
\centering 
\includegraphics[width=0.35\textwidth]{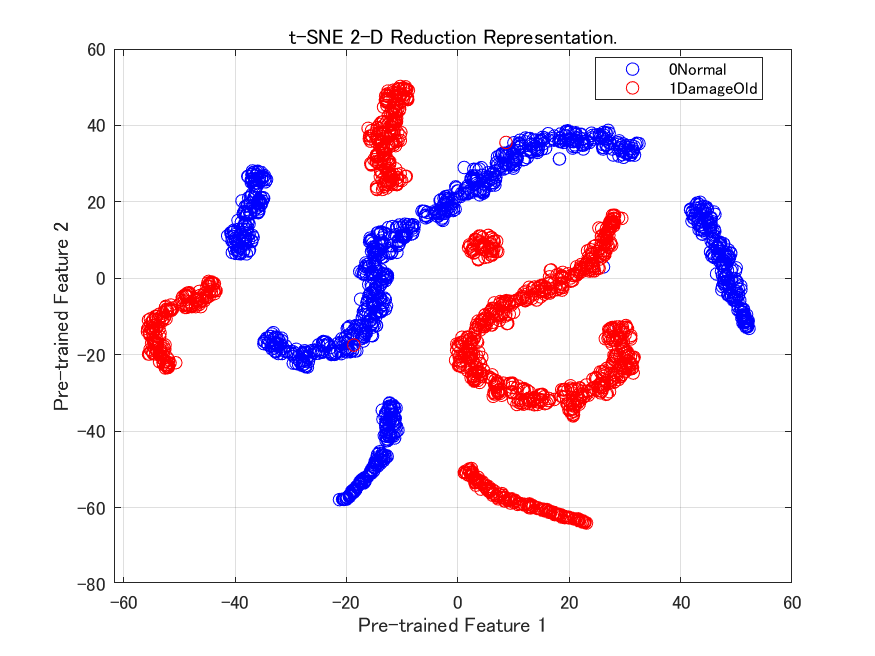} \\ 
\includegraphics[width=0.35\textwidth]{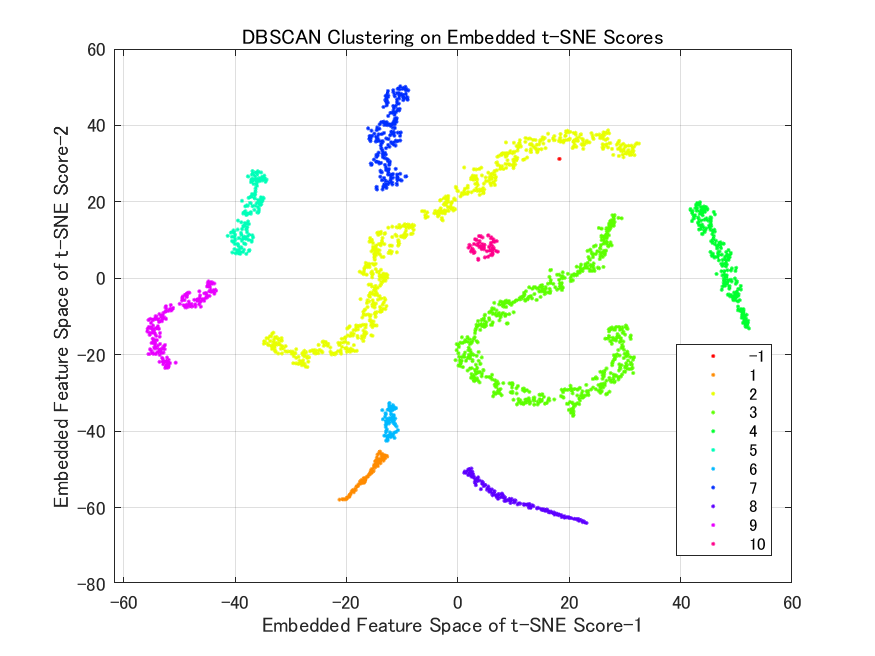} 
\caption{MN-pair contrastive damage representation (top) and density-based clustering (bottom) of vegetable damage.} 
\label{fig:mndbVeg} 
\end{figure} 

\subsubsection{Damage-mark Heatmaps on Vegetable Damage} 
We visualised the damage features using Gaussian upsampling in our deeper FCDD-ResNet101 network. In addition, we generated a histogram of the anomaly scores of the test images in the imbalanced case with a positive ratio of $1/16$. In Fig. \ref{fig:rawVeg}, a damage-mark explanation is represented. The red region in the heatmap represents the old and damaged features in the vegetable image.  
Fig. \ref{fig:histVeg} illustrates that a few overlapping bins exist in the boundary of the normal class and vegetable damage class along the horizontal anomaly scores. Thus, to detect old and damaged vegetables, the score range was well separated in the vegetable damage image dataset. 
 
\subsubsection{Feedback Effect on Damage Class Mining} 
As shown in Figure \ref{fig:EffectVeg}, from the viewpoint of accuracy, the imbalanced studies on vegetable damage imply that all accuracies consistently converged into significant phases.  
Ranging with a positive ratio of less than $1/16$, we could understand that there were damage vision mining opportunities with an accuracy gain in terms of the AUC and recall. In contrast, when ranged with a positive ratio of over $1/8$, it shifted to the over-mining phase without any gain in the AUC and recall. The former phase of damage vision mining opportunities has become beneficial owing to its promising advantage of higher accuracy. 
 
\subsubsection{Embedding Damage Representation} 
As shown in Figure \ref{fig:mndbVeg}, we analysed the feature imbalance in the vegetable damage embedding space and implemented MN-pair contrastive damage representation learning and density-based clustering.  
The number of vegetable damage clusters increased by 10 more than the initial four classes, that is, tomato, bell pepper, chili pepper, and New Mexico chili, which included normal and anomalies. Regarding the vegetable-damaged feature, several narrow clusters were distributed in the embedding space. 
 
\subsection{Plant Infection} 
\subsubsection{Backbone Studies of Supervised Detection} 
As shown in Table~\ref{tab:accBackbonePlant}, our deeper FCDD-Inceptionv3 outperformed in terms of $F_1$, precision rather than the baseline CNN27 and other backbone-based deeper FCDDs in the plant disease image dataset for detecting infected leaves. 
\begin{table}[h] 
\caption{Backbone ablation studies on infected leaves detection using our proposed deeper FCDDs.} 
\label{tab:accBackbonePlant} 
\centering 
\begin{tabular}{|c|c|c|c|c|} 
\hline 
\textbf{Backbone} & \textbf{AUC} & \boldmath{$F_1$} & \textbf{Precision} & \textbf{Recall} \\ 
\hline 
CNN27 & 0.9914 & 0.9439 & 0.9404 & 0.9475 \\ \hline 
VGG16 & 0.9999 & 0.9900 & 0.9851 & 0.9950 \\ 
ResNet101 & 0.9999 & 0.9888 & 0.9827 & 0.9950 \\ 
\textbf{Inceptionv3} & \textbf{0.9997} & \textbf{0.9912} & \textbf{0.9924} & \textbf{0.9900} \\ \hline 
\end{tabular} 
\end{table} 
 
\subsubsection{Imbalanced-to-unsupervised Training Results} 
As shown in Table~\ref{tab:accImbalancePlant}, we implemented ablation studies on the imbalanced data containing smaller anomalous and relatively large normal images. In this study, we applied our deeper FCDD-Inceptionv3 and achieved high performance in the aforementioned supervised results.   
Compared with the balanced case of a positive ratio of 1/1, we found that there was an applicable range from an imbalanced ratio of 1/2 to 1/8, where the accuracy loss of recall was consistently less than 3\%.  
However, in the extremely imbalanced range of 1/16 to 1/1300, the accuracy was inferior to the applicable range, that is, the loss of recall was greater than 3\%. The rare positive ratio of 1/16 represents imbalanced data that contain a small number of anomalous images (81) and relatively large normal images (1300). In this case, additional anomalous images were acquired and added to the initial dataset.    
The marginal gain in accuracy was relatively high when infected leaf images were added.  
\begin{table}[h] 
\caption{Imbalanced data studies using our deeper FCDD-Inceptionv3 for Plant infected leaves detection $N_9=1300$.} 
\label{tab:accImbalancePlant} 
\centering 
\begin{tabular}{|c|c|c|c|c|} 
\hline 
\textbf{Positive ratio} & \textbf{AUC} & \boldmath{$F_1$} & \textbf{Precision} & \textbf{Recall} \\ 
\hline 
\textbf{1/1(ano.$N_9$)} & \textbf{0.9997} & \textbf{0.9912} & \textbf{0.9924} & \textbf{0.9900} \\ \hline 
1/2(ano.650) & 0.9989 & 0.9912 & 0.9949 & 0.9875 \\  
1/4(ano.325) & 0.9981 & 0.9861 & 0.9949 & 0.9775 \\  
1/8(ano.163) & 0.9983 & 0.9848 & 0.9949 & 0.9750 \\ \hline  
\textbf{1/16(ano.81)}&\textbf{0.9988} & \textbf{0.9709} & \textbf{0.9821} & \textbf{0.9600} \\ 
\textbf{1/32(ano.41)}&\textbf{0.9981} & \textbf{0.9722} & \textbf{0.9797} & \textbf{0.9650} \\  
\textbf{1/64(ano.20)}&\textbf{0.9936} & \textbf{0.9533} & \textbf{0.9618} & \textbf{0.9450} \\  
\textbf{1/128(ano.10)}&\textbf{0.9826} & \textbf{0.9023} & \textbf{0.9553} & \textbf{0.8550} \\  
\textbf{1/$N_9$(ano.1)} &\textbf{0.7882} & \textbf{0.6943} & \textbf{0.7770} & \textbf{0.6275} \\ \hline 
\end{tabular} 
\end{table} 
 
\begin{figure}[h] 
\centering 
\includegraphics[width=0.4\textwidth]{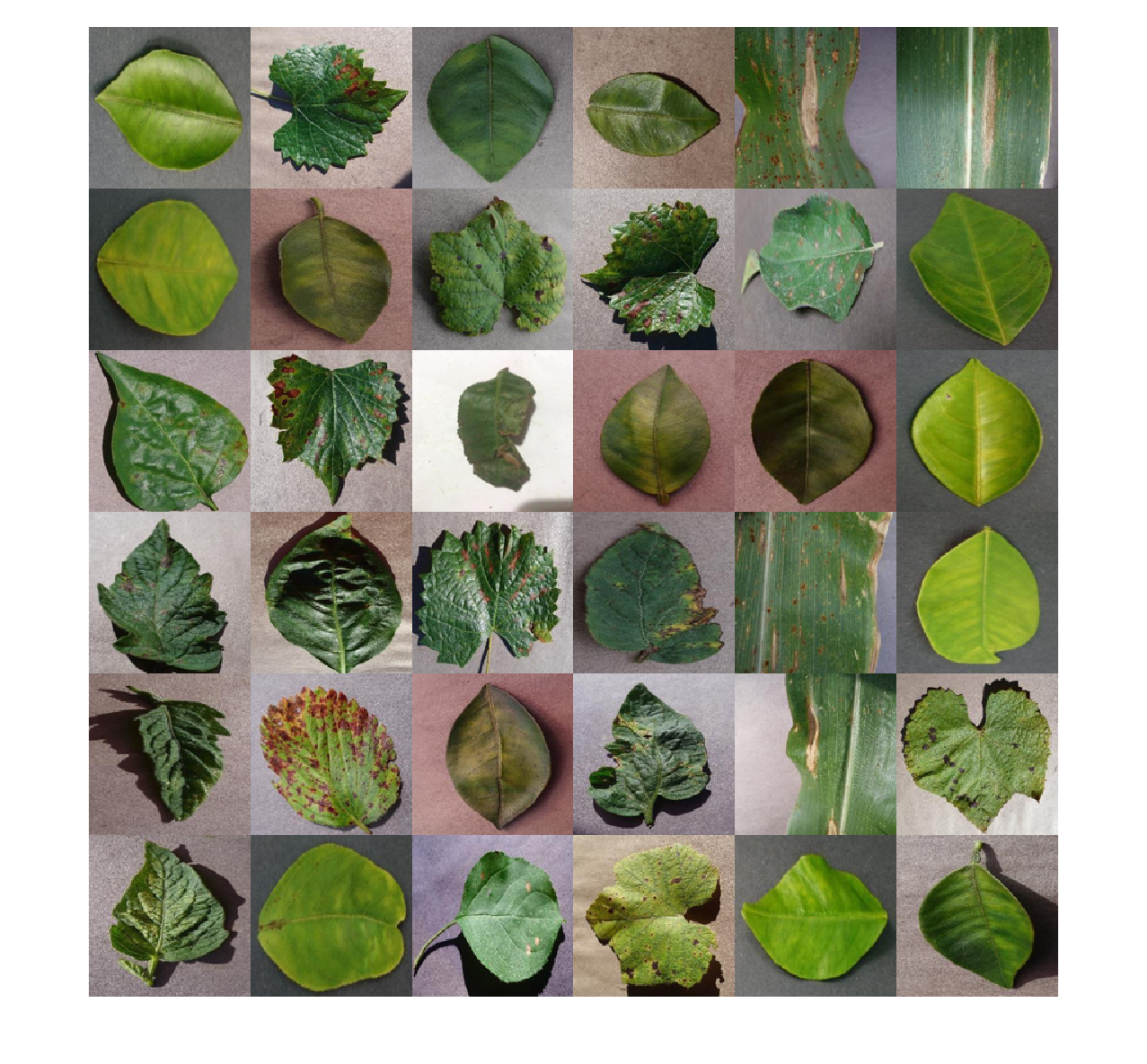} \\ 
\includegraphics[width=0.4\textwidth]{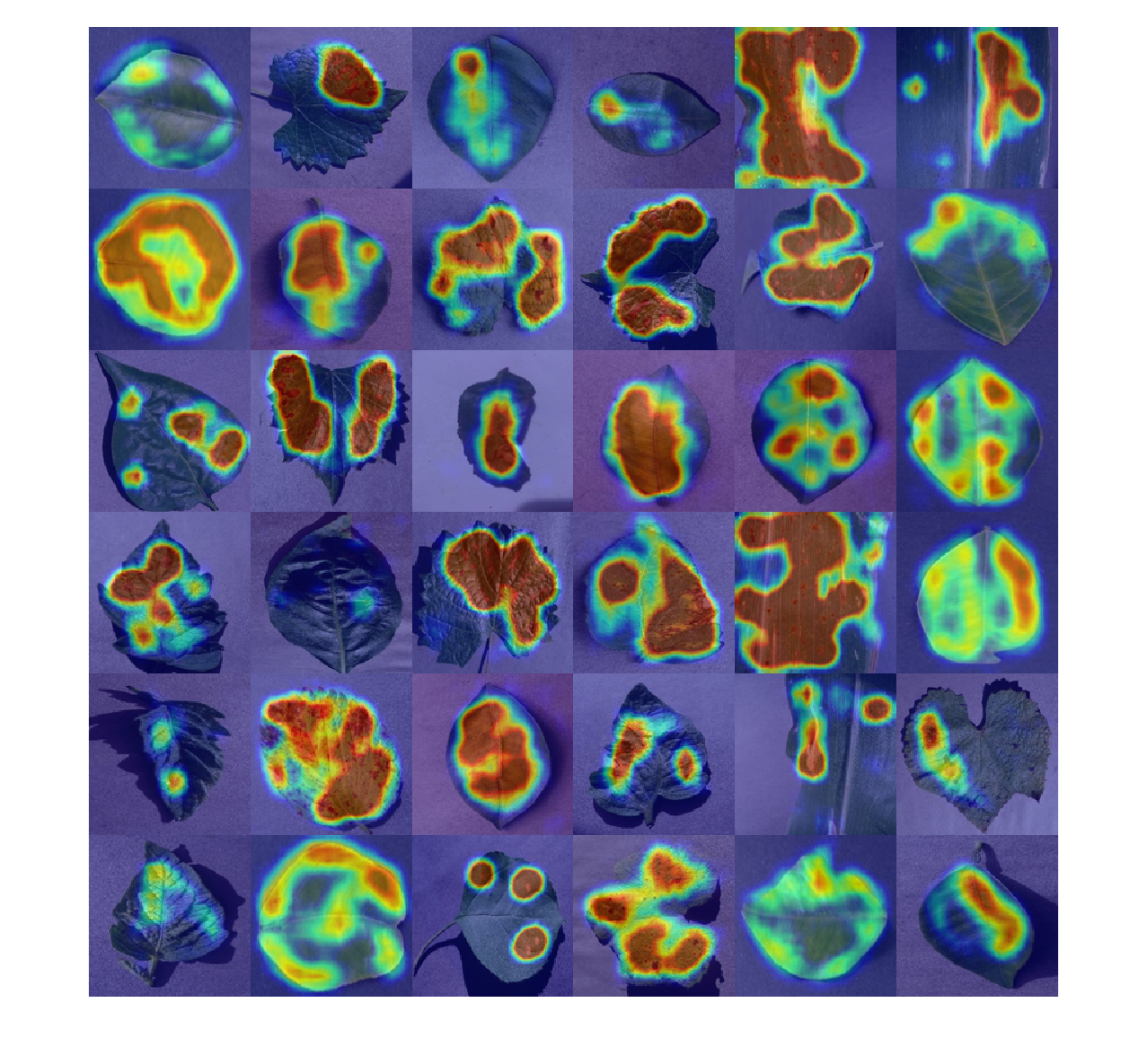} 
\caption{Imbalanced damage images (top) with positive ratio $1/8$, and damage-mark heatmaps (bottom) of plant leaves infection using our deeper FCDD-Inceptionv3.} 
\label{fig:rawPlant} 
\end{figure} 
\begin{figure}[h] 
\centering 
\includegraphics[width=0.37\textwidth]{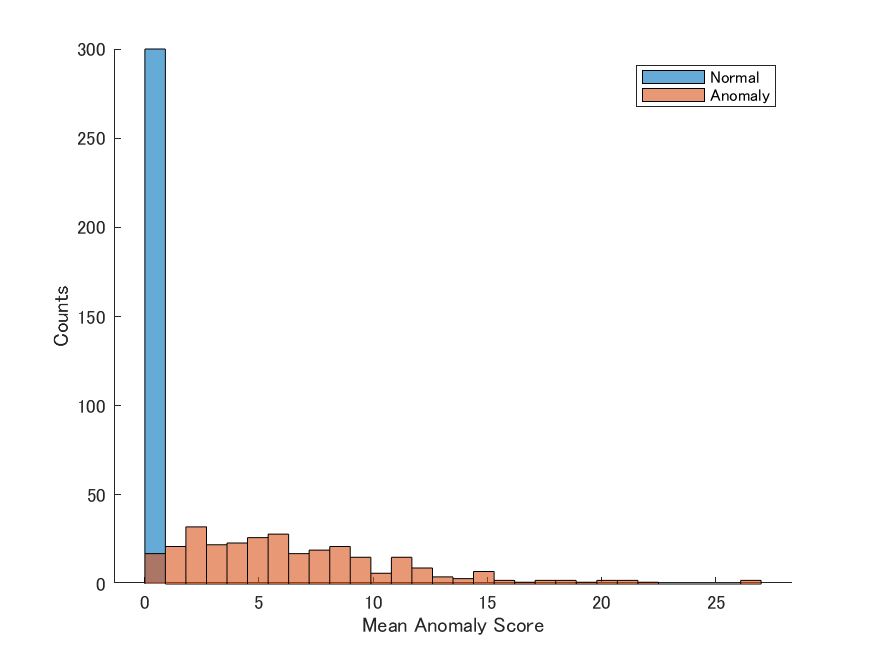} 
\caption{Histogram of plant leaves infection scores using our deeper FCDD-Inceptionv3 corresponding to the imbalanced damage images with positive ratio $1/8$.} 
\label{fig:histPlant} 
\end{figure} 
 
\begin{figure}[h] 
\centering 
\includegraphics[width=0.37\textwidth]{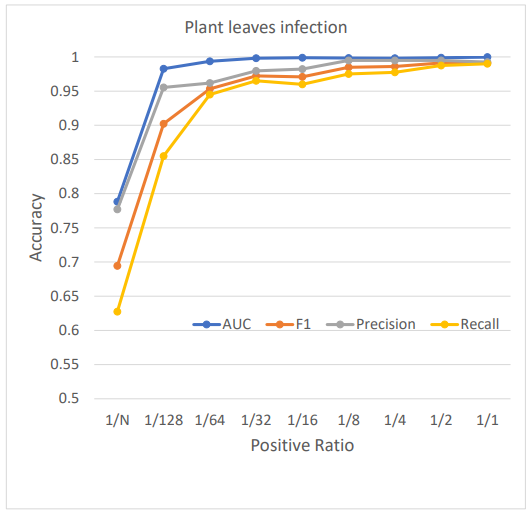} 
\caption{Anomalous vision mining studies on plant leaves  infection indicate that the more anomalies of damage vision, the higher the performance of anomaly detection.} 
\label{fig:EffectPlant}  
\end{figure} 
 
\subsubsection{Damage-mark Heatmaps on Plant Disease} 
We visualised the damage features using Gaussian upsampling in our deeper FCDD-Inceptionv3 network. In addition, we generated a histogram of the anomaly scores of the test images in the imbalanced case with a positive ratio of $1/8$. In Fig. \ref{fig:rawPlant}, a damage-mark explanation is represented. The red region in the heatmap represents infected leaves.   
Fig. \ref{fig:histPlant} illustrates that a few overlapping bins exist in the boundary of the health leaves class and plant disease class along the horizontal anomaly scores. Thus, to detect infected leaves, the score range was well separated in the plant disease image dataset. 
 
\subsubsection{Feedback Effect on Damage Class Mining} 
In Figure \ref{fig:EffectPlant}, from the viewpoint of accuracy, the imbalanced studies on plant leaf infections implied that all accuracies consistently converged into significant phases.  
Ranging with a positive ratio of less than $1/8$, we could understand that there were damage vision mining opportunities with an accuracy gain in terms of recall. In contrast, ranging with a positive ratio of over $1/4$, it shifted in the over-mining phase without any recall gain. The former phase of damage vision mining opportunities has become beneficial owing to its promising advantage of higher accuracy. 
 
\subsubsection{Embedding Damage Representation} 
\begin{figure}[h] 
\centering 
\includegraphics[width=0.35\textwidth]{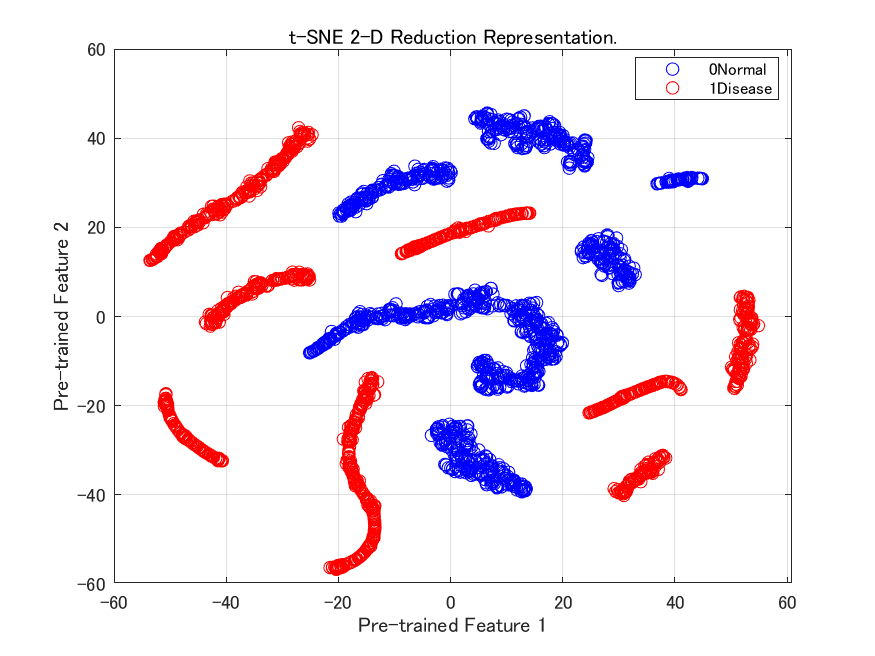} \\ 
\includegraphics[width=0.35\textwidth]{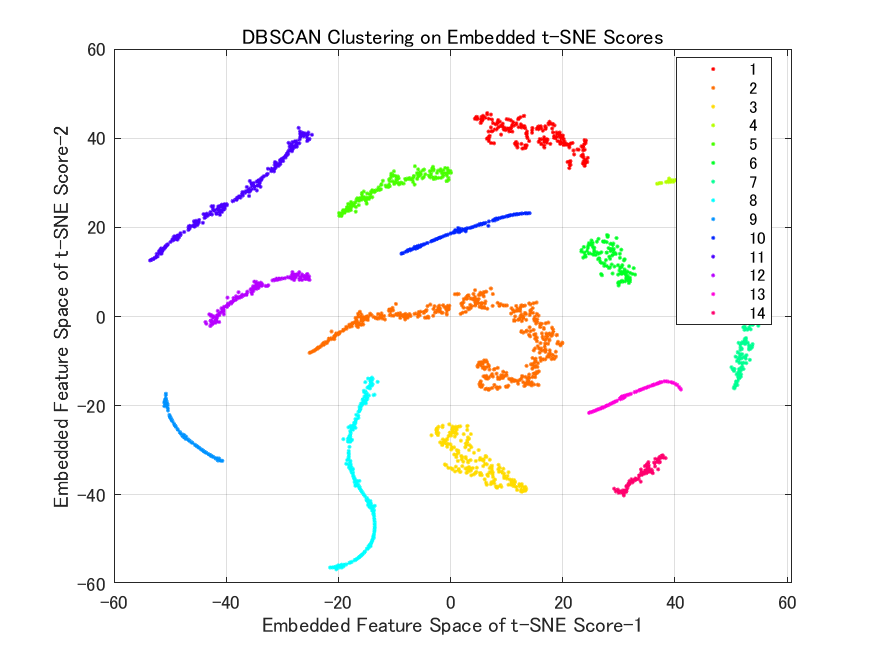} 
\caption{MN-pair contrastive damage representation (top) and density-based clustering (bottom) of plant disease.} 
\label{fig:mndbPlant} 
\end{figure}

As shown in Figure \ref{fig:mndbPlant}, we analysed feature imbalance in the plant disease embedding space and implemented MN-pair contrastive damage representation and density-based clustering.  
The number of plant disease clusters decreased to 14, less than twice the initial 12 classes, such as apple, blueberry, cherry, corn, grape, peach, potato, raspberry, soybean, strawberry, and tomato, which included normal and anomalies.  
For the plant disease features, several narrow clusters were distributed in the embedding space. 
 
\subsection{River Sludge} 
\subsubsection{Backbone Studies of Supervised Detection} 
As shown in Table~\ref{tab:accBackboneRiver}, our deeper FCDD-ResNet101 outperformed in terms of the $F_1$, precision, and recall rather than the baseline CNN27 and other backbone-based deeper FCDDs in this river surface image dataset for detecting river sludge floating on the surface. 
\begin{table}[h] 
\caption{Backbone ablation studies on river sludge detection using our proposed deeper FCDDs.} 
\label{tab:accBackboneRiver} 
\centering 
\begin{tabular}{|c|c|c|c|c|} 
\hline 
\textbf{Backbone} & \textbf{AUC} & \boldmath{$F_1$} & \textbf{Precision} & \textbf{Recall} \\ 
\hline 
CNN27 & 0.9518 & 0.8567 & 0.8940 & 0.8225 \\ \hline 
VGG16 & 0.9666 & 0.9065 & 0.9291 & 0.8850 \\ 
\textbf{ResNet101} &\textbf{0.9681} & \textbf{0.9214} & \textbf{0.9496} & \textbf{0.8950} \\ 
Inceptionv3 &0.9623 & 0.9010 & 0.9402 & 0.8650 \\ \hline 
\end{tabular} 
\end{table} 
 
\subsubsection{Imbalanced-to-unsupervised Training Results} 
As shown in Table~\ref{tab:accImbalanceRiver}, we conducted ablation studies on the imbalanced data containing smaller anomalous and relatively large normal images. In this study, we applied our deeper FCDD-ResNet101 and achieved high performance in the aforementioned supervised results.   
Compared with the balanced case of a positive ratio of 1/1, we found that there was an applicable range from an imbalanced ratio of 1/2 to 1/4, where the accuracy of $F_1$ was consistently greater than 90\%.  
However, in the extremely imbalanced range of 1/8–1/1300, the accuracy was inferior to that of the applicable range. A positive ratio of 1/8 represents imbalanced data containing 163 anomalous images and relatively large 1300 normal images. In this case, additional anomalous images were acquired and added to the initial dataset.    
The marginal gain in accuracy was relatively high when the river sludge images were added to the surface.  
\begin{table}[h] 
\caption{Imbalanced data studies using our deeper FCDD-ResNet101 for River sludge  detection $N_6=1300$.} 
\label{tab:accImbalanceRiver} 
\centering 
\begin{tabular}{|c|c|c|c|c|} 
\hline 
\textbf{Positive ratio} & \textbf{AUC} & \boldmath{$F_1$} & \textbf{Precision} & \textbf{Recall} \\ 
\hline 
\textbf{1/1(ano.$N_{10}$)} & \textbf{0.9681} & \textbf{0.9214} & \textbf{0.9496} & \textbf{0.8950} \\ \hline 
1/2(ano.650) & 0.9635 & 0.9175 & 0.9468 & 0.8900 \\  
1/4(ano.325) & 0.9457 & 0.9077 & 0.8934 & 0.9225 \\ \hline  
\textbf{1/8(ano.163)} & \textbf{0.9588} & \textbf{0.8736} & \textbf{0.9222} & \textbf{0.8300} \\  
\textbf{1/16(ano.81)}& \textbf{0.9428} & \textbf{0.8579} & \textbf{0.9150} & \textbf{0.8075} \\ 
\textbf{1/32(ano.41)}&\textbf{0.9267} & \textbf{0.8382} & \textbf{0.8686} & \textbf{0.8100} \\  
\textbf{1/64(ano.20)}&\textbf{0.9131} & \textbf{0.8175} & \textbf{0.8746} & \textbf{0.7675} \\  
\textbf{1/128(ano.10)}&\textbf{0.9123} & \textbf{0.7973} & \textbf{0.8676} & \textbf{0.7375} \\  
\textbf{1/$N_{10}$(ano.1)} &\textbf{0.8221} & \textbf{0.7139} & \textbf{0.7513} & \textbf{0.6800} \\ \hline 
\end{tabular} 
\end{table} 
 
\begin{figure}[h] 
\centering 
\includegraphics[width=0.4\textwidth]{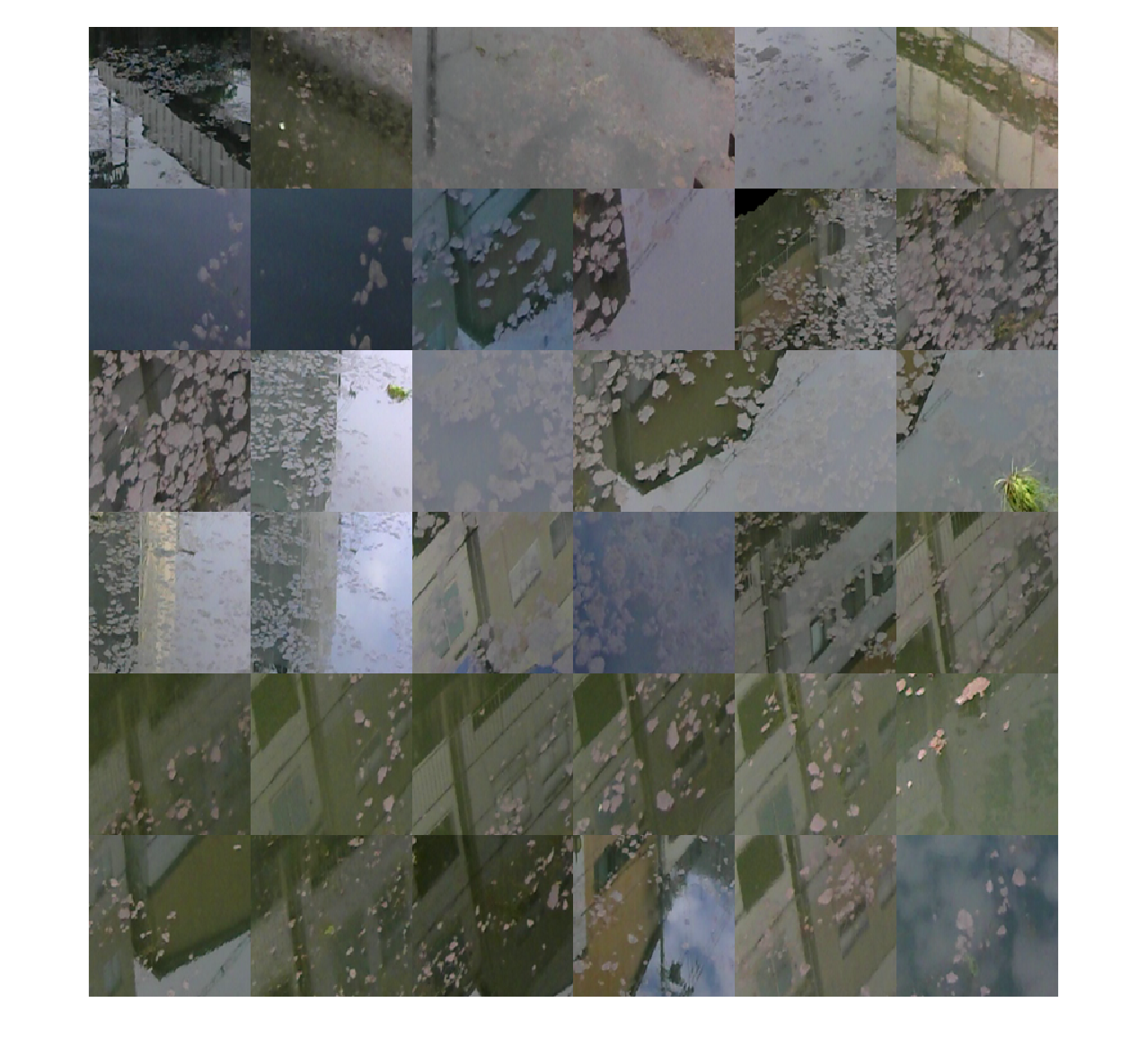} \\ 
\includegraphics[width=0.4\textwidth]{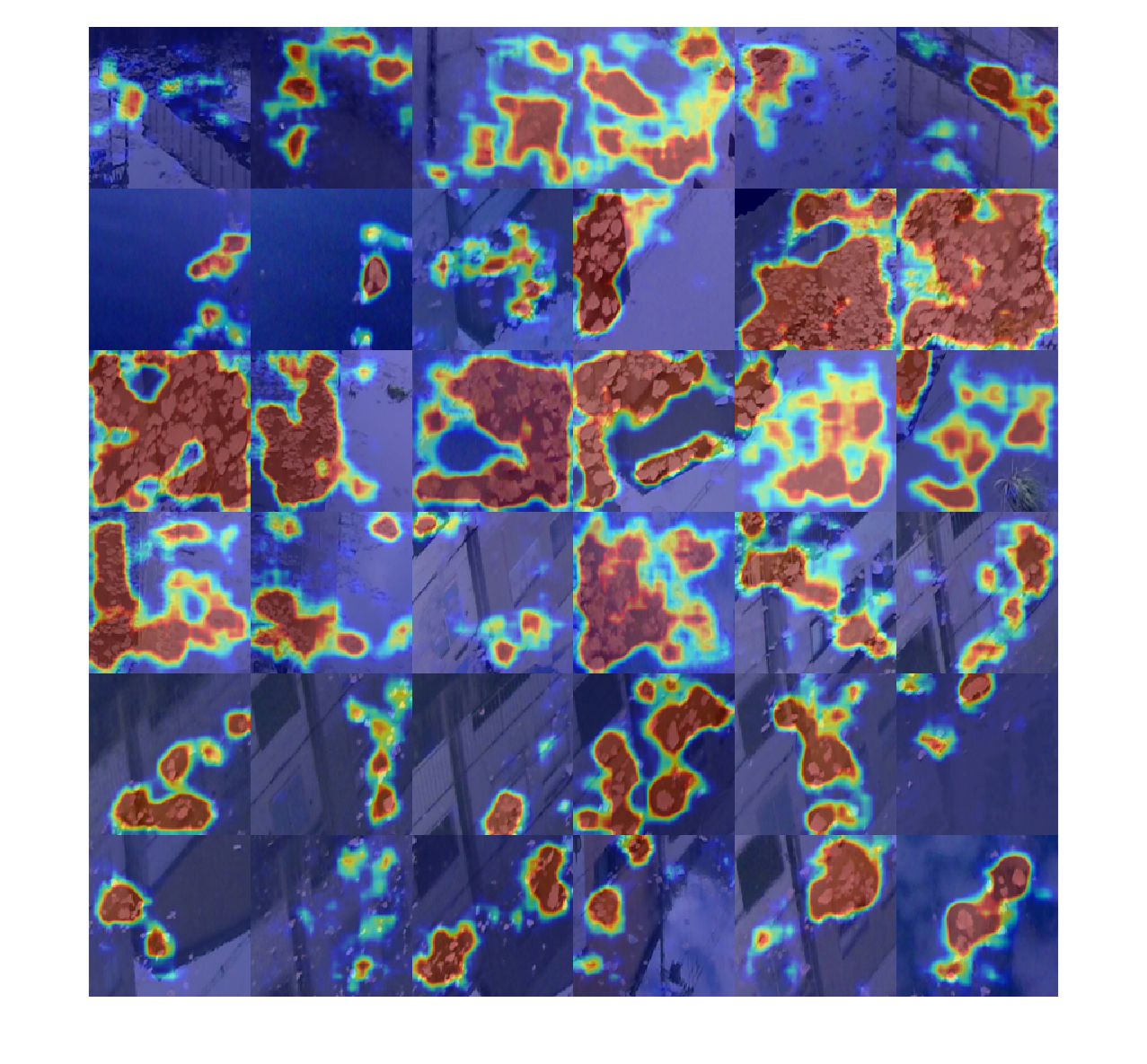} 
\caption{Imbalanced damage images (top) with positive ratio $1/4$, and damage-mark heatmaps (bottom) of river sludge using our deeper FCDD-ResNet101.} 
\label{fig:rawRiver} 
\end{figure} 
\begin{figure}[h] 
\centering 
\includegraphics[width=0.37\textwidth]{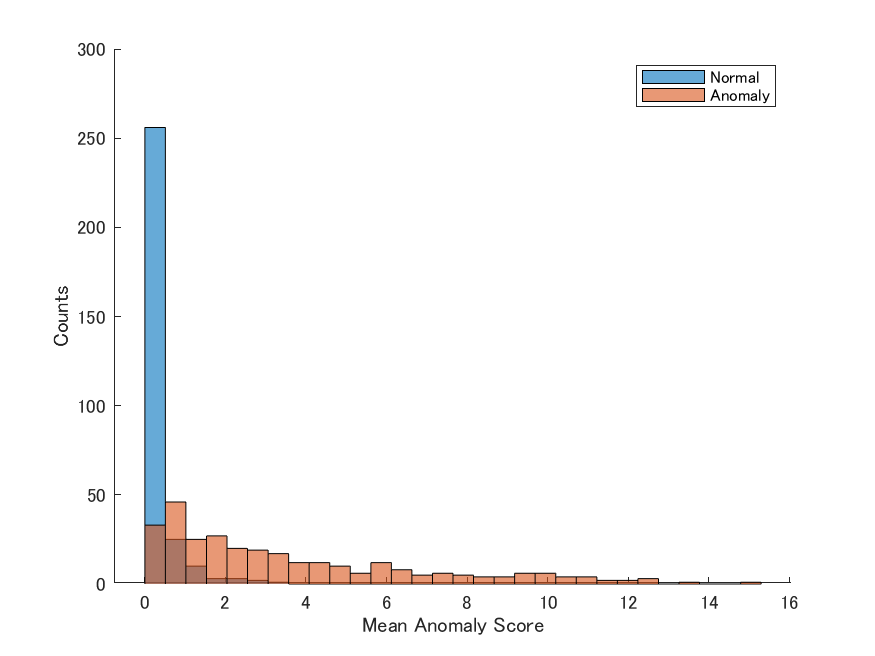} 
\caption{Histogram of river sludge scores using our deeper FCDD-ResNet101 corresponding to the imbalanced damage images with positive ratio $1/4$.} 
\label{fig:histRiver} 
\end{figure} 
 
\begin{figure}[h] 
\centering 
\includegraphics[width=0.37\textwidth]{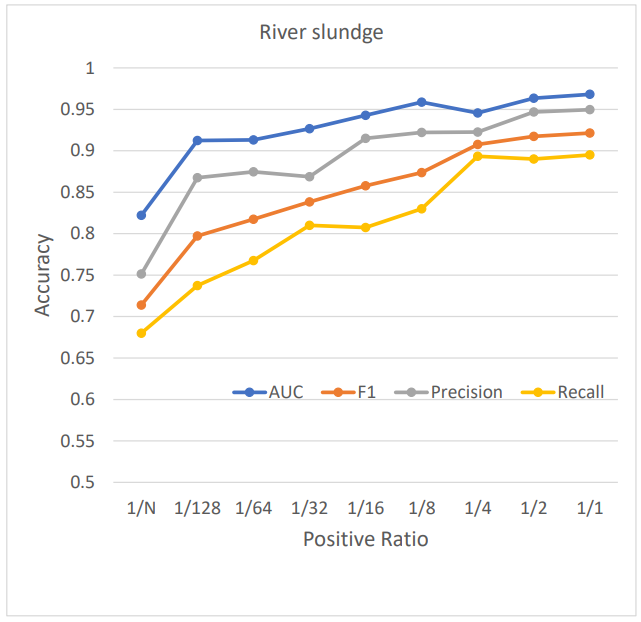} 
\caption{Anomalous vision mining studies on river sludge indicate that the more anomalies of damage vision, the higher the performance of anomaly detection.} 
\label{fig:EffectRiver}  
\end{figure} 
 
\subsubsection{Damage-mark Heatmaps on River Sludge} 
We visualised the damage features using Gaussian upsampling in our deep FCDD-ResNet101 network. In addition, we generated a histogram of the anomaly scores of the test images in the imbalanced case with a positive ratio of $1/4$. In Fig. \ref{fig:rawRiver}, a damage-mark explanation is represented. The red region in the heatmap represents the river sludge features in the river surface images.  
Fig. \ref{fig:histRiver} illustrates that a few overlapping bins exist in the boundary of the normal class and river sludge class along the horizontal anomaly scores. Thus, to detect river sludge on the surface, the score range was well separated in the river sludge image dataset. 
 
\subsubsection{Feedback Effect on Damage Class Mining} 
As shown in Figure \ref{fig:EffectRiver}, from the viewpoint of accuracy, imbalanced studies on river sludge imply that all accuracies consistently converge into significant phases.  
Ranging with a positive ratio of less than $1/8$, we could understand that there were damage vision mining opportunities with an accuracy gain in terms of the AUC. In contrast, ranging with a positive ratio of over $1/4$, it shifted in the over-mining phase without any gain in the AUC. The former phase of damage vision mining opportunities has become beneficial owing to its promising advantage of higher accuracy. 
 
\subsubsection{Embedding Damage Representation} 
\begin{figure}[h] 
\centering 
\includegraphics[width=0.35\textwidth]{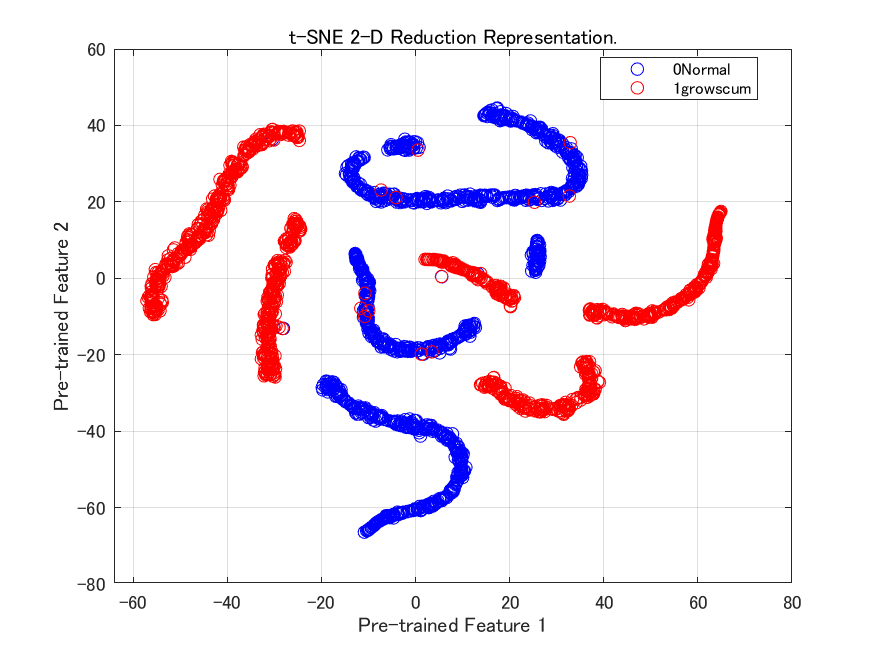} \\ 
\includegraphics[width=0.35\textwidth]{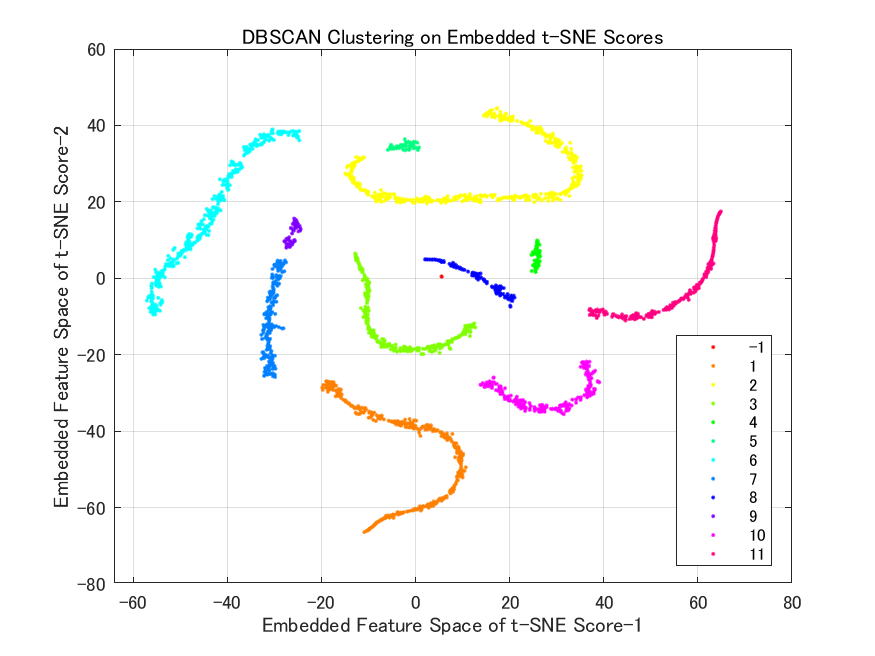} 
\caption{MN-pair contrastive damage representation (top) and density-based clustering (bottom) of river sludge.} 
\label{fig:mndbRiver} 
\end{figure}

As shown in Figure \ref{fig:mndbRiver}, we analysed the feature imbalance in the river sludge embedding space and implemented MN-pair contrastive damage representation learning and density-based clustering.  
Appropriately, the number of river sludge clusters increased to 11, rather than the initial two classes that contained normal river surface and variational sludge.  
Regarding the river sludge features, a few narrow clusters were distributed in the embedding space. 
 
\begin{figure}[h] 
\centering 
\includegraphics[width=0.4\textwidth]{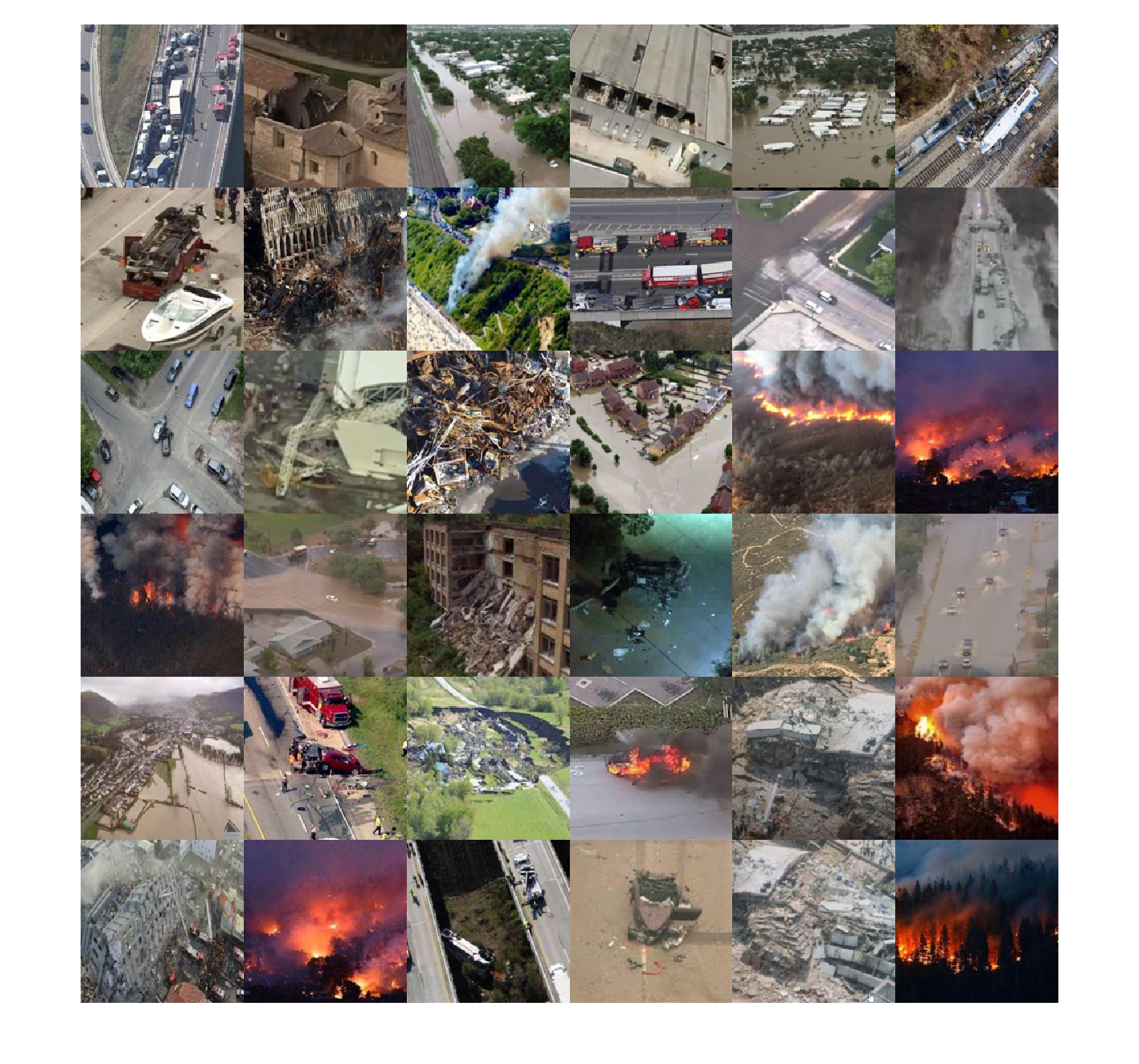} \\ 
\includegraphics[width=0.4\textwidth]{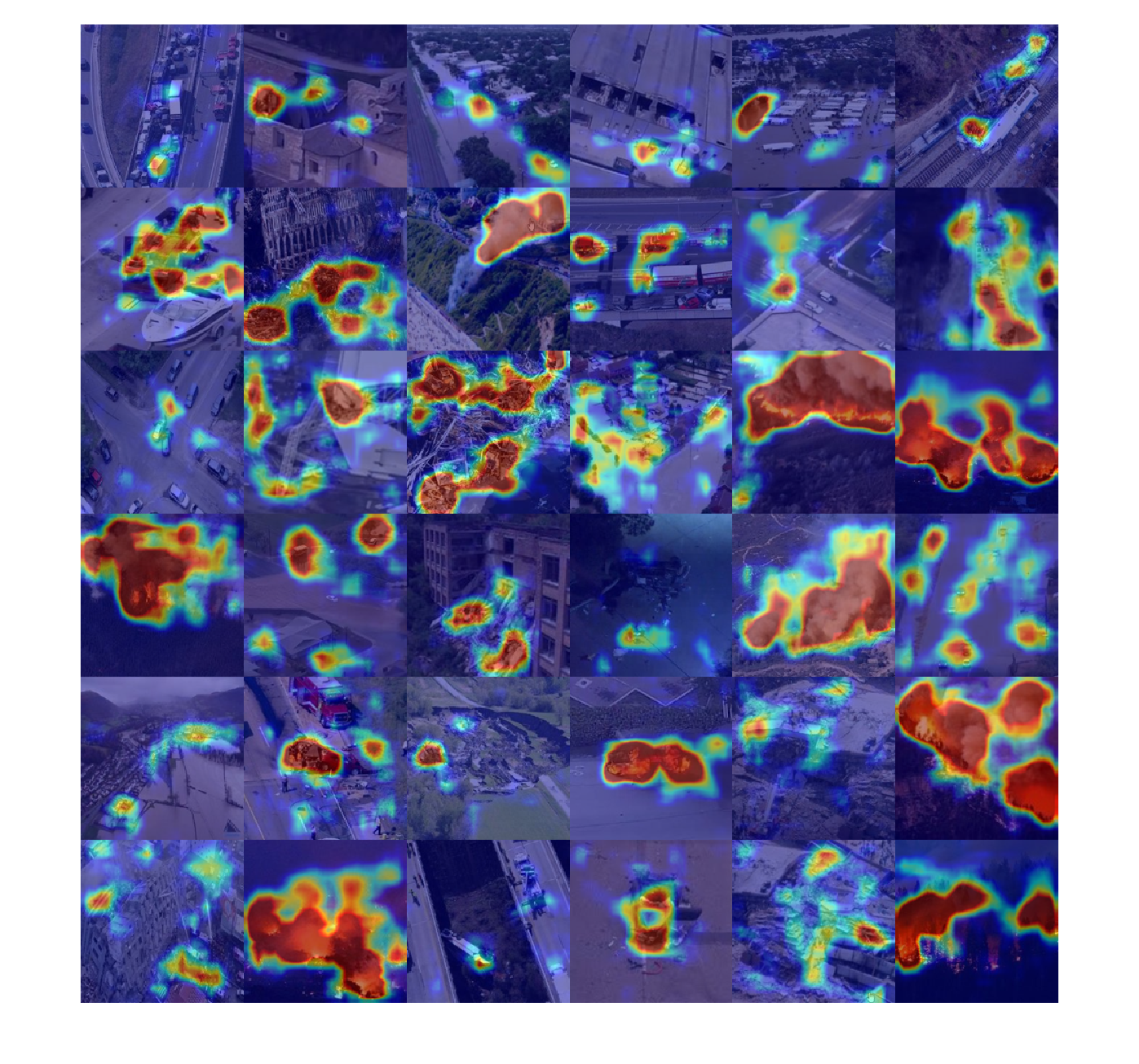} 
\caption{Imbalanced damage images (top) with positive ratio $1/16$, and damage-mark heatmaps (bottom) of disaster damage using our deeper FCDD-VGG16.} 
\label{fig:rawBuild} 
\end{figure} 
\begin{figure}[h] 
\centering 
\includegraphics[width=0.37\textwidth]{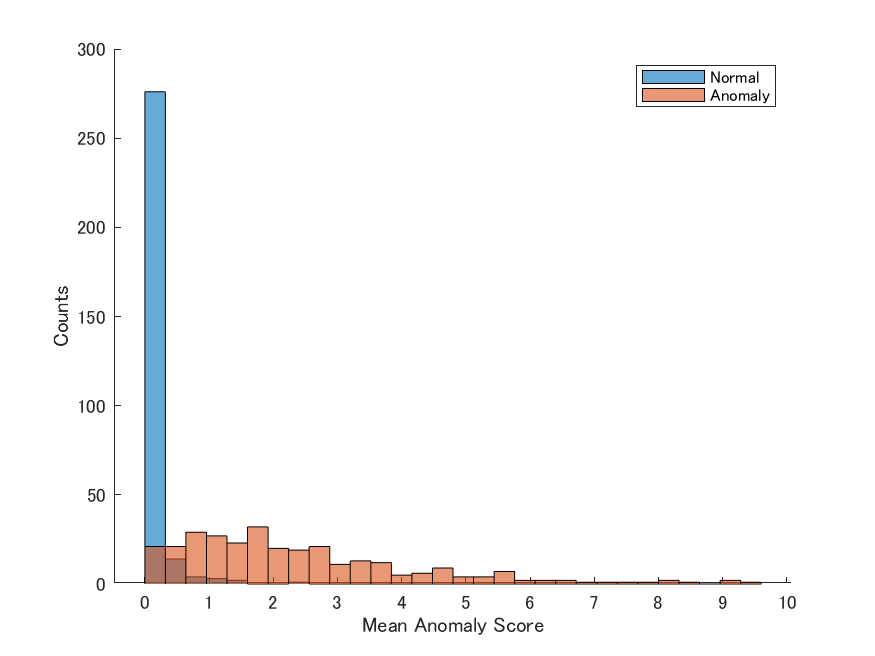} 
\caption{Histogram of disaster damage scores using our deeper FCDD-VGG16 corresponding to the imbalanced damage images with positive ratio $1/16$.} 
\label{fig:histBuild} 
\end{figure} 
 
\subsection{Disaster Damage} 
\subsubsection{Backbone Studies of Supervised Detection} 
As shown in Table~\ref{tab:accBackboneBuild}, our deeper FCDD-VGG16 outperformed in terms of the AUC, $F_1$, and precision, rather than the baseline CNN27 and other backbone-based deeper FCDDs in this disaster damage dataset for detecting building collapse, traffic incidents, fire, and flooding area. 
\begin{table}[h] 
\caption{Backbone ablation studies on disaster damage detection using our proposed deeper FCDDs.} 
\label{tab:accBackboneBuild} 
\centering 
\begin{tabular}{|c|c|c|c|c|} 
\hline 
\textbf{Backbone} & \textbf{AUC} & \boldmath{$F_1$} & \textbf{Precision} & \textbf{Recall} \\ 
\hline 
CNN27 & 0.9433 & 0.7896 & 0.7523 & 0.8307 \\ \hline 
\textbf{VGG16} & \textbf{0.9969} & \textbf{0.9622} & \textbf{0.9589} & \textbf{0.9655} \\ 
ResNet101 &0.9916 & 0.9323 & 0.8985 & 0.9687 \\ 
Inceptionv3 &0.9925 & 0.9319 & 0.9189 & 0.9453 \\ \hline 
\end{tabular} 
\end{table} 
 
\subsubsection{Imbalanced-to-unsupervised Training Results} 
As shown in Table~\ref{tab:accImbalanceBuild}, we implemented ablation studies on the imbalanced data containing smaller anomalous and relatively large normal images. In this study, we applied our deeper FCDD-VGG16 to achieve high performance in the aforementioned supervised results.   
Compared with the balanced case with a positive ratio of 1/1, we found that there was an applicable range from an imbalanced ratio of 1/2 to 1/16, where the accuracy was consistently high.  
However, in the extremely imbalanced range of 1/32–1/1300, the accuracy was inferior to that of the applicable range. The rare positive ratio of 1/32 represents imbalanced data containing quite a little 41 anomalous images and relatively large 1300 normal images. In this case, additional anomalous images were acquired and added to the initial dataset.    
The marginal gain in accuracy was relatively high when rare events in the devastated images were added.  
\begin{table}[h] 
\caption{Imbalanced data studies using our deeper FCDD-VGG16 for Disaster damage detection $N_7=1300$.} 
\label{tab:accImbalanceBuild} 
\centering 
\begin{tabular}{|c|c|c|c|c|} 
\hline 
\textbf{Positive ratio} & \textbf{AUC} & \boldmath{$F_1$} & \textbf{Precision} & \textbf{Recall} \\ 
\hline 
\textbf{1/1(ano.$N_{11}$)} & \textbf{0.9844} & \textbf{0.9491} & \textbf{0.9410} & \textbf{0.9575} \\ \hline 
1/2(ano.650) & 0.9850 & 0.9398 & 0.9422 & 0.9375 \\  
1/4(ano.325) & 0.9843 & 0.9422 & 0.9469 & 0.9375 \\  
1/8(ano.163) & 0.9809 & 0.9423 & 0.9447 & 0.9400 \\  
1/16(ano.81)& 0.9766 & 0.9413 & 0.9401 & 0.9425 \\ \hline 
\textbf{1/32(ano.41)}&\textbf{0.9606} & \textbf{0.9075} & \textbf{0.9075} & \textbf{0.9075} \\  
\textbf{1/64(ano.20)}&\textbf{0.9444} & \textbf{0.8924} & \textbf{0.8826} & \textbf{0.9025} \\  
\textbf{1/128(ano.10)}&\textbf{0.8971} & \textbf{0.8221} & \textbf{0.8484} & \textbf{0.7975} \\  
\textbf{1/$N_{11}$(ano.1)} &\textbf{0.6870} & \textbf{0.5783} & \textbf{0.6721} & \textbf{0.5075} \\ \hline 
\end{tabular} 
\end{table} 
 
\subsubsection{Damage-mark Heatmaps on Disaster Damage} 
We visualised the damage features using Gaussian upsampling in our deeper FCDD-VGG16 network. In addition, we generated a histogram of the anomaly scores of the test images in the imbalanced case with a positive ratio of $1/16$. In Fig. \ref{fig:rawBuild}, a damage-mark explanation is represented. The red region in the heatmap represents four classes of disaster features: fire and smoke, building collapse, traffic incidents, and flooding.  
Fig. \ref{fig:histBuild} illustrates that a few overlapping bins exist in the boundary of the ordinary class and disaster class along the horizontal anomaly scores. Thus, to detect devastating features, the score ranges were well separated in the disaster-damage dataset. 
 
\begin{figure}[h] 
\centering 
\includegraphics[width=0.37\textwidth]{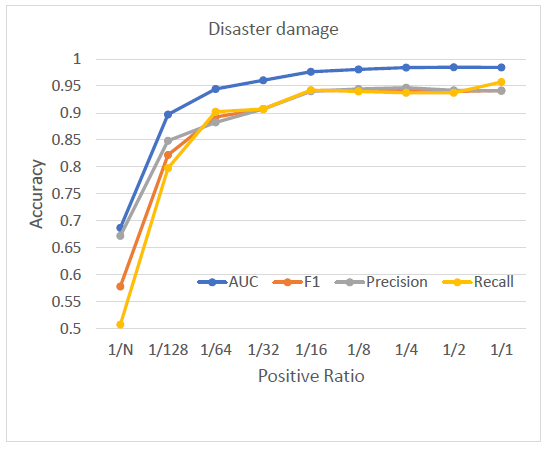} 
\caption{Anomalous vision mining studies on disaster  damage indicate that the more anomalies of damage vision, the higher the performance of anomaly detection.} 
\label{fig:EffectBuild} 
\end{figure} 
 
\subsubsection{Feedback Effect on Damage Class Mining} 
As shown in Figure \ref{fig:EffectBuild}, from the viewpoint of accuracy, imbalanced studies on disaster damage imply that all accuracies consistently converged into significant phases.  
Ranging with a positive ratio of less than $1/8$, we can understand that there were damage vision mining opportunities with an accuracy gain. By contrast, when ranging with a positive ratio of over $1/4$, it shifted in the over-mining phase without any gain in accuracy. The former phase of damage vision mining opportunities has become beneficial owing to its promising advantage of higher accuracy. 
 
\subsubsection{Embedding Damage Representation} 
\begin{figure}[h] 
\centering 
\includegraphics[width=0.35\textwidth]{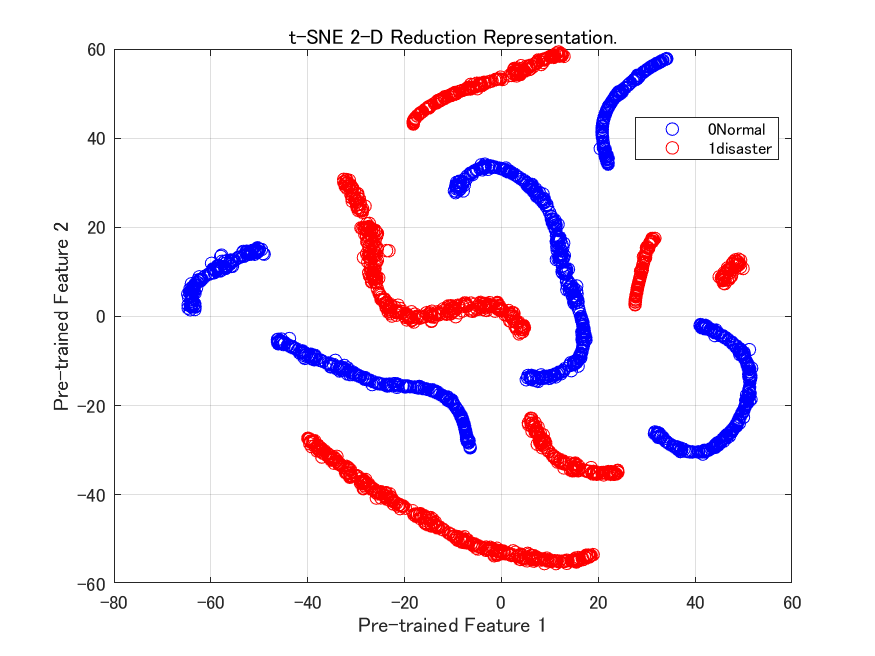} \\ 
\includegraphics[width=0.35\textwidth]{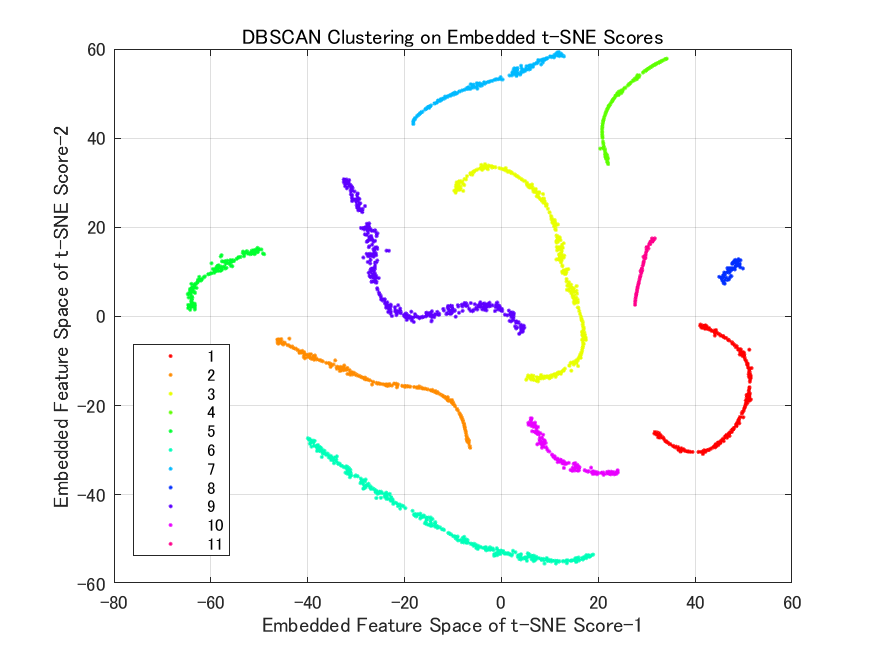} 
\caption{MN-pair contrastive damage representation (top) and density-based clustering (bottom) of disaster damage.} 
\label{fig:mndbBuild} 
\end{figure}

As shown in Figure \ref{fig:mndbBuild}, we analysed the feature imbalance in the disaster-damage embedding space and implemented our MN-pair contrastive damage-representation learning and density-based clustering.  
The number of disaster damage clusters increased by approximately 11 more than the initial four classes, that is, building collapse, flooding area, traffic incidents, and fire/smoke, which contained normal and anomalies.  
For the disaster-damage feature, a few narrow clusters were distributed in the embedding space. 
 
\subsection{Hurricane Damage} 
\subsubsection{Backbone Studies of Supervised Detection} 
As shown in Table~\ref{tab:accBackboneHurri}, our deeper FCDD-ResNet101 outperformed in terms of the $F_1$, and recall rather than the baseline CNN27 and other backbone-based deeper FCDDs in this satellite imagery dataset by remote sensing for detecting hurricane damage. 
\begin{table}[h] 
\caption{Backbone ablation studies on hurricane damage detection using our proposed deeper FCDDs.} 
\label{tab:accBackboneHurri} 
\centering 
\begin{tabular}{|c|c|c|c|c|} 
\hline 
\textbf{Backbone} & \textbf{AUC} & \boldmath{$F_1$} & \textbf{Precision} & \textbf{Recall} \\ 
\hline 
CNN27 & 0.9858 & 0.9297 & 0.9172 & 0.9425 \\ \hline 
VGG16 & 0.9980 & 0.9671 & 0.9770 & 0.9575 \\ 
\textbf{ResNet101} &\textbf{0.9982} & \textbf{0.9753} & \textbf{0.9611} & \textbf{0.9900} \\ 
Inceptionv3 &0.9979 & 0.9623 & 0.9671 & 0.9575 \\ \hline 
\end{tabular} 
\end{table} 
 
\subsubsection{Imbalanced-to-unsupervised Training Results} 
As shown in Table~\ref{tab:accImbalanceHurri}, we implemented ablation studies on the imbalanced data containing smaller anomalous and relatively large normal images. In this study, we applied our deeper FCDD-ResNet101 and achieved high performance in the aforementioned supervised results.   
Compared to the balanced case of the positive ratio 1/1, we found that there was an applicable range from the imbalanced ratio 1/2 to 1/16, where the accuracy of recall was consistently more than 97\%.  
However, in the extremely imbalanced range of 1/32–1/1300, the accuracy was inferior to that of the applicable range. The rare positive ratio of 1/32 represents imbalanced data containing quite a little 41 anomalous images and relatively large 1300 normal images. In this case, additional anomalous images were acquired and added to the initial dataset.    
The marginal gain in accuracy was relatively high with the addition of remote sensing satellite imagery of hurricane damage.  
\begin{table}[h] 
\caption{Imbalanced data studies using our deeper FCDD-ResNet101 for Hurricane damage detection $N_8=1300$.} 
\label{tab:accImbalanceHurri} 
\centering 
\begin{tabular}{|c|c|c|c|c|} 
\hline 
\textbf{Positive ratio} & \textbf{AUC} & \boldmath{$F_1$} & \textbf{Precision} & \textbf{Recall} \\ 
\hline 
\textbf{1/1(ano.$N_{12}$)} & \textbf{0.9982} & \textbf{0.9753} & \textbf{0.9611} & \textbf{0.9900} \\ \hline 
1/2(ano.650) & 0.9986 & 0.9851 & 0.9754 & 0.9950 \\  
1/4(ano.325) & 0.9981 & 0.9739 & 0.9656 & 0.9825 \\  
1/8(ano.163) & 0.9956 & 0.9558 & 0.9375 & 0.9750 \\  
1/16(ano.81)& 0.9962 & 0.9641 & 0.9535 & 0.9750 \\ \hline 
\textbf{1/32(ano.41)}&\textbf{0.9889} & \textbf{0.9564} & \textbf{0.9815} & \textbf{0.9325} \\  
\textbf{1/64(ano.20)}&\textbf{0.9818} & \textbf{0.9420} & \textbf{0.9492} & \textbf{0.9350} \\  
\textbf{1/128(ano.10)}&\textbf{0.9734} & \textbf{0.9324} & \textbf{0.9702} & \textbf{0.8975} \\  
\textbf{1/$N_{12}$(ano.1)} &\textbf{0.8155} & \textbf{0.7376} & \textbf{0.6807} & \textbf{0.8050} \\ \hline 
\end{tabular} 
\end{table} 
 
\begin{figure}[h] 
\centering 
\includegraphics[width=0.4\textwidth]{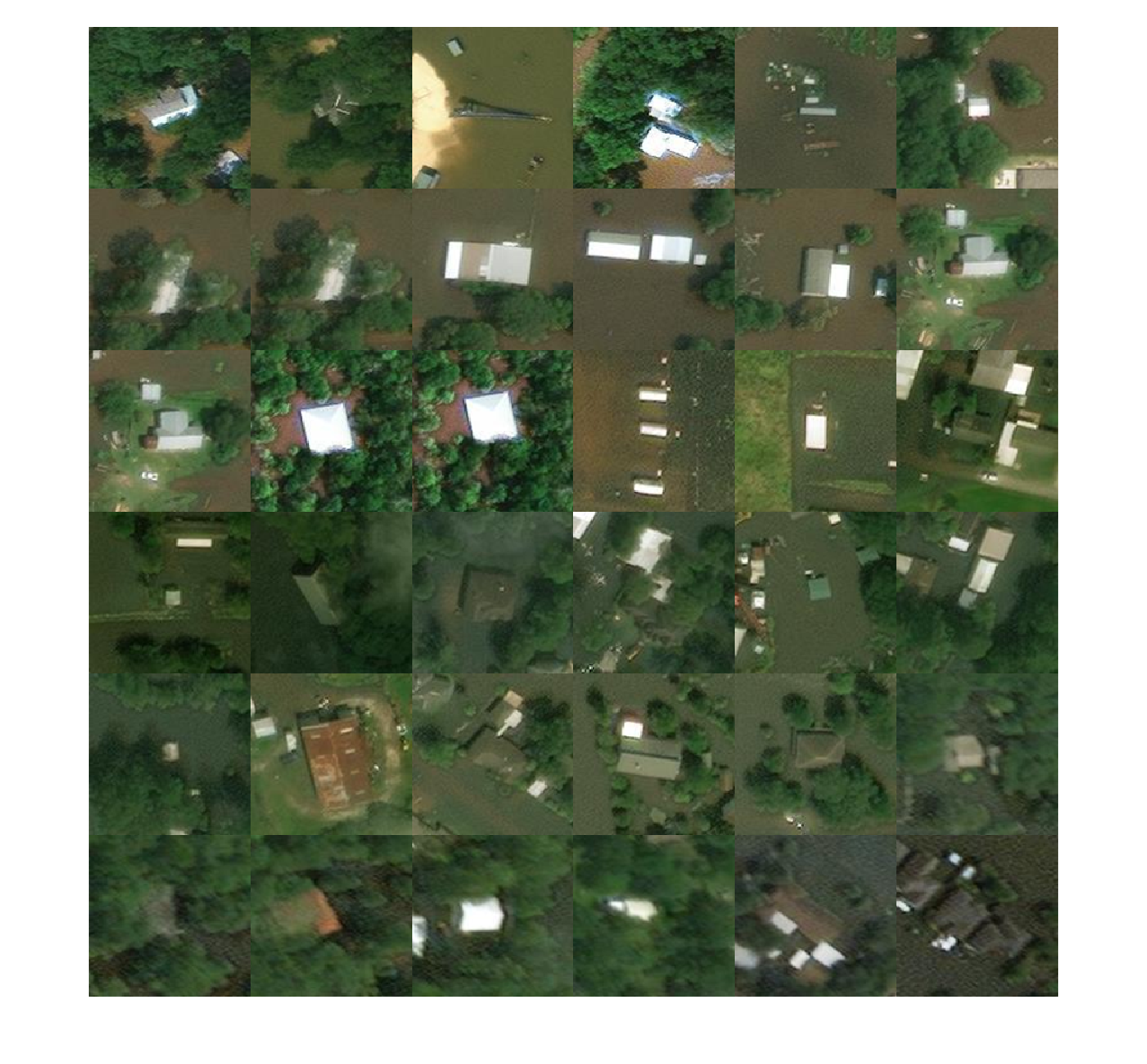} \\ 
\includegraphics[width=0.4\textwidth]{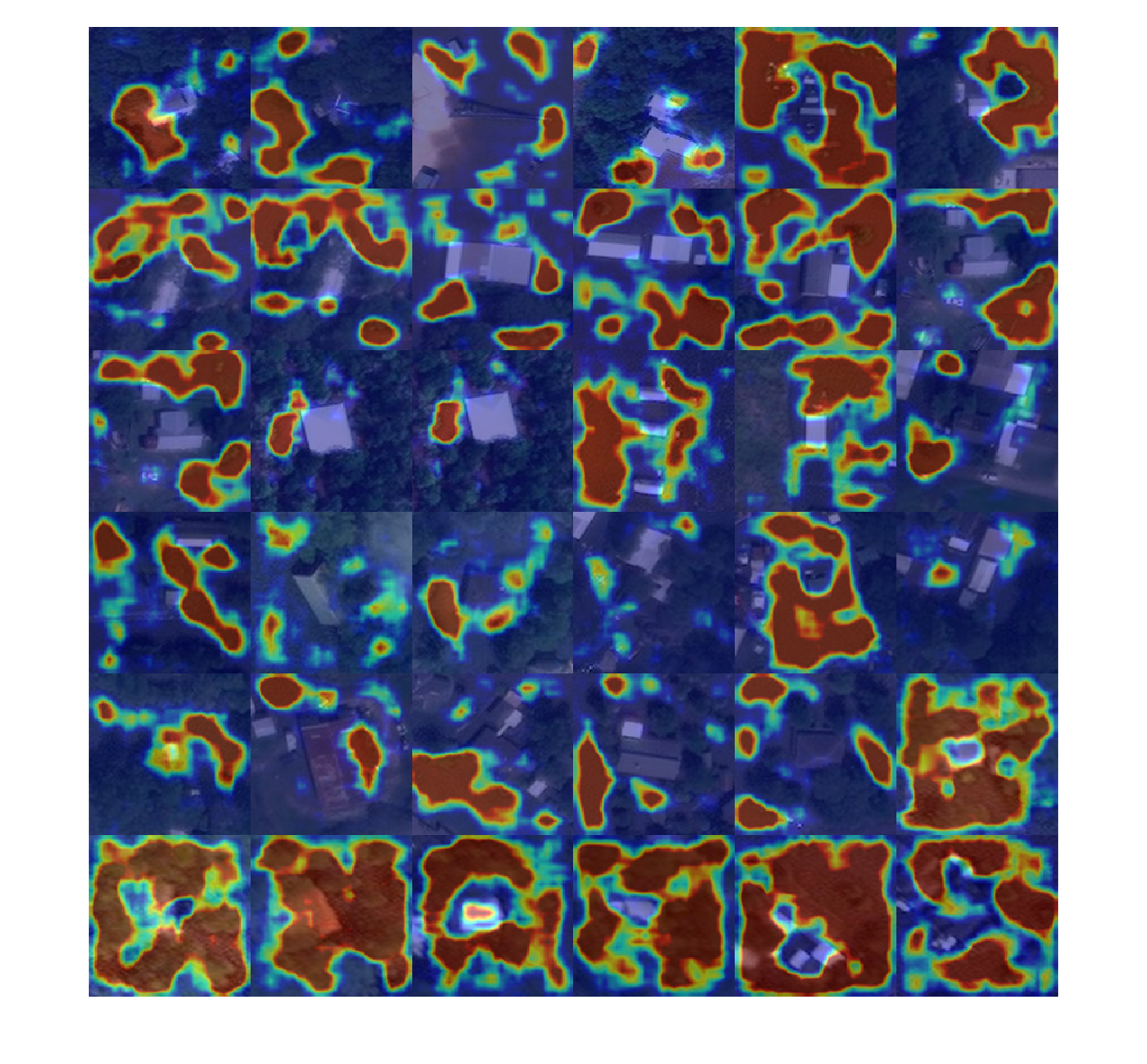} 
\caption{Imbalanced damage images (top) with positive ratio $1/16$, and damage-mark heatmaps (bottom) of hurricane damage using our deeper FCDD-ResNet101.} 
\label{fig:rawHurri} 
\end{figure} 
\begin{figure}[h] 
\centering 
\includegraphics[width=0.37\textwidth]{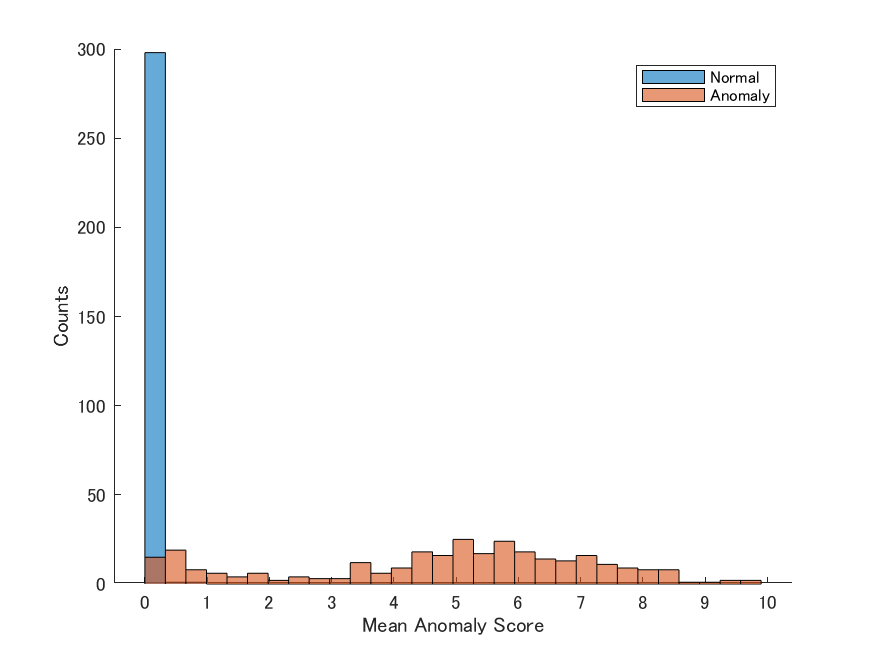} 
\caption{Histogram of hurricane damage scores using our deeper FCDD-ResNet101 corresponding to the imbalanced damage images with positive ratio $1/16$.} 
\label{fig:histHurri} 
\end{figure} 
 
\begin{figure}[h] 
\centering 
\includegraphics[width=0.37\textwidth]{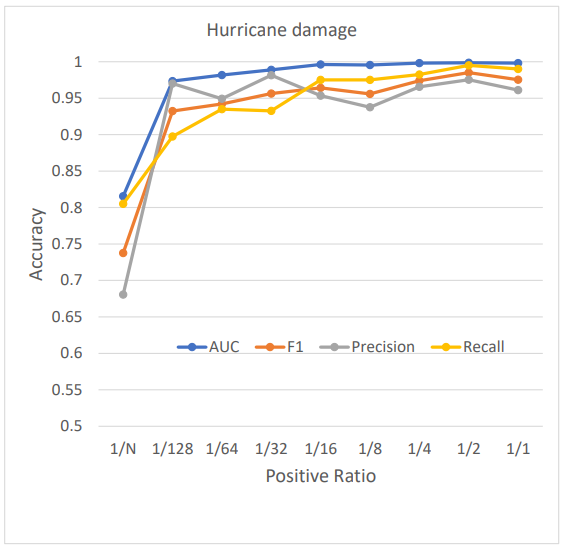} 
\caption{Anomalous vision mining studies on hurricane damage, indicates that the more anomalies of damage vision, the higher performance of anomaly detection.} 
\label{fig:EffectHurri} 
\end{figure} 

\begin{figure}[h] 
\centering 
\includegraphics[width=0.35\textwidth]{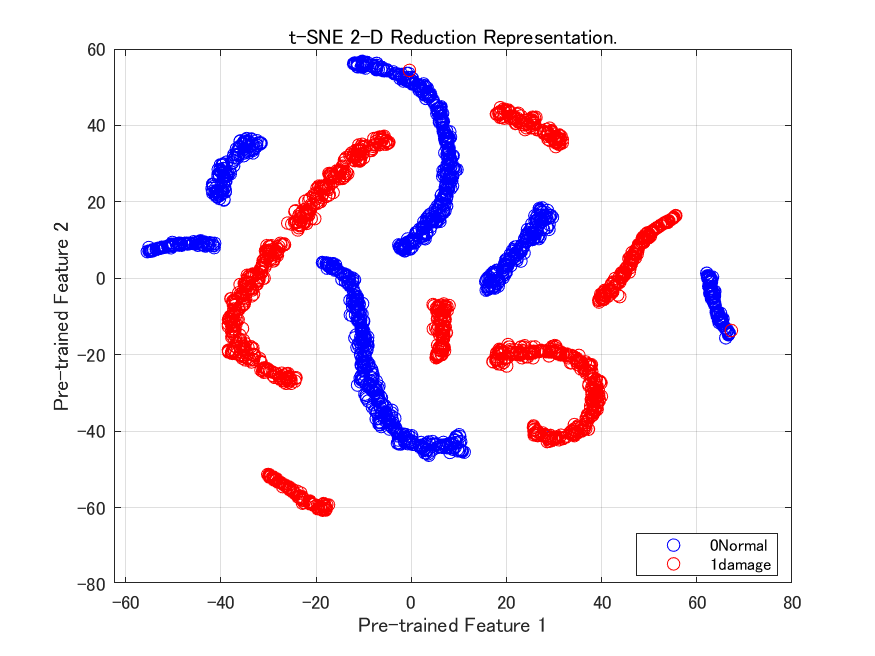} \\ 
\includegraphics[width=0.35\textwidth]{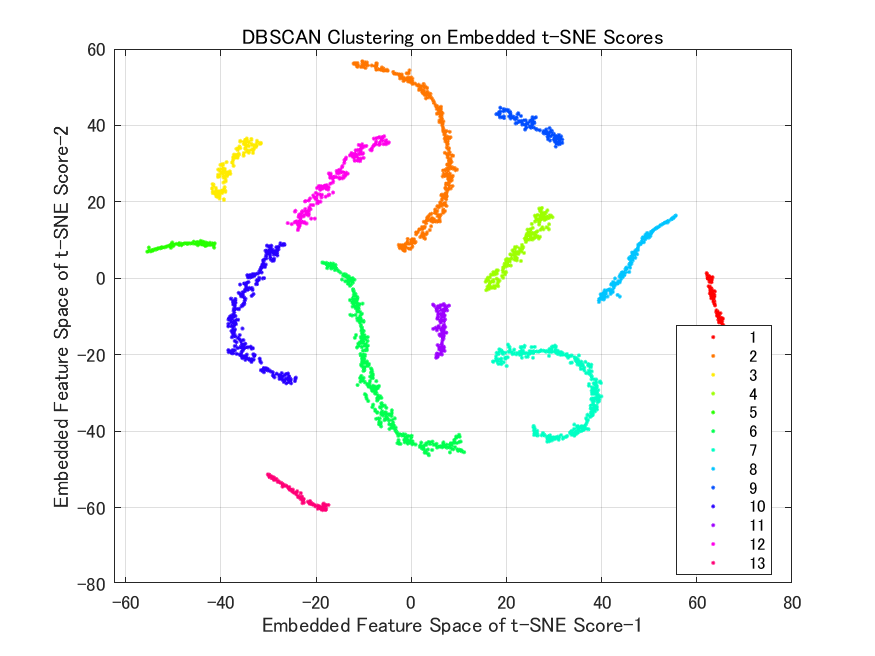} 
\caption{MN-pair contrastive damage representation (top) and density-based clustering (bottom) of hurricane damage.} 
\label{fig:mndbHurri} 
\end{figure} 
 
\subsubsection{Damage-mark Heatmaps on Hurricane} 
We visualised the damage features using Gaussian upsampling in our deep FCDD-ResNet101 network. In addition, we generated a histogram of the anomaly scores of the test images in the imbalanced case with a positive ratio of $1/16$. In Fig. \ref{fig:rawHurri}, a damage-mark explanation is represented. The red region in the heatmap represents hurricane damage to flooding features in the remote sensing satellite imagery.  
Fig. \ref{fig:histHurri} illustrates that a few overlapping bins exist in the boundary of the normal class and hurricane-damaged class along the horizontal anomaly scores. Thus, to detect hurricane damage due to flooding, the score range was well separated in the remote-sensing satellite imagery dataset. 
 
\subsubsection{Feedback Effect on Damage Class Mining} 
As shown in Figure \ref{fig:EffectHurri}, from the viewpoint of all accuracies, the imbalanced studies on hurricane imply that all accuracies consistently converged into significant phases.  
Ranging by a positive ratio of less than $1/32$, we could understand that there were damage vision mining opportunities with an accuracy gain in terms of the AUC. By contrast, with a positive ratio of over $1/16$, it shifted in the over-mining phase without any gain in the AUC. The former phase of damage vision mining opportunities has become beneficial owing to its promising advantage of higher accuracy. 
 
\subsubsection{Embedding Damage Representation} 
As shown in Figure \ref{fig:mndbHurri}, we analysed the feature imbalance in the hurricane damage embedding space and implemented our MN-pair contrastive damage representation learning and density-based clustering.  
Surprisingly, the number of hurricane damage clusters increased to 13 rather than the initial two classes because of variational features that contained trees, devastated buildings, and flooding areas.  
In the hurricane damage feature, many narrow clusters were distributed in the embedding space. 
 
\begin{table*}[h] 
\caption{$1/a$ Few-shot anomalies feedback effect on accuracy AUC studied on the former 6 datasets.} 
\label{tab:accFeedEffect1-6} 
\centering 
\begin{tabular}{|c|c|c|c|c|c|c|} 
\hline 
\textbf{Model} & \textbf{Blood} & \textbf{Lung} & \textbf{Breast} & \textbf{Driving} & \textbf{Wood} & \textbf{Concrete}\\ 
\hline 
PaDiM      & 0.7378 & 0.8285 & 0.5867 & 0.9715 & 0.6132 & 0.6106 \\  
PatchCore& 0.9415 & 0.7180 & 0.4108 & 0.9204 & 0.5758 & 0.7179 \\  
1/$N_d$ one-shot  & 0.9174 & 0.7494  & 0.8331 & 0.8378 & 0.6136& 0.7927 \\  
$1/(2a^{\ast})$ few-shot & 0.9919 & 0.9911 & 0.9555 & 0.9916 & 0.8947 & 0.8956 \\  
$\mathbf{1/a^{\ast}}$ \textbf{few-shot} & \textbf{0.9907} & \textbf{0.9908} & \textbf{0.9622} & \textbf{0.9965} & \textbf{0.9101} & \textbf{0.9147} \\ \hline 
$\mathbf{1/a^{\ast}}$ &\textbf{1/16} & \textbf{1/16} & \textbf{1/8} & \textbf{1/16} & \textbf{1/16} & \textbf{1/8} \\ 
adapted backbone & ResNet101 & ResNet101 & ResNet101 & Inceptionv3 & VGG16 & VGG16 \\ 
\# feature clusters & \#9 & \#12 & \#13 & \#10 & \#10 & \#21 \\ \hline 
\end{tabular} 
\end{table*} 
\begin{table*}[h] 
\caption{$1/a$ Few-shot anomalies feedback effect on accuracy AUC studied on the latter 6 datasets.} 
\label{tab:accFeedEffect7-12} 
\centering 
\begin{tabular}{|c|c|c|c|c|c|c|} 
\hline 
\textbf{Model} & \textbf{Logical} & \textbf{Vegetable} & \textbf{Plant} & \textbf{River} & \textbf{Disaster} & \textbf{Hurricane}\\ 
\hline 
PaDiM      & 0.5307 & 0.6021 & 0.9042 & 0.7753 & 0.8134 & 0.2959 \\  
PatchCore& 0.6885 & 0.8257 & 0.8754 & 0.6288 & 0.7435 & 0.5364 \\  
1/$N_d$ one-shot               & 0.5520 & 0.7886 & 0.7882 & 0.8221 & 0.6870 & 0.8155 \\  
$1/(2a^{\ast})$ few-shot& 0.7881 & 0.9771 & 0.9988 & 0.9588 & 0.9606 & 0.9889 \\  
$\mathbf{1/a^{\ast}}$ \textbf{few-shot} & \textbf{0.8504} & \textbf{0.9838} & \textbf{0.9983} & \textbf{0.9457} & \textbf{0.9766} & \textbf{0.9962} \\ \hline 
$\mathbf{1/a^{\ast}}$ &\textbf{1/8} & \textbf{1/16} & \textbf{1/8} & \textbf{1/4} & \textbf{1/16} & \textbf{1/16} \\ 
adapted backbone & VGG16 & ResNet101 & Inceptionv3 & ResNet101 & VGG16 & ResNet101 \\ 
\# feature clusters & \#11 & \#10 & \#14 & \#11 & \#11 & \#13 \\ \hline 
\end{tabular} 
\end{table*} 

\subsection{Few-shot Anomalies Feedback Effect} 
In Table~\ref{tab:accFeedEffect1-6}, we summarized the aforementioned six trained results in terms of AUC accuracy compared to our imbalanced few-shot detection using the adapted backbone-based deeper FCDDs with previous unsupervised normalising methods. In addition, we indicate the number of feature clusters from contrastive learning results using our MN pair contrastive damage representation.  
In the case of two damage vision regarding blood infection and driving distraction, the one-shot anomaly detection method was inferior to the unsupervised normalising methods. However, we found that $1/a^{\ast}$ few-shot anomaly detection, where the positive ratios were $1/8$ and  $1/16$, achieved higher accuracy without over-mining the damaged vision.  
Furthermore, in the case of breast cancer and concrete deterioration, there were more feature clusters, and a more positive ratio of $1/a$ few-shot anomalies was required in the imbalanced feature space, as shown in the embedding scores by the t-SNE.     
 
In Table~\ref{tab:accFeedEffect7-12}, we summarise the remaining six trained results in terms of AUC accuracy compared to our imbalanced few-shot detection using the adapted backbone-based deeper FCDDs with the previous unsupervised normalising methods. In addition, we indicate the number of feature clusters from contrastive learning results using our MN pair contrastive damage representation.  
In the case of the four damage vision, such as MVTec LOCO products, vegetable damage, plant leaf infection, and disaster damage, the one-shot anomaly detection method was inferior to the unsupervised normalising methods. Only one-shot anomaly learning was insufficient for achieving higher performance than the previous normalising methods.  
However, we found that the $1/a^{\ast}$ few-shot anomaly detection, where the positive ratios ranged from $1/16$ to $1/4$, achieved higher accuracy without over-mining the  damage vision.  
Furthermore, in the case of plant leaf infection, there was a greater number of feature clusters, and a more positive ratio of $1/a$ few-shot anomalies was required in the imbalanced feature space using embedding scores.     
 
\section{Concluding Remarks} 
\subsection{$\mathbf{1/a}$ Few-shot Anomalies Feedback for Class Imbalance} 
We developed an imbalanced-vision application to automate one-class anomaly detection. To ensure the feasibility of imbalanced damage vision datasets for typical targets of medical diseases, material deterioration, hazardous behaviour, plant diseases, river sludge, and disaster damage. We found that there was an adapted backbone when our deeper FCDDs were used for imbalanced damage vision detection. In addition, we visualised the damage-mark heatmaps using direct Gaussian upsampling of the receptive field of the FCN. This created a damage-mark heatmap for visual explanation without annotating the target of damage in the localised regions.  
Furthermore, we compared the accuracy of our method with that of previous typical models of patch-wise embedding similarity. Consequently, the $1/a$ few-shot anomaly feedback using our deeper FCDDs outperformed previous normalising methods, that is, PaDiM and PatchCore. 
 
From our imbalanced studies, compared with the balanced case with a positive ratio of $1/1$, we found that there was an appropriate positive ratio of $1/a$, where the accuracy was consistently high. However, the extremely imbalanced range was from one-shot to $1/2a$, the accuracy of which was inferior to that of the applicable ratio. In contrast, when ranging with a positive ratio over $2/a$, it shifted in the over-mining phase without an effective gain in accuracy.   
Thus, we found an effective positive ratio of anomalies versus a relatively large normal class without over-mining to ensure stable accuracy and avoid wasting time and resources on damage vision mining.  
 
\subsection{Greater Imbalanced Embedding Clusters, More Anomalies Feedback} 
We considered a {\it feature imbalance} in which minor clusters were sparsely distributed in the feature embedding space. The author implemented our MN-pair contrastive damage representation learning and density-based clustering. Surprisingly, the number of damage feature clusters increased, rather than the number of initial predefined classes that contained normal and anomalies, and the damage feature clusters were distributed into a narrow region in the embedding space. 
The authors hypothesised that the greater the damage feature imbalance in the embedding space, the more few-shot anomaly feedback is required.  
As a result, as summarised in 12 damage vision experiments, in the three cases of breast cancer, concrete deterioration, and plant leaf infection, there were a greater number of damage feature clusters in the embedding space, and a more positive ratio of $1/a$ few-shot anomaly feedback was required. 
     
\subsection{Limitation and Robustness for Unseen Damage} 
This study applied a specific anomaly detection model and limited the targets of available datasets in the present. 
Another unbalanced feature remains for practical use in each domain. Examples include medical diseases, material deterioration, damage from natural disasters, and environmental damage.  
In the outdoor field, we should consider the temperature variation of damaged vision in the background; for example, the winter season may include unseen noise, that is, decayed grass and snow.  
Damage vision mining in imbalanced rare events requires more time to achieve stable accuracy. To overcome this hurdle, the anomaly score can be used in edge devices for the effective data acquisition of rare classes. By unseen data mining, only the anomalous vision that has damage marks with particularly high anomaly scores, the data acquisition process can be made more efficient. 
From the key results of our imbalanced studies, we found that an appropriate positive ratio of $1/a^{\ast}$ resulted in a promising accuracy gain as a damage-vision mining merit. In contrast, ranging with the positive ratio over $2/a^{\ast}$, the accuracy has not been further gained effectively. This implies that the over-mining phase is not beneficial for the early stopping of that phase of further data acquisition.   
 
We are going to tackle the opportunities of unseen damage vision mining and improve the robustness of damage detection applications to overcome the combination of imbalance type in damage vision, based on our initial studies on flood inflow using regression\cite{Yasuno2021L2norm,Yasuno2020RainCode}, typhoon damage and river scum using image classification\cite{Yasuno2020Natural,Yasuno2022River},  
construction material condition clustering using contrastive metric learning\cite{Yasuno2023MNPair}, concrete exfoliation and snow-covered roads using semantic segmentation\cite{Yasuno2019Popouts,Yasuno2020GeneraSyn,Yasuno2020Perpixel,Yasuno2021Road}, and earthquake disasters and bridge-slab deterioration using anomaly detection\cite{Yasuno2019Color,Yasuno2021Bridge,Yasuno2020Generative}. 
However, these are independent domain-based damage vision modelling and application tasks for specific users.   
 
\subsection{Damage Vision Feedback on Multi-modality} 
Building a foundation model for damage vision tasks using large-scale available damage vision data is important. Since 2021, the large language models have been making fast progress utilizing the vision transformers and diffusion models, such as GLIDE \cite{Nichol2022glide}, DALL-E2 \cite{Ramesh2022hierarchical}, PaLM \cite{Chowdhery2022palm}, GPT-4 \cite{Openai2023gpt4}, ImageBind \cite{Girdhar2023imagebind}, Segment Anything Model (SAM) \cite{Kirillov2023segment}, and PaLM-SayCan \cite{Ahn2022saycan}. To improve the fairness and explainability of damage vision tasks, we plan to create a beneficial $1/a^{\ast}$few-shot anomaly feedback algorithm using pre-trained foundation models with multi-modality connecting damage vision, diagnostic text, defective sounds, and inspection robotics. We believe that appropriate feedback by $1/a^{\ast}$ few-shot anomalies results in a fair and unbiased diagnosis of damaged vision with an imbalanced feature problem. 
 
\section*{Acknowledgment} 
We gratefully acknowledge the conductive comments of Professor George A. Tsihrintzis. We also thank the editorial team of Machine Learning Paradigms for the opportunity of imbalanced damage-vision studies. 
The authors wish to thank MathWorks and Computer Vision Toolbox Team, Takuji Fukumoto for providing helpful MATLAB resources for Automated Visual Inspection.   
{\small 
\bibliographystyle{ieee_fullname} 
\bibliography{mlparabib} 
} 
 
\end{document}